\documentclass[11pt]{article}

\usepackage[preprint]{acl}
\usepackage{times}
\usepackage{latexsym}
\usepackage[T1]{fontenc}
\usepackage[utf8]{inputenc}
\usepackage{microtype}
\usepackage{amsmath,amssymb,amsthm}
\usepackage{booktabs}
\usepackage{array}
\usepackage{longtable}
\usepackage{multirow}
\usepackage{graphicx}
\usepackage{xcolor}
\usepackage{enumitem}
\usepackage{float}
\usepackage{listings}
\newcommand{\A}[1]{\href{https://oeis.org/A#1}{A#1}}
\newcommand{\nov}[1]{\textit{N#1}}
\newcommand{\novna}{---}
\newcommand{\sig}[1]{\textit{S#1}}
\newcommand{\lantern}{\textsc{Lantern}}

\title{\lantern{}: Illuminating Hidden Mathematical Knowledge\\
in Language Models}

\author{Pavel Tikhonov, Elena Tutubalina, Ivan Oseledets,\\\textbf{Dmitry I. Ignatov, Mikhail Seleznyov}}
\date{}

\begin{document}
\maketitle

\begin{abstract}
Language models can now prove theorems, but people still decide which problems to pursue. We ask whether a model's internal representations can help identify promising mathematical connections. We develop \lantern{}, a fast, cost-efficient pipeline that uses a classifier over pretrained-model activations to rank candidate relations, followed by staged filtering, hypothesis generation, executable verification, and analytical checking. Applied to the On-Line Encyclopedia of Integer Sequences (OEIS), \lantern{} ranked 50 million pairs among 10,000 frequently referenced sequences and produced 62 verified relations between pairs without an existing OEIS cross-reference. A content screen retained 13 relations worth presenting; nine of these are informative or insightful, including four which are entirely novel to the best of our knowledge: none appears in the OEIS or in our targeted literature search. The entire end-to-end process including classifier training, candidate ranking, filtering and verification took under 8 hours.
\end{abstract}

\begin{figure}[!t]
\centering
\includegraphics[width=\linewidth]{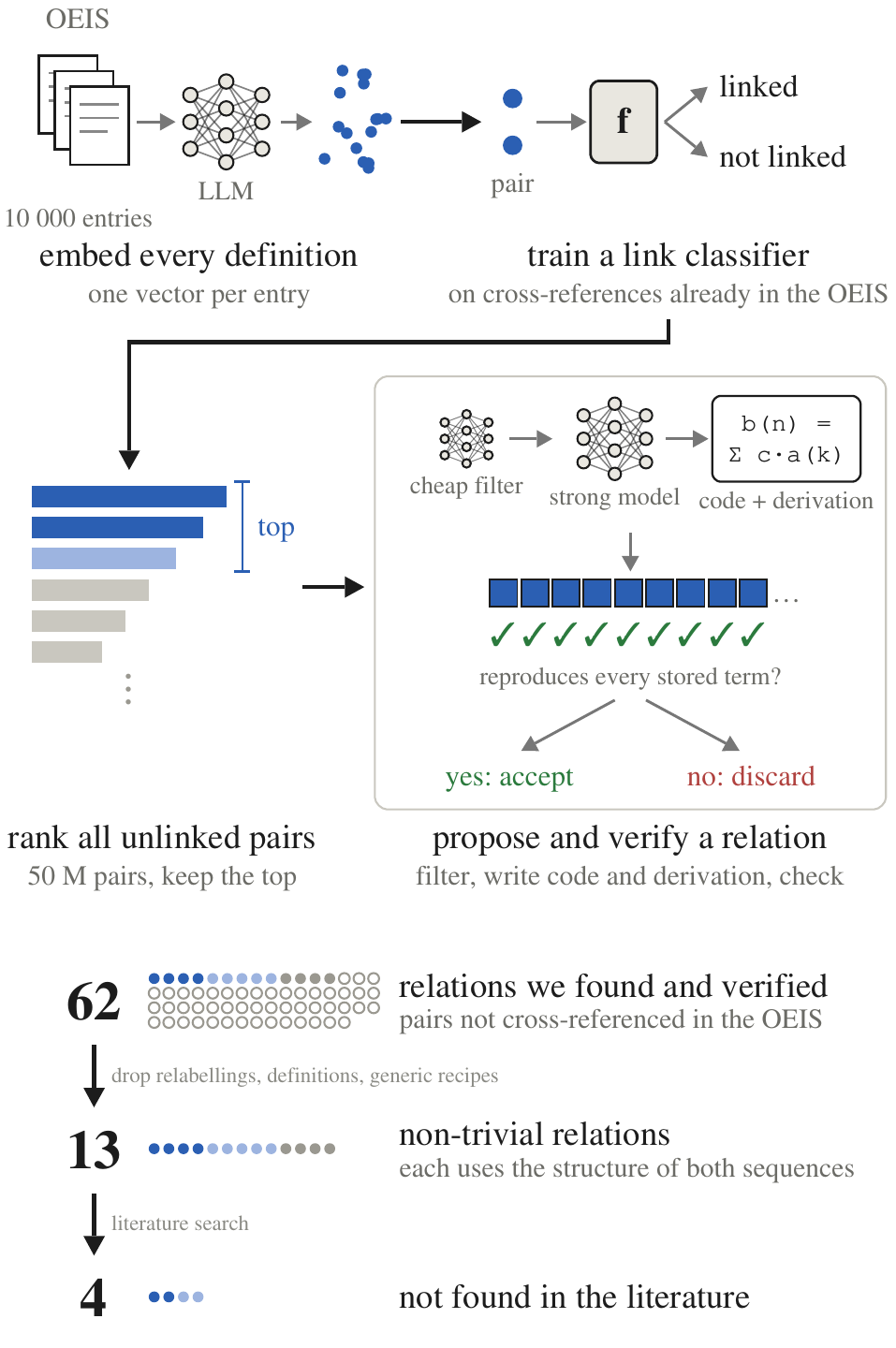}
\caption{Overview of \lantern{}. A classifier over pretrained LLM representations guides the search for relations between OEIS entries with no existing cross-reference. We found and verified 62 such relations; 13 of them are non-trivial, and for four of these we found no prior statement in the literature.}

\label{fig:pipeline}
\end{figure}

\section{Introduction}
\label{sec:intro}
Language models now prove olympiad problems~\citep{hubert2025alphaproof}, find constructions for objectives that people specify~\citep{funsearch,georgiev2025exploration}, and resolve items from curated lists of open problems~\citep{deepmind2026proofsearch,oeisopen2026}. Still, a mathematician usually chooses what to look at. This has been called the bottleneck of AI-assisted mathematics~\citep{zheng2026problem} and the first criterion of machine discovery~\citep{he2024birch}. A number of systems choose for themselves: they take candidates from statements already written in the literature~\citep{zheng2026problem}, from missing edges of a knowledge graph scored by its topology~\citep{hygrail2026}, or from the model's own generation within a topic~\citep{leanconjecturer,long2026meca}. We propose a different approach. It rests on the assumption that during pretraining a language model has already learned properties of mathematical objects, that this knowledge is encoded in its internal representation, and that the representation can be used directly to search for relations between objects: we rank every pair of objects in a knowledge base by what the model's hidden states say about them.

On the OEIS~\citep{oeis}, whose cross-references record the relations people have noticed, we trained a linear classifier on Qwen3-32B activations with the substantive cross-references as labels and scored all 50\,million pairs of the 10,000 most-referenced entries; an agent turned the top pairs into executable relations checked on every stored term, and an audit checked each against the text of the entries (Fig.~\ref{fig:pipeline}). The technical pipeline produced 62 verified statements between entries the OEIS does not connect. We then applied a content screen to remove bookkeeping, definitional, and mechanically reproducible relations. The screen retained 13 relations; a two-axis assessment classifies nine as informative or insightful. Four of these nine were not found in the OEIS or in our targeted literature search: two are informative and two are insightful. On cross-references that people added after the training data, ranked against unlinked pairs of the same topics, the classifier stays above a scorer trained the same way on the surface text of the definitions, and a direct question to the same model ranks held-out pairs about as well as the classifier (Sec.~\ref{sec:controls}).

Our main contributions are as follows:
\begin{itemize}[nosep,leftmargin=*]
\item We show that a pretrained language model already contains knowledge of previously undocumented relations between mathematical objects, and that this knowledge can be extracted from its hidden states.
\item We propose \lantern{}, a cost-effective and efficient method for discovering such relations.
\item Using our method, we obtained 62 technically verified relations between previously unlinked OEIS sequences. A content screen retained 13; nine are informative or insightful, of which four were not found in the OEIS or in our targeted literature search.
\end{itemize}

\section{Related Work}
\label{sec:rw}

\paragraph{Human-directed discovery.}
Most AI discovery systems, including as FunSearch, AlphaProof, and co-scientist agents~\citep{hubert2025alphaproof,funsearch,georgiev2025exploration,deepmind2026proofsearch,oeisopen2026,coscientist2026} rely on human-provided objectives. Recent work like \citet{zheng2026problem} reduces human input to high-level research directions by extracting and grading open statements from papers.

\paragraph{Self-proposed conjectures.}
Automated conjecture generation typically relies on axioms, self-play, LLM prompting, or symbolic enumeration~\citep{minimo,dong2025stp,leanconjecturer,chen2026problems,das2026hardness,long2026meca,wong2026nextrh,davila2026txgraffiti,ramanujan2021,svatos2023}. In these methods, candidates are newly generated, and ``novelty'' usually means absence from a formal library or relies on expert evaluation~\citep{zhang2026survey,he2024birch}.

\paragraph{Missing links as hypotheses.}
Predicting unrecorded connections between known concepts has roots in literature-based discovery~\citep{swanson1986} and graph-based link prediction~\citep{krenn2023,gu2024,frohnert2024,marwitz2026}. Architecturally, we are closest to HyGRAIL~\citep{hygrail2026}, which predicts missing edges in materials graphs using a GNN and an LLM.

\paragraph{Latent knowledge and the OEIS.}
Language models often encode latent knowledge beyond their surface text, as shown by material discovery via word embeddings~\citep{tshitoyan2019} and probing of hidden layers~\citep{schut2023,davies2021,venugopal2026,burns2022}. The OEIS, in turn, is a knowledge base that mathematicians have been building for decades, and extending an entry can itself take decades: the ninth Dedekind number (A000372) was computed 32 years after the eighth~\citep{jakel2023dedekind}, and the sequence is known only asymptotically. The database has been mined for new identities by transforming stored sequences and matching the results against other entries~\citep{nguyen2013mining}, and a coincidence of counts noticed in it has led to a proof: closure systems under the $T_1$ separation axiom (A334254) turned out to be in bijection with Davis' set-union lattices (A235604) and with Mapes' atomic lattices~\citep{ignatov2022cryptomorphism}. In machine learning the OEIS serves as a test set for recurrence inference with transformers~\citep{dascoli2022recurrence}, as a benchmark for term prediction~\citep{intseqbert2026,fact2022}, and as a source of formalized conjectures and agent tasks~\citep{deepmind2026proofsearch,oeisopen2026}.

To our knowledge, we are the first to combine these directions, mining a model's latent representations to predict and mathematically verify missing cross-references within the OEIS.
\section{Setup}
\label{sec:setup}
The main difficulty in searching for hypotheses with language models is that the hypotheses they produce are hard to verify. For our method we needed a test bed with three properties:
\begin{itemize}[nosep,leftmargin=*]
\item well-defined objects;
\item documented relations between the objects;
\item relatively cheap verification of generated hypotheses.
\end{itemize}

\begin{figure}[t]
\centering
\includegraphics[width=\linewidth]{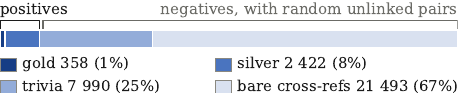}
\caption{Composition of the 32\,263 cross-referenced pairs in the corpus, text twins and the test pairs excluded. Only gold and silver, 9\% of them, are used as positives; trivia, bare cross-references and random unlinked pairs are the negatives. }
\label{fig:links}
\end{figure}

We chose to work with the On-Line Encyclopedia of Integer Sequences~\citep{oeis} because it has all three: each entry is a sequence given by a one-line definition and its first terms, the cross-references record which relations people have already noticed, and agreement with the stored terms gives a cheap first check of any exact claim about two sequences. The encyclopedia is also versioned, so for any cross-reference we can tell when it appeared.

\subsection{The corpus}
\label{sec:corpus}
We work with three snapshots of the encyclopedia: $S_0$ of 2025-04-30, the day after the embedding model's release; $S_1$ of 2026-02-17, the day after the release of Qwen3.5-27B; and $S_2$ of 2026-08-15, the latest available when we ran the experiments. We built the corpus from $S_2$. Of its 398\,320 entries we kept those with at least 25 terms, a definition of at least 15 characters that names no other entry, and none of the keywords \texttt{dead}, \texttt{dupe}, \texttt{allocated}, \texttt{recycled} or \texttt{uned}, which mark empty or retired entries. From these we selected the 10\,000 most often mentioned by other entries (at least 17 mentions); they form 50\,million pairs.

\subsection{Labeling the pairs}
\label{sec:labels}
\begin{figure}[!b]
\centering
\includegraphics[width=\linewidth]{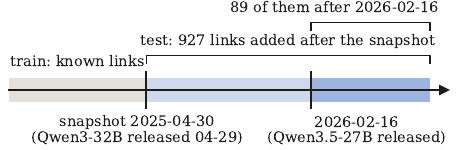}
\caption{The split by date. The classifier is trained on cross-references present at $S_0$ (2025-04-30, one day after the embedding model's release) and tested on links people added afterwards: 927 pairs (265 with no word shared between the definitions) and 89 more added in 2026. Of these, 74 were selected by a frontier model as substantive relations. }
\label{fig:timeline}
\end{figure}

\begin{figure*}[!t]
\centering
\includegraphics[width=\linewidth]{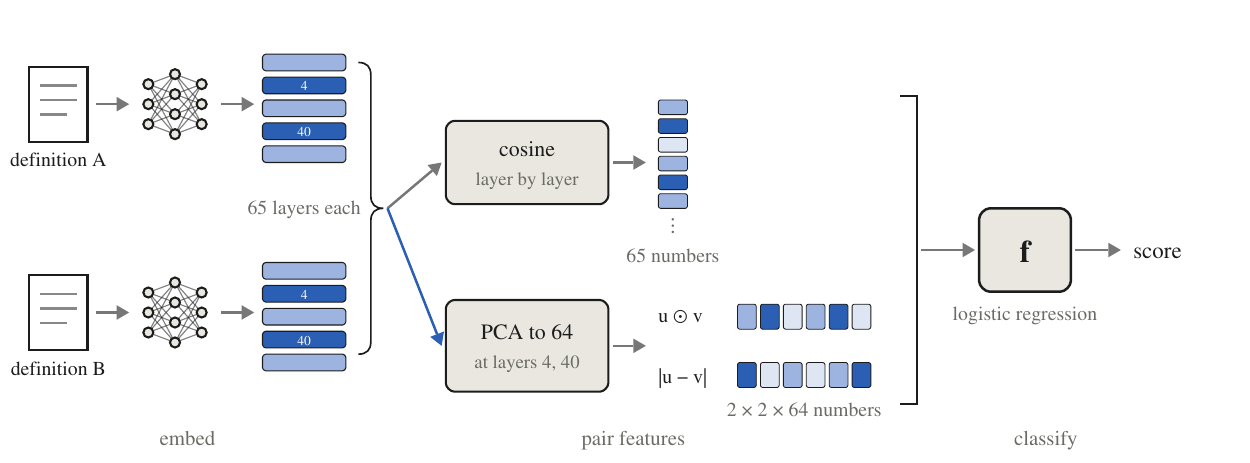}
\caption{The link classifier, drawn for Qwen3-32B. Each definition is embedded once into 65 layer vectors. A pair is described by the centred cosine between the two vectors at every layer and, at two layers (highlighted), by the element-wise product and absolute difference of the PCA-reduced vectors, as defined in Sec.~\ref{sec:classifier}; a logistic regression maps these features to a score.}
\label{fig:embed}
\end{figure*}

Entries come with cross-references, so the obvious way to label a pair is by whether the two entries are cross-referenced. Most cross-references, however, connect an entry to something trivially related (e.g.\ its generating function or a rescaled copy), with no interesting mathematical statement behind them, so a classifier trained on these labels may mostly learn to recognise such trivial links. We therefore classified every linked pair into one of three kinds (Fig.~\ref{fig:links}) by the mathematical depth of the relation.

We took every pair of corpus entries in which one entry mentions the other in $S_0$. From these we removed text twins, pairs whose definitions have a character-level TF-IDF cosine of at least 0.85 and whose relation is a change of parameter, and the pairs we reserve for testing below; 32\,263 pairs remained. In 10\,770 of them a formula or comment line of one entry names the partner, and in 21\,493 the partner appears only in the CROSSREFS section. We ran a screening model (Claude Sonnet 5) over the first kind; it read the two definitions and the verbatim formula and comment lines of both entries and assigned one of three grades:
\begin{itemize}[nosep,leftmargin=*]
\item \textbf{gold}: a theorem-grade identity, bijection, asymptotic or density result, or a stated conjecture. Example: \A{000584} $\times$ \A{123125}, where the link is Worpitzky's identity, $n^5$ written as a sum of Eulerian numbers times binomial coefficients.
\item \textbf{silver}: real but routine or textbook content. Example: \A{000332} $\times$ \A{000580}, where one sequence is a binomial convolution of the other.
\item \textbf{trivia}: a routine link, such as the same object in two representations, a rescaling or shift, or a template twin. Example: \A{000290} $\times$ \A{005408}, where the squares are the partial sums of the odd numbers.
\end{itemize}
Our rubric asked the screener to judge the relation itself, whatever the fame of the sequences. We then had every gold verdict re-read in full by a frontier model (Claude Fable 5.1); 91--94\% held, the rest we downgraded.

We use gold and silver, 2\,780 pairs or 9\% of all linked pairs, as positives, and trivia-graded links, the pairs named only in CROSSREFS and random unlinked pairs as negatives (Appendix~\ref{app:links}).
Some random pairs may turn out to be unrecorded relations; we assume this is rare enough to neglect.

We split the data at the release date of the embedding model, so that the classifier is tested on relations that appeared after the model's training data (Fig.~\ref{fig:timeline}). We freeze the cross-reference graph at $S_0$ and call a pair \emph{future} if both entries existed then, neither mentioned the other, and a cross-reference appeared later: 927 such pairs, 265 of them with no word shared between the definitions. Links added after $S_1$ form a window of 89 pairs that postdates all three embedding models. From these future pairs the same frontier model selected 42 in the 2025 set and 48 in the 2026 window, 74 distinct pairs, as substantive relations; in what follows we use them as relations that appeared after the model's knowledge cutoff and after the snapshot date of our training set.

\section{Method}
\label{sec:method}
\lantern{} as a whole is shown in Figure~\ref{fig:pipeline}. As the embedding model we chose Qwen3-32B; for the size comparison in Appendix~\ref{app:scorers} we also embedded the corpus with Qwen3-8B and Qwen3.5-27B.

\subsection{Embedding}
\label{sec:embed}
We embed the definition $d_i$ of each entry, as plain text, with the embedding model. From one forward pass we keep the residual stream after every layer: with $h_t^{(\ell)}(d_i)$ the residual vector at token $t$ after layer $\ell$, $\ell = 0, \ldots, L$, $d$ the width of the model and $T_i$ the number of tokens,
\begin{equation}
x_i^{(\ell)} = \frac{1}{T_i}\sum_{t=1}^{T_i} h_t^{(\ell)}(d_i) \in \mathbb{R}^{d},
\end{equation}
normalised to unit length, $L+1$ vectors per entry; $L = 64$ and $d = 5\,120$ for Qwen3-32B and Qwen3.5-27B, $L = 36$ and $d = 4\,096$ for Qwen3-8B; we keep all layers and let the classifier weight them.

\subsection{Link classifier}
\label{sec:classifier}
A pair of embeddings has far more numbers than labelled pairs (two vectors of $(L+1)\times d$, that is $65\times5\,120$ for Qwen3-32B, against 2\,800 positives), so a classifier on raw embeddings may overfit. We therefore reduce a pair to a few hundred features of two kinds (Fig.~\ref{fig:embed}, drawn for Qwen3-32B). With $x_i^{(\ell)}$ the vectors of Sec.~\ref{sec:embed}, the first kind is the cosine at each of the $L+1$ layers. Some entries have a high cosine with most of the corpus, so we centre the cosine: with $R$ a fixed random sample of 512 corpus entries, $m_i^{(\ell)} = \frac{1}{|R|}\sum_{r\in R}\langle x_i^{(\ell)}, x_r^{(\ell)}\rangle$ the mean cosine of entry $i$ to the sample and $\bar m^{(\ell)}$ its average over the corpus, the feature is
\begin{equation}
c_{ij}^{(\ell)} = \langle x_i^{(\ell)}, x_j^{(\ell)}\rangle - m_i^{(\ell)} - m_j^{(\ell)} + \bar m^{(\ell)}.
\end{equation}
$c_{ij}$ is the similarity of the pair in excess of the two entries' typical similarity to the corpus. The second kind is taken at two layers, $\ell = 4$ and $\ell = 40$ for the two 65-layer models and $\ell = 2$ and $\ell = 22$ for Qwen3-8B, the same relative depths: with $u_i^{(\ell)} = W_\ell\,(x_i^{(\ell)} - \mu_\ell)$ the projection onto the top 64 principal directions of the corpus at that layer, the features are $u_i^{(\ell)} \odot u_j^{(\ell)}$ and $|u_i^{(\ell)} - u_j^{(\ell)}|$, $2 \times 2 \times 64 = 256$ numbers. With the $L+1$ cosines this gives 321 features for Qwen3-32B; on these features, standardised, we train a logistic regression with the labels of Sec.~\ref{sec:labels}.

The 2\,780 substantive links (Sec.~\ref{sec:labels}) are one pair in eighteen thousand of the 50\,million in the corpus, and a classifier trained on that proportion would see almost only negatives. We therefore rebalance the training set: all 2\,780 links as positives, capped at eight per entry and fifty per contributor (the signature on the formula line) so that the most-referenced entries and the most prolific contributors do not dominate, and ten negatives per positive, drawn in equal parts from trivia-graded links, bare cross-references and random unlinked pairs.

\subsection{Hypothesis generation}
\label{sec:generation}
\begin{table}[t]
\centering
\small
\setlength{\tabcolsep}{4pt}
\begin{tabular}{@{}lrr@{}}
\toprule
Stage & Pairs remaining & From previous stage \\
\midrule
Ranked queue & 500 & --- \\
Cheap filter & 118 & 23.6\% \\
Hypothesis proposal & 44 & 37.3\% \\
Technical verification & 44 & 100\% \\
\bottomrule
\end{tabular}
\caption{The staged selection process for main-configuration  run before the content audit. The queue contains the first 500 eligible pairs in the classifier ranking; the final stage requires both agreement with every stored term and an analytical derivation.}
\label{tab:funnel}
\end{table}
From the corpus to verified relations there are four stages: (1) we score every pair of the corpus with the classifier and form a queue from the top of the ranking; (2) then we pass them through a first filter to keep only the candidates worth a closer look; (3) then we run a stronger agent on the pairs the filter kept, and it proposes a hypothesis for each or discards it; (4) finally we verify every surviving hypothesis, first on the finite amount of sequence terms and then analytically.
\begin{itemize}[nosep,leftmargin=*]
\item \textbf{Ranking.} With the classifier of Sec.~\ref{sec:classifier} we score every pair of the corpus. From the ranking we remove the pairs that the encyclopedia already links and the text twins (Sec.~\ref{sec:labels}), and then walk down it keeping a pair only if neither of its entries has appeared in a pair kept above, so that no entry takes part in more than one queued pair.
\item \textbf{Filter.} We then pass the queue through a relatively cheap filter: Claude Sonnet 5 without tools, asked for each pair only whether an exact relation could plausibly exist.
\item \textbf{Hypothesis proposal.} Each pair the filter keeps we hand to a stronger model, Claude Opus 5, with a shell, a local mirror of the encyclopedia ($S_2$) and a small library for exact arithmetic of truncated power series. It receives both definitions, the first terms and the formula lines and may read any entry in full. Its task is to return a derivation whose steps are named facts or checks it ran, a proof class saying how the finite check closes into a proof, and a Python function computing the terms of one sequence from the other, or to discard the pair.
\item \textbf{Verification.} We verify every hypothesis first on the data: the function runs in a sandbox without files, network or third-party libraries, so that it has to compute the terms, and its output is compared with every stored term of the target, 25 to 102 per pair; any mismatch discards the pair, and we also read every passing function, to catch hard-coded terms. Agreement on the stored terms does not prove the relation, so we then verify the derivation of every hypothesis analytically.
\end{itemize}

\subsection{Search configurations}
\label{sec:configs}
\begin{table*}[t]
\centering
\begin{tabular*}{\textwidth}{@{\extracolsep{\fill}}lrrrr@{}}
\toprule
 & \sig{0} routine & \sig{1} informative & \sig{2} insightful & Total \\
\midrule
\nov{0} documented & 22 & 0 & 0 & 22 \\
\nov{1} known elsewhere & 2 & 3 & 2 & 7 \\
\nov{2} not found in search & 0 & 4 & 2 & 6 \\
Novelty not assessed & 27 & 0 & 0 & 27 \\
\midrule
Total & 51 & 7 & 4 & 62 \\
\bottomrule
\end{tabular*}
\caption{Cross-tabulation of novelty and significance for all 62 technically verified relations, before the content screen. \sig{0}--\sig{2} measure what the connection contributes; \nov{0}--\nov{2} measure what was already documented. \nov{2} means not found by the OEIS and targeted literature searches, not established priority. Of the six \nov{2} relations, four pass the screen: two \sig{1} and two \sig{2}.}
\label{tab:ns-matrix}
\end{table*}

The main configuration of Sec.~\ref{sec:classifier} is one of three configurations used to explore the ranked queue. We report them separately because the 62 statements are the union of their outputs, rather than the result of one single run (Table~\ref{tab:configs} in Appendix~\ref{app:configs}). The main configuration processed the top 500 pairs; the two smaller configurations processed their first 100 pairs.

The main configuration yielded the bulk of the verified statements (44), while two variants explored different training setups:
\begin{itemize}[nosep,leftmargin=*]
\item \textbf{Variant 1}: trained to predict the uncurated cross-reference graph, taking all linked pairs in a 5\,000-entry corpus as positives against random unlinked pairs. The first hundred pairs yielded 3 verified statements.
\item \textbf{Variant 2}: ran on the same 5\,000 entries but shifted the objective to substantive relations, using the graded labels of Sec.~\ref{sec:labels} (gold and silver against trivia, bare cross-references, and random pairs). Its first hundred pairs yielded 15 verified statements.
\end{itemize}
The main configuration uses the same graded labels on 10\,000 entries, with the caps of Sec.~\ref{sec:classifier}. The technical pipeline therefore produced 62 verified statements. The full output and the audit ledger are in Appendix~\ref{app:list}.

\subsection{Content screen}
\label{sec:audit}
Technical verification is necessary but not sufficient for the main result: a relation can reproduce every stored term while being mathematically empty. We therefore applied a post-verification content screen to all 62 statements. We retained a relation only when it was exact, specific to the pair, and used substantive structure from both objects. We excluded relations that were primarily a change of notation or parameter, an elementary index or value transformation, a direct consequence of one entry's definition, a generic reconstruction recipe, a finite-value coincidence, or a redundant member of a repeated family. The full screening rule is in Appendix~\ref{app:screen}; the item-level decisions are in Appendix~\ref{app:list}.

The screen is separate from the two-axis assessment applied to the retained relations. The significance (\sig{}) axis measures what the connection contributes, while the novelty (\nov{}) axis measures whether the connection was already documented.
\paragraph{Significance.}
This axis describes what the connection contributes:
\begin{itemize}[nosep,leftmargin=*]
\item \textbf{\sig{0} (routine):} an exact relation that provides a routine transformation or reconstruction, without an informative route or structural interpretation of the kind below.
\item \textbf{\sig{1} (informative):} a specific and useful route for obtaining one object from the other.
\item \textbf{\sig{2} (insightful):} an interpretation, common specification, or structural explanation of what one object is in terms of another.
\end{itemize}
These levels describe the contribution of the connection, not the length of its proof. A relation may pass the content screen while remaining \sig{0}; \sig{1} and \sig{2} identify the retained relations that are informative or insightful.
\paragraph{Novelty.}
This axis describes whether the connection between the two objects was already documented: \textbf{\nov{0}} means documented in the OEIS; \textbf{\nov{1}}, absent from the OEIS but known elsewhere; and \textbf{\nov{2}}, not found in the OEIS or in our targeted literature search. The label concerns the connection between the pair, not the novelty of its proof ingredients; \nov{2} does not establish priority. We mark novelty as not assessed (\novna) for routine cases not subjected to a literature search.

\section{Results}
\label{sec:results}

The main run narrowed the 50\,million possible pairs to 500 ranked candidates, of which 118 passed the inexpensive filter, 44 received a hypothesis from the stronger agent, and all 44 passed sandbox verification and analytical checking (Table~\ref{tab:funnel}). The time of each stage is in Table~\ref{tab:stage-time} (Appendix~\ref{app:configs}). Across the three configurations, the technical pipeline produced 62 verified statements: 44 from the main configuration and 18 from the two variants (Table~\ref{tab:configs}). The content screen then retained 13; among these, nine were informative or insightful (\sig{1} or \sig{2}). This gives the main result as a 62--13--9 funnel; the complete item-level decisions are in the screening ledger (Table~\ref{tab:audit-ledger} in Appendix~\ref{app:list}).

Reproducing the stored terms and supplying a derivation establishes that a relation is correct, but does not by itself make it informative or undocumented. The nine \sig{1}/\sig{2} relations are the retained results that go beyond routine transformations: five give a specific route from one object to the other, while four provide a structural interpretation of the pair. Novelty is assessed separately: \nov{0} means documented in the OEIS, \nov{1} means not found there but known elsewhere, and \nov{2} means not found in the OEIS or in our targeted literature search. The full cross-tabulation is in Table~\ref{tab:ns-matrix}, and the item-level \sig{}/\nov{} labels are listed in Appendix~\ref{app:list}.

\subsection{Significance of the retained relations}
\label{sec:significance}
Table~\ref{tab:showcase} illustrates the range of retained relations, from computational routes between objects to structural interpretations.
\paragraph{Cross-domain bridges.}
Two retained relations connect objects from different mathematical domains. The first links one-dimensional and two-dimensional cellular automata; the second connects analytic $q$-series with 5-core partitions. Both are \nov{2}/\sig{2}: we did not find the connections in the corresponding OEIS records or in our literature search, although some ingredients are known. These are the most discovery-like outputs of the pipeline, although we do not claim that the underlying identities are new theorems.

The \nov{1}/\sig{1} relations recover established mathematics that was absent as a cross-reference between the paired OEIS entries. The two retained \nov{2}/\sig{1} relations supply useful computational routes between the paired objects that we did not find documented in our search. Together with the two \nov{2}/\sig{2} bridges, they make up the four novel connections reported in the abstract: novelty is shared, but the level of significance differs.

\begin{table*}[t]
\centering
\footnotesize
\setlength{\tabcolsep}{4pt}
\renewcommand{\arraystretch}{1.15}
\begin{tabular}{@{}>{\raggedright\arraybackslash}p{0.24\textwidth}>{\raggedright\arraybackslash}p{0.40\textwidth}>{\raggedright\arraybackslash}p{0.31\textwidth}@{}}
\toprule
Objects & Connection in plain language & Exact relation \\
\midrule
\textbf{1D and 2D cellular automata}\newline
\A{071053} $\times$ \A{160239} &
\textbf{\nov{2}/\sig{2} bridge.} The live-cell counts of a planar automaton are recovered from the block factors of a one-dimensional rule (Sec.~\ref{sec:automata-bridge}). &
$g(k)=f(k)^2-f(k-2)^2$;\newline
$A(n)=\prod_i f(k_i)$,\newline
$B(n)=\prod_i g(k_i)$, where $k_i$ are binary run lengths. \\
\addlinespace
\textbf{Continued fraction and theta quotient}\newline
\A{007325} $\times$ \A{227216} &
\textbf{\nov{2}/\sig{2} bridge.} Two analytic expressions combine to count a family of integer partitions, drawn as cell diagrams (Sec.~\ref{sec:partition-bridge}). &
$T(q)^2R(q)^5=\sum_{n\ge0}c_5(n)q^n$, where $c_5(n)$ counts 5-core partitions of $n$. \\
\addlinespace
Greedy sequence and Conway's permutation\newline
\A{003278} $\times$ \A{006368} &
A sequence built by greedily avoiding three-term arithmetic progressions can instead be generated recursively through Conway's permutation $B$. &
$c(0)=0$, $c(2m)=B(2c(m))$,\newline
$c(2m+1)=B(4c(m)+1)$;\newline
$S(n)=c(n-1)+1$. \\
\addlinespace
Pascal matrix and a change of basis\newline
\A{132440} $\times$ \A{039755} &
A change of basis turns the square of the dense lower-triangular Pascal matrix $P$ into a matrix with just two nonzero diagonals. &
$M^{-1}P^2M=I+2T$, where $M$ is the B-Stirling array and $T=\log P$ has only its first subdiagonal nonzero. \\
\addlinespace
Partitions and necklaces\newline
\A{066633} $\times$ \A{185651} &
Deconvolving the counts $P(n,k)$ of parts of size $k$ in partitions of $n$ reveals divisibility. This recovers the totients used to count necklaces. &
$\sum_{i=1}^n e(n-i)P(i,k)=[k\mid n]$,\newline
where $e(m)=[q^m]\prod_{j\ge1}(1-q^j)$. \\
\addlinespace
Planar maps and a Catalan triangle\newline
\A{039598} $\times$ \A{000139} &
Ternary trees connect non-separable planar-map counts $M_n$ to a Catalan triangle $Q$. This recovers a classical connection missing from the OEIS. &
$T_3(x)=\sum_{n\ge0}\frac{n+1}{2}M_nx^n$,\newline
$C(x)=T_3(x/C(x))$,\newline
$Q(n,k)=[x^{n-k}]C(x)^{2k+2}$. \\
\bottomrule
\end{tabular}
\caption{Selected highlights from the 13 screened relations. All six pairs lack an OEIS cross-reference; the first two are the \nov{2}/\sig{2} connections explained in Sec.~\ref{sec:bridges}. The examples span the \nov{}/\sig{} matrix in Table~\ref{tab:ns-matrix} and were selected for interpretability. Full statements and derivations, including the 62 technically verified outputs, are in Appendices~\ref{app:list} and~\ref{app:proofs}. }
\label{tab:showcase}
\end{table*}

\subsection{Two discoveries, explained}
\label{sec:bridges}
To give a concrete sense of the mathematical content of the pipeline's outputs, we examine two retained relations with especially clear mathematical interpretations. For each one, we describe the paired objects, state the exact relation obtained by the search, and give the derivation used to verify it. Both relations were classified as \nov{2} (not found in the OEIS or in our targeted literature search) and \sig{2} (insightful).

\begin{figure*}[t]
\centering
\includegraphics[width=\linewidth]{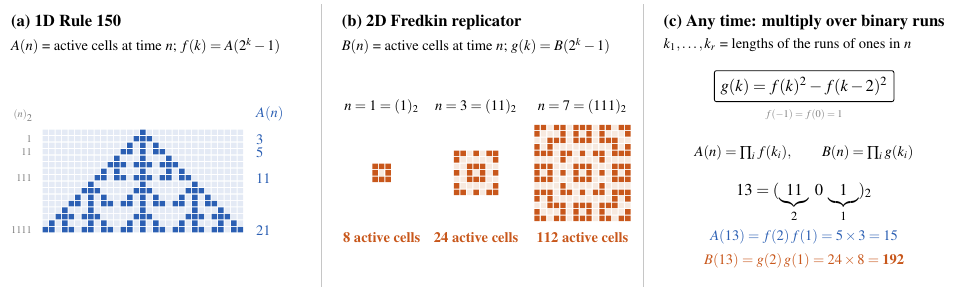}
\caption{\textbf{A 1D rule predicts 2D population counts.} (a)~The space--time diagram of Rule~150, one row per time step; $A(n)$ counts the active cells in row $n$, and $f(k)=A(2^k-1)$ is the count when $n$ is a single run of $k$ ones in binary. (b)~The eight-neighbor Fredkin replicator, grown from a single active cell, at $n=1,3,7$; its single-run counts $g(k)=B(2^k-1)$ are differences of squares of the 1D counts. (c)~For any $n$, the lengths of the runs of ones in its binary expansion select the factors to multiply. At $n=13=(1101)_2$, runs of length two and one give $B(13)=g(2)\,g(1)=24\times8=192$. Colors distinguish the two automata; filled cells are active.}
\label{fig:automata-bridge}
\end{figure*}

\subsubsection{Reconstructing a 2D population sequence from a 1D one}
\label{sec:automata-bridge}
The search found an exact bridge between the population sequences of two cellular automata with different dimensions (Fig.~\ref{fig:automata-bridge}). Starting from one active cell, let $A(n)$ be the number of active cells after $n$ steps of one-dimensional \emph{Rule~150}, and let $B(n)$ be the corresponding number for the two-dimensional \emph{Fredkin replicator}. Although the automata have different local rules and geometries, every value of $B(n)$ can be reconstructed exactly from the one-dimensional sequence $A$.

Rule~150 evolves cells on a line: the next state is the XOR of a cell and its two neighbors. The Fredkin replicator evolves cells on a square grid: a cell is active at the next step when an odd number of its eight neighbors is active. Their population sequences share a factorization over runs of ones in the binary expansion of $n$. Define the single-run counts

\[
f(k)=A(2^k-1),\qquad g(k)=B(2^k-1).
\]
If the run lengths are $k_1,\ldots,k_r$, then
\begin{equation}
A(n)=\prod_{i=1}^r f(k_i),\qquad B(n)=\prod_{i=1}^r g(k_i).
\label{eq:run-factors}
\end{equation}
For example, $13=(1101)_2$ has two runs, of lengths two and one: $A(13)=f(2)\,f(1)$ and $B(13)=g(2)\,g(1)$. The empty product gives $A(0)=B(0)=1$.

\paragraph{The bridge is a difference of squares.}
The relation found by the pipeline is
\begin{equation}
\boxed{g(k)=f(k)^2-f(k-2)^2,}
\label{eq:automata-bridge}
\end{equation}
for $k\ge1$, with $f(0)=1$ and the boundary convention $f(-1)=1$. Thus the 1D counts determine every 2D block factor, and Eq.~\eqref{eq:run-factors} assembles the count at any time. For instance,
\[
\begin{aligned}
g(1)&=3^2-1^2=8,\\
g(2)&=5^2-1^2=24,\\
g(3)&=11^2-3^2=112.
\end{aligned}
\]
At time 13 this gives $A(13)=5\times3=15$ and $B(13)=24\times8=192$. These are exact population counts, not approximations to the pictured patterns.

\paragraph{Why it holds.}
The entries record the block formulas
\[
\begin{aligned}
f(k)&=\frac{2^{k+2}-(-1)^k}{3},\\
g(k)&=\frac{5\cdot4^k-2(-2)^k}{3}.
\end{aligned}
\]
Substituting the first into the difference of squares yields the second:
\[
f(k)^2-f(k-2)^2
=\frac{15\cdot4^k-6(-2)^k}{9}=g(k).
\]
The individual counting laws were already known; the recovered connection expresses the planar law directly through the one-dimensional one. The block identity itself is recorded in OEIS A246030 (Conrad, 2023) in terms of Jacobsthal numbers; what was not recorded is its reading as a bridge between the two automata.

\begin{figure*}[t]
\centering
\includegraphics[width=\linewidth]{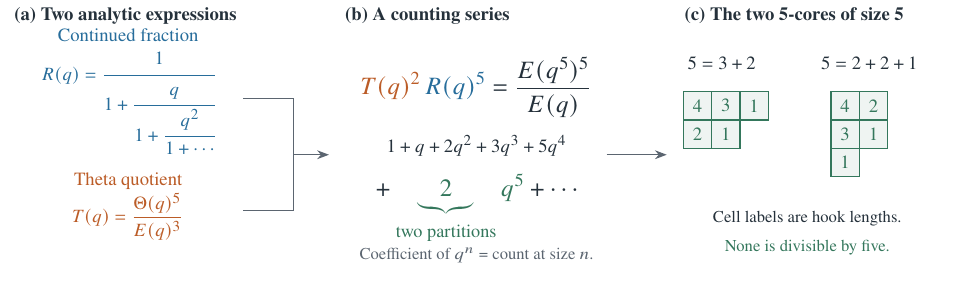}
\caption{\textbf{From analytic expressions to finite diagrams.} The continued fraction $R$ and theta quotient $T$ combine as $T^2R^5$ to give the generating function of 5-core partitions. The coefficient of $q^5$ is two: the diagrams shown are exactly the two qualifying partitions of five. Numbers inside cells are hook lengths, counting the cell itself plus those to its right and below; none is divisible by five.}
\label{fig:partition-bridge}
\end{figure*}

\subsubsection{A continued fraction that counts partitions}
\label{sec:partition-bridge}
A continued fraction and a theta quotient look like analytic objects. Combined in the right powers, they count a family of integer partitions (Fig.~\ref{fig:partition-bridge}). A partition expresses an integer as a sum of positive integers, ignoring order; for example, $5=3+2$ is drawn as left-aligned rows of three and two cells. A cell's \emph{hook length} is one plus the number of cells to its right and below it. A partition is a \emph{5-core} if none of these lengths is divisible by five. Exactly two partitions of five qualify: $(3,2)$ and $(2,2,1)$.

\paragraph{An identity with a counting interpretation.}
Let $R(q)$ be the Rogers--Ramanujan continued fraction in the normalization of \A{007325}, with constant term one:
\[
R(q)=\cfrac{1}{1+\cfrac{q}{1+\cfrac{q^2}{1+\cfrac{q^3}{1+\cdots}}}}.
\]
For the partner \A{227216}, define
\[
\begin{aligned}
E(q)&=\prod_{m\ge1}(1-q^m),\\
\Theta(q)&=\sum_{j\in\mathbb Z}(-1)^j q^{j(5j-1)/2},\\
T(q)&=\Theta(q)^5/E(q)^3.
\end{aligned}
\]
Here $\Theta(q)=1-q^2-q^3+q^9+q^{11}-\cdots$ is a theta series. The bridge reads
\begin{equation}
\boxed{T(q)^2R(q)^5=\sum_{n\ge0}c_5(n)q^n,}
\label{eq:partition-bridge}
\end{equation}
where $c_5(n)$ counts 5-core partitions of $n$. Such a generating function stores the count for size $n$ as the coefficient of $q^n$; multiplying series is convolution of their coefficient arrays. Here the product starts
\[
1+q+2q^2+3q^3+5q^4+\underbrace{2}_{\text{two diagrams}}q^5+\cdots.
\]

\paragraph{Why it holds.}
Group the factors of $E(q)$ by their exponents modulo five. Let $a(q)$ contain the factors with residues 1 and 4, and $b(q)$ those with residues 2 and 3:
\[
\begin{aligned}
a(q)&=\prod_{m\ge0}(1-q^{5m+1})(1-q^{5m+4}),\\
b(q)&=\prod_{m\ge0}(1-q^{5m+2})(1-q^{5m+3}).
\end{aligned}
\]
The remaining factors give $E(q^5)$, so $E(q)=a(q)b(q)E(q^5)$. The classical product representation of the continued fraction gives $R=a/b$, and the Jacobi triple product gives $\Theta=bE(q^5)$. Substituting and cancelling,
\[
\begin{aligned}
T^2R^5
&=\frac{b^{10}E(q^5)^{10}}{E(q)^6}\frac{a^5}{b^5}\\
&=\frac{E(q^5)^5}{E(q)}.
\end{aligned}
\]
The final quotient is the classical generating function of 5-core partitions (\A{053723}). The connection therefore supplies a concrete combinatorial meaning for the product of the two selected analytic expressions.

\section{Controls}
\label{sec:controls}
\lantern{} relies on information encoded in the model's hidden states. Three alternative explanations must therefore be ruled out. The ranking might come from surface features of the definitions, from memorization of OEIS pages, or from the labels used to train the classifier rather than from mathematical knowledge. We test these explanations separately:

\begin{enumerate}[nosep,leftmargin=*]
\item a surface-text scorer tests whether the definitions alone are sufficient;
\item a post-cutoff test tests whether the classifier needs to have memorized the relevant pages;
\item a label-free question to the model tests whether the signal is present without our training labels.
\end{enumerate}

The controls use held-out cross-references and matched negative examples whenever possible.

\begin{figure}[t]
\centering
\includegraphics[width=\linewidth]{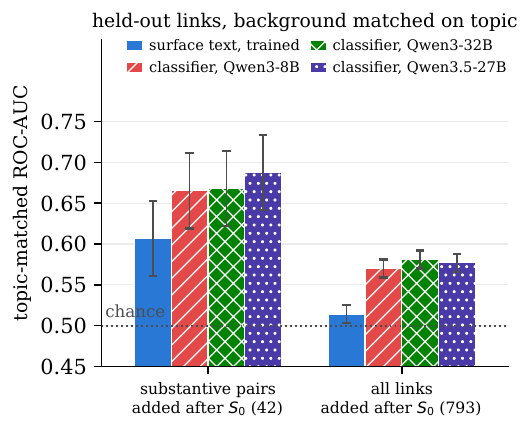}
\caption{Cross-references added after $S_0$: ROC-AUC of the surface-text scorer and of the classifier on three model sizes, all trained on the same labels, against unlinked pairs whose entries carry the same two topics as the linked pair (Sec.~\ref{sec:heldout}); the 42 substantive pairs and the 793 links with a matched background; whiskers are $\pm$1 SE. Qwen3.5-27B was released after $S_0$ and may have seen the 42 pairs. }
\label{fig:controls}
\end{figure}

\subsection{Activations against the text of the definitions}
\label{sec:textscorer}
\label{sec:heldout}

\paragraph{Question.}
Could the same ranking be obtained from the definitions alone? If so, the activations would add no evidence about latent mathematical knowledge.

\paragraph{Control.}
We built a surface-text scorer with the same pairwise classifier as in Sec.~\ref{sec:classifier}. Each definition was represented by character 3--5-gram TF--IDF features, reduced to 128 LSA dimensions. For each pair we used the elementwise product, absolute difference, and word-set overlap. The activation-based and text-based scorers therefore differed only in their input representations and were trained on the same labels and negatives.

\paragraph{Evaluation.}
We evaluated both scorers on 927 cross-references added after $S_0$, including 42 pairs graded as substantive. Random negatives are unsuitable here because linked entries tend to share a topic. We therefore matched each held-out pair against unlinked pairs with the same two topics, excluding text twins.
This gives the scorer credit only for distinguishing related entries from unrelated entries within the same topical background.

\paragraph{Result.}
On the 793 held-out links with a valid matched background, the text scorer was at chance (AUC $=0.51$), whereas the activation-based classifiers reached AUC $=0.57$--$0.58$ across all three model sizes. On the 42 substantive pairs, the corresponding values were $0.61$ and $0.67$--$0.69$. Thus the definitions reveal the topic of an entry, but the activations provide additional information about which entries within that topic are related.

\subsection{Is it memory of the pages?}
\label{sec:memory}
The previous control does not exclude page memorization. A model may have seen the two OEIS pages during pretraining and may use information about those pages rather than properties of the sequences. We test this in two ways.

\paragraph{Edits after the training snapshot.}
A definition edited after $S_0$ could contain a hint about a cross-reference added later. Only 30 of the 927 held-out links involve such an edit. Removing them leaves the results of Fig.~\ref{fig:controls} unchanged.

\paragraph{Entries created after pretraining.}
We next used 2\,379 entries created in 2026, after the model's knowledge cutoff. Their 5\,557 cross-references to older entries were scored by a classifier trained only on cross-references among older entries. The classifier reached AUC $=0.88$ on the new entries, compared with $0.89$ on a random held-out split among older entries. The corresponding text scorer dropped from $0.85$ to $0.78$.

The classifier therefore does not need to have seen an entry's page in order to rank its links. This result is consistent with the interpretation that the signal is transferred from the model's representations of the mathematical definitions, rather than retrieved from memorized pages.

\subsection{Is it in the model, or in our labels?}
\label{sec:question}
The two classifiers were trained on the 2\,780 substantive links selected in Sec.~\ref{sec:labels}. Their performance could therefore reflect our labelling decisions rather than mathematical information in the model. We address this concern with a label-free readout.

We showed Qwen3-32B the two definitions and asked it to rate the depth of their relation on a 0--9 scale, using the same anchors as the grading rubric in Sec.~\ref{sec:labels}. The score was the expected first-token digit, averaged over both orders of the pair. Unlike the classifier, this readout received the two definitions jointly and used no training labels.

Because a direct question is too expensive to run on all 50\,million pairs, we evaluated it on the 74 substantive future pairs from Sec.~\ref{sec:labels}. Each pair was ranked against 100 unlinked pairs from the same popularity bins. The question and the classifier achieved nearly identical median percentiles: 88.5 and 88.0, respectively.

The label-free question therefore recovers a signal comparable to that of the trained classifier.  The practical advantage of the classifier is throughput: it converts one embedding pass over the corpus into scores for all pairs.

\section{Limitations}
\label{sec:limitations}

Our results have two main limitations. First, the OEIS is a particularly structured test bed: its objects come with human-written definitions, formulas, initial terms, and explicit cross-references, while our corpus is biased toward frequently mentioned sequences. It is therefore unclear whether the same approach would work for less formal mathematical objects or for knowledge bases in which relations are not recorded as explicit links. We also evaluated only one family of pretrained models.

\section{Conclusion}
\label{sec:conclusion}

We showed that a pretrained language model can help choose mathematical questions, not only answer those posed by a researcher. By ranking millions of pairs of OEIS entries with hidden-state representations and verifying the most promising candidates with an agent, \lantern{} produced 62 technically verified relations. A content screen retained 13, of which nine are informative or insightful. Four of these nine were not found in the OEIS or in our targeted literature search: two provide informative computational routes and two provide insightful structural interpretations. These results suggest that pretrained representations can serve as a practical guide for discovering and investigating unrecorded mathematical relations in large structured knowledge bases.

\bibliography{refs}

\clearpage
\onecolumn
\appendix
\section{The 62 statements}
\label{app:list}
Each item gives the pair, its novelty/significance assessment, and the statement; the screening ledger below gives its independent decision and reason. The derivations are in Appendix~\ref{app:proofs}, and the verification code is in the repository. The list contains all 62 technically verified outputs; the main text reports the 13 relations retained by the content screen. For the 20 statements marked $\dagger$ the screen read the entries' formula and comment lines alone (Sec.~\ref{sec:limitations}).
\begin{longtable}{@{}p{0.035\textwidth}p{0.25\textwidth}p{0.10\textwidth}p{0.10\textwidth}p{0.43\textwidth}@{}}
\toprule
No. & Pair & \nov{}/\sig{} & Decision & Reason \\
\midrule
\endhead
1 & \A{230877}$\times$\A{334302} & \nov{2}/\sig{1} & excluded & The ordering is the definition of A334302; the substantive statistic is a relation to the A125106 encoding. \\
2 & \A{053120}$\times$\A{060187} & \nov{2}/\sig{1} & excluded & Chebyshev coefficients only produce the odd powers used in the recorded formula for A060187. \\
3 & \A{103438}$\times$\A{001498} & \nov{1}/\sig{1} & retained & Specific route through central factorial numbers and the Bessel coefficients. \\
4 & \A{124733}$\times$\A{266213} & \nov{2}/\sig{1} & retained & Specific Riordan-array composition and closed-form identity. \\
5 & \A{000127}$\times$\A{076839} & \novna/\sig{0} & excluded & Any two period-five sequences can be matched by a five-value table. \\
6 & \A{132440}$\times$\A{039755} & \nov{2}/\sig{1} & retained & Specific conjugation identity in the exponential Riordan group. \\
7 & \A{008282}$\times$\A{162975} & \novna/\sig{0} & retained & A useful route through Euler numbers between two independently recorded triangles. \\
8 & \A{004009}$\times$\A{122135} & \nov{1}/\sig{0} & retained & A useful exact route from the partner's Rogers--Ramanujan functions to $E_4$. \\
9 & \A{013973}$\times$\A{093829} & \nov{1}/\sig{0} & retained & A useful exact route from the partner's cubic theta function to $E_6$. \\
10 & \A{007325}$\times$\A{227216} & \nov{2}/\sig{2} & retained & Structural interpretation as the generating function of 5-core partitions. \\
11 & \A{004011}$\times$\A{000003} & \nov{1}/\sig{1} & retained & Specific theta-series and class-number route. \\
12 & \A{039598}$\times$\A{000139} & \nov{1}/\sig{2} & retained & Structural bridge from planar maps to ternary trees and Catalan numbers. \\
13 & \A{122856}$\times$\A{303363} & \novna/\sig{0} & excluded & Deconvolution and arithmetic progression are definitional; the partner adds no route. \\
14 & \A{112555}$\times$\A{132440} & \nov{0}/\sig{0} & excluded & Recorded generating-function expansion; A132440 only names the Pascal logarithm. \\
15 & \A{005043}$\times$\A{021009} & \nov{0}/\sig{0} & excluded & Recorded inverse binomial transform and signed Pascal row. \\
16 & \A{034807}$\times$\A{063967} & \nov{0}/\sig{0} & excluded & Reindexing and removal of a normalisation in a recorded formula. \\
17 & \A{003278}$\times$\A{006368} & \novna/\sig{0} & excluded & Conway's defining branches reduce the claim to the definition of A003278. \\
18 & \A{130534}$\times$\A{297328} & \novna/\sig{0} & excluded & Generic Newton-series substitution into a column definition. \\
19 & \A{049310}$\times$\A{046534} & \novna/\sig{0} & excluded & A046534 only recovers Pascal's triangle for a recorded formula. \\
20 & \A{007429}$\times$\A{007437} & \novna/\sig{0} & excluded & Algebra of two recorded Dirichlet series. \\
21 & \A{047749}$\times$\A{008315} & \novna/\sig{0} & excluded & Ratio of recorded closed forms. \\
22 & \A{039599}$\times$\A{294175} & \novna/\sig{0} & excluded & Recorded formula rewritten by Pascal's identity. \\
23 & \A{035513}$\times$\A{101864} & \nov{1}/\sig{1} & retained & The BB-numbers give a specific route to the Wythoff array. \\
24 & \A{009766}$\times$\A{060693} & \novna/\sig{0} & excluded & Row reversal and a recorded binomial factor. \\
25 & \A{214262}$\times$\A{081360} & \novna/\sig{0} & excluded & Generic telescoping of eta quotients. \\
26 & \A{002288}$\times$\A{109389} & \novna/\sig{0} & excluded & Generic Euler-product regrouping. \\
27 & \A{079006}$\times$\A{002513} & \novna/\sig{0} & excluded & Generic eta-product calculus. \\
28 & \A{087808}$\times$\A{227736} & \novna/\sig{0} & excluded & Unrolling the defining binary-run recursion. \\
29 & \A{002654}$\times$\A{227454} & \novna/\sig{0} & excluded & Standard square-root and eta identities. \\
30 & \A{006318}$\times$\A{000669} & \nov{1}/\sig{2} & retained & Structural bridge between multiset and sequence tree grammars. \\
31 & \A{081362}$\times$\A{080054} & \novna/\sig{0} & excluded & Change of parameter $x\mapsto-x$. \\
32 & \A{001935}$\times$\A{050468} & \novna/\sig{0} & excluded & Generic eta-product calculus. \\
33 & \A{000108}$\times$\A{005802} & \nov{0}/\sig{0} & excluded & Recorded Gessel formula with a Catalan rewrite. \\
34 & \A{004018}$\times$\A{227216} & \novna/\sig{0} & excluded & Quotient of recorded Euler transforms. \\
35 & \A{008284}$\times$\A{240009} & \novna/\sig{0} & excluded & Parameter substitution in a recorded bivariate generating function. \\
36 & \A{027641}$\times$\A{259072} & \novna/\sig{0} & excluded & A constant rewritten through one Bernoulli number. \\
37 & \A{006352}$\times$\A{121361} & \novna/\sig{0} & excluded & Generic logarithmic derivative of an eta product. \\
38 & \A{000118}$\times$\A{122856} & \novna/\sig{0} & excluded & The full $r_2$ sequence is reconstructed before taking its convolution square. \\
39 & \A{130534}$\times$\A{144150} & \nov{0}/\sig{0} & excluded & Inverse of a recorded Stirling transform. \\
40 & \A{007332}$\times$\A{001936} & \novna/\sig{0} & excluded & Generic eta-product substitution. \\
41 & \A{352361}$\times$\A{193842} & \novna/\sig{0} & excluded & Universal Lucas-polynomial expansion. \\
42 & \A{000032}$\times$\A{295862} & \novna/\sig{0} & excluded & The recorded formula has Lucas numbers as its homogeneous part. \\
43 & \A{227543}$\times$\A{251592} & \novna/\sig{0} & excluded & Two common Catalan deformations meet at one parameter value. \\
44 & \A{071053}$\times$\A{160239} & \nov{2}/\sig{2} & retained & Structural bridge between one- and two-dimensional cellular automata. \\
45 & \A{066633}$\times$\A{185651} & \nov{0}/\sig{0} & retained & Useful route from partition counts through divisibility and totients to necklaces. \\
46 & \A{003557}$\times$\A{065483} & \nov{0}/\sig{0} & excluded & Generic Euler-product reconstruction from a multiplicative function. \\
47 & \A{101036}$\times$\A{001917} & \nov{0}/\sig{0} & excluded & Covering-set definition restated through a table of orders. \\
48 & \A{101211}$\times$\A{245562} & \nov{0}/\sig{0} & excluded & Odd positions of the run list. \\
49 & \A{101211}$\times$\A{245563} & \nov{0}/\sig{0} & excluded & Odd positions of the run list in reverse order. \\
50 & \A{010815}$\times$\A{001399} & \nov{0}/\sig{0} & excluded & First three factors of a recorded Euler product. \\
51 & \A{007325}$\times$\A{354650} & \nov{0}/\sig{0} & excluded & Definition of the inverse series plus a recorded comment. \\
52 & \A{302997}$\times$\A{286354} & \novna/\sig{0} & excluded & Reconstruction of $\theta_3$ and the Euler product. \\
53 & \A{056911}$\times$\A{002129} & \nov{0}/\sig{0} & excluded & Primality test followed by a generic sieve. \\
54 & \A{002313}$\times$\A{259149} & \nov{0}/\sig{0} & excluded & Prime list used only to compute $r_2$ and $\theta_3$. \\
55 & \A{013661}$\times$\A{051716} & \nov{0}/\sig{0} & excluded & Standard Bernoulli and zeta formulas. \\
56 & \A{172236}$\times$\A{168561} & \nov{0}/\sig{0} & excluded & Two descriptions of the same Fibonacci-polynomial object. \\
57 & \A{030203}$\times$\A{001817} & \nov{0}/\sig{0} & excluded & Recorded transforms with an index shift. \\
58 & \A{000594}$\times$\A{002445} & \nov{0}/\sig{0} & excluded & Coincidence of two numerical constants. \\
59 & \A{065330}$\times$\A{002175} & \nov{0}/\sig{0} & excluded & The first entry contributes only an equivalent set of coprime integers. \\
60 & \A{048675}$\times$\A{341783} & \nov{0}/\sig{0} & excluded & The partner serves only as a list of primes. \\
61 & \A{046523}$\times$\A{069279} & \nov{0}/\sig{0} & excluded & Characterisation following directly from the definitions. \\
62 & \A{000203}$\times$\A{006954} & \nov{0}/\sig{0} & excluded & Primality test inserted into a standard Bernoulli formula. \\
\bottomrule
\caption{Screening ledger for the 62 statements listed below. The screen retains 13 relations that give a useful exact route or interpretation for a reader; nine are \sig{1} or \sig{2}. The remaining 49 are primarily definitional, representational, generic, or mechanically reproducible.}
\label{tab:audit-ledger}
\end{longtable}
\small
\begin{enumerate}[nosep,leftmargin=*]
\item \A{230877} $\times$ \A{334302} (\nov{2}, \sig{1}). With $\lambda(n)$ the partition encoded by $n$ as in \A{125106} and $k$ its number of parts, $w(n)=k\max\lambda(n)+\frac{k(k+1)}{2}-|\lambda(n)|$; the partitions $\lambda(1),\lambda(2),\ldots$ ordered by sum, length and reverse lexicographic order are the rows of \A{334302}.
\item \A{053120} $\times$ \A{060187} (\nov{2}, \sig{1}). $e^{(2i-1)u}=T_{2i-1}(\cosh u)+\sinh u\,\bigl(T_0(\cosh u)+2\sum_{j=1}^{i-1}T_{2j}(\cosh u)\bigr)$; row $n$ of the type-B Eulerian numbers is the coefficients of $x^1,\ldots,x^n$ in $(1-x)^n\sum_{i=1}^n(2i-1)^{n-1}x^i$, with $(2i-1)^{n-1}=(n-1)!\,[u^{n-1}]$ of the first series.
\item \A{103438} $\times$ \A{001498} (\nov{1}, \sig{1}). With the central factorial numbers $\mathrm{CF}$ (\A{036969}), $T(2p+1,n)=\sum_{k=1}^{p+1}(2k-2)!!\,\mathrm{CF}(p+1,k)\,B(n,k)$ and $T(2p,n)=(2n+1)\sum_{k=1}^{p}\frac{k}{2k+1}(2k-2)!!\,\mathrm{CF}(p,k)\,B(n,k)$.
\item \A{124733} $\times$ \A{266213} (\nov{2}, \sig{1}). With $h$ the Riordan function of $W$ ($h=x(1+3h+h^2)$), $\sum_{k=0}^n N(m,k)\bigl(W(n-1,k-1)+W(n-1,k)\bigr)=[x^n]\bigl(\frac{1-x}{1-5x}\bigr)^{m/2}$.
\item \A{000127} $\times$ \A{076839} (\novna, \sig{0}). $y(n)=1+\lfloor (r(n+4)\bmod 5)/2\rfloor$.
\item \A{132440} $\times$ \A{039755} (\nov{2}, \sig{1}). $M^{-1}P^2M=I+2T$, with $P$ the Pascal matrix.
\item \A{008282} $\times$ \A{162975} (\novna, \sig{0}). Both triangles reduce to the generating function $e^x\sec x$; from it the Euler numbers, and from them by Deutsch's formula the whole first triangle.
\item \A{004009} $\times$ \A{122135} (\nov{1}, \sig{0}). $E_4=(x;x)_\infty^8\,(G^{20}+228xG^{15}H^5+494x^2G^{10}H^{10}-228x^3G^5H^{15}+x^4H^{20})$, $G$ and $H$ the Rogers--Ramanujan series, recovered from the partner.
\item \A{013973} $\times$ \A{093829} (\nov{1}, \sig{0}). $E_6=-27a^6+36a^3b^3-8b^6$, $a$ and $b$ the Borwein cubic theta functions; $a$ telescopes from the partner.
\item \A{007325} $\times$ \A{227216} (\nov{2}, \sig{2}). $T(q)^2R(q)^5=(q^5;q^5)_\infty^5/(q;q)_\infty$, the generating function of 5-core partitions.
\item \A{004011} $\times$ \A{000003} (\nov{1}, \sig{1}). $N(n)=\varepsilon(n)\sum_{s\in\mathbb Z} r_3(n-s^2)$, $\varepsilon(n)=3$ for odd $n$ and $1$ otherwise, with $r_3$ given by Gauss's class-number formula through $h(-4m)$.
\item \A{039598} $\times$ \A{000139} (\nov{1}, \sig{2}). With $T_3(x)=\sum_n \frac{n+1}{2}M(n)x^n$, $T_3=1+xT_3^3$, $C(x)=T_3(x/C(x))$ and $Q(n,k)=[x^{n-k}]\,C(x)^{2k+2}$: the maps are counted by ternary trees up to the factor $(n+1)/2$, followed by the Fuss--Catalan substitution.
\item \A{122856} $\times$ \A{303363} (\novna, \sig{0}). Deconvolving $N$ by the set $\{2^c+2^d: c \text{ even}, c\le d\}$ gives $U(m)$, the number of unordered representations of $m$ as a sum of two triangular numbers, and $t(n)=2U(m)-[m \text{ pronic}]$ with $4m+1$ the odd part of $3n+2$; $t(n)=0$ when that odd part is $3 \bmod 4$.
\item \A{112555} $\times$ \A{132440} (\nov{0}, \sig{0}). With $P=\exp(B)$ the Pascal matrix, $T(n,k)=[n=k]+\sum_{j=0}^{n-1}(-1)^jP(j,n-1-k)$.
\item \A{005043} $\times$ \A{021009} (\nov{0}, \sig{0}). $\sum_{k=0}^n L(n,k)\frac{(2k)!}{(k+1)!}=(-1)^n n!\,R(n)$.
\item \A{034807} $\times$ \A{063967} (\nov{0}, \sig{0}). $U(n,k)=\sum_{i=0}^{\lfloor n/2\rfloor}\frac{n-i}{n}L(n,i)\binom{n-i}{k}$.
\item \A{003278} $\times$ \A{006368} (\novna, \sig{0}). $c(0)=0$, $c(2m)=B(2c(m))$, $c(2m+1)=B(4c(m)+1)$ gives $S(n)=c(n-1)+1$ for all $n\ge1$.
\item \A{130534} $\times$ \A{297328} (\novna, \sig{0}). With $P_m(k)=\sum_{i=0}^{m-1}s(m-1,i)k^{i+1}=k(k+1)\cdots(k+m-1)$, $Q(n,k)=[x^n]\prod_{j\ge1}\sum_{m\ge0}\frac{P_m(k)}{m!}j^m x^{jm}$.
\item \A{049310} $\times$ \A{046534} (\novna, \sig{0}). Reading the entry as a rational triangle $P(n,k)$, $3\binom nk=3P(n,k)+5\binom{n-2}{k-1}$ recovers Pascal's triangle, and then $S(n,k)=(-1)^{(n-k)/2}\binom{(n+k)/2}{k}$ for $n\ge k$ with $n-k$ even, else $0$.
\item \A{007429} $\times$ \A{007437} (\novna, \sig{0}). $f(n)+\sum_{d\mid n}d\,\varphi(d)f(n/d)=2\sum_{d\mid n}g(d)$.
\item \A{047749} $\times$ \A{008315} (\novna, \sig{0}). $a(2m)=T(3m,m)/(m+1)$, $a(2m+1)=2T(3m+1,m)/(m+2)$.
\item \A{039599} $\times$ \A{294175} (\novna, \sig{0}). The row sums of the triangle are $\binom{2n}{n}$, and $a(n)=\bigl(2^n-\binom{2\lceil n/2\rceil}{\lceil n/2\rceil}\bigr)/2$.
\item \A{035513} $\times$ \A{101864} (\nov{1}, \sig{1}). The BB-numbers are the third column of the array plus 2; from them $\lfloor n\varphi\rfloor$ and the whole array are recovered.
\item \A{009766} $\times$ \A{060693} (\novna, \sig{0}). The rows of one are the reversed rows of the other multiplied by $\binom{n+2}{k+1}/(n+2)$, for all $0\le k\le n$.
\item \A{214262} $\times$ \A{081360} (\novna, \sig{0}). The first is an infinite product of values of the second at $x^{m\cdot2^k}$; the multiplicities telescope and the factorisation is unique.
\item \A{002288} $\times$ \A{109389} (\novna, \sig{0}). Every $n$ is uniquely a 3-smooth number times a number coprime to 6, so the full Euler product is assembled from values of the partner at $q^k$.
\item \A{079006} $\times$ \A{002513} (\novna, \sig{0}). The first equals $(B(-x)/B(x^2))^2$, $B$ the generating function of cubic partitions.
\item \A{087808} $\times$ \A{227736} (\novna, \sig{0}). Unrolling the recursion is a pass over the reversed row of runs: a run of $r$ ones adds $r$, a run of $r$ zeros multiplies by $2^r$.
\item \A{002654} $\times$ \A{227454} (\novna, \sig{0}). From the first, $\theta_3^2=1+4\sum$ is assembled; a square root gives $\theta_3$, and from it the Euler product and the eta quotient.
\item \A{006318} $\times$ \A{000669} (\nov{1}, \sig{2}). Replacing ``multiset of subtrees'' by ``sequence of subtrees'' turns the tree grammar into a quadratic equation whose solution is the generating function of the large Schr\"oder numbers.
\item \A{081362} $\times$ \A{080054} (\novna, \sig{0})$^\dagger$. With $A(x)$ the generating function of \A{081362}, that of \A{080054} is $A(-x)/A(x)=A(x^2)/A(x)^2$.
\item \A{001935} $\times$ \A{050468} (\novna, \sig{0})$^\dagger$. With $U(q)$ the generating function of \A{001935} and $E(q)=\prod_{j\ge0}U(q^{4^j})^{-1}$, $\sum_{n\ge1}\A{050468}(n)q^n=qU(q)^4E(q^2)^2\bigl(E(q)^8+20qE(q^4)^8\bigr)$.
\item \A{000108} $\times$ \A{005802} (\nov{0}, \sig{0})$^\dagger$. $(n+1)^2\,\A{005802}(n)=\sum_{k=0}^n C_k\binom{n+1}{k}\binom{n+1}{k+1}$.
\item \A{004018} $\times$ \A{227216} (\novna, \sig{0})$^\dagger$. The generating function of \A{004018} is that of \A{227216} times $\prod_{k\ge1}(1-x^k)^{c_k}$ with $c_k$ of period 20, $c=[-1,4,-6,5,-6,9,-6,0,-1,4,-1,0,-6,9,-6,5,-6,4,-1,0]$.
\item \A{008284} $\times$ \A{240009} (\novna, \sig{0})$^\dagger$. With $P(t,x)=\prod_{j\ge1}(1-tx^j)^{-1}=\sum_{s,j}\A{008284}(s,j)t^jx^s$, $\sum_{n,k}\A{240009}(n,k)u^kq^n=P(u/q,q^2)\,P(1/u,q^2)$.
\item \A{027641} $\times$ \A{259072} (\novna, \sig{0})$^\dagger$. $-\zeta'(-7)=\frac{B_8}{8}\bigl(\log 2\pi+\gamma-H_7-\zeta'(8)/\zeta(8)\bigr)$ with $B_8=\A{027641}(8)/30$ and $\zeta(8)=-B_8(2\pi)^8/(2\cdot8!)$.
\item \A{006352} $\times$ \A{121361} (\novna, \sig{0})$^\dagger$. With $g(m)=\bigl(-a(m)+2a(m/2)+3a(m/3)+4a(m/4)-6a(m/6)+12a(m/12)\bigr)/24$, $a=\A{006352}$, the sequence $b=\A{121361}$ satisfies $nb(n)=\sum_{m=1}^n g(m)b(n-m)$.
\item \A{000118} $\times$ \A{122856} (\novna, \sig{0})$^\dagger$. $r_2(k)$ is read off \A{122856} ($r_2(k)=0$ when $v_3(k)$ is odd, else $4\,\A{122856}((w-2)/3)$ with $w$ determined by $k/3^{v_3(k)} \bmod 3$), and $\A{000118}(n)=\sum_{i+j=n}r_2(i)r_2(j)$.
\item \A{130534} $\times$ \A{144150} (\nov{0}, \sig{0})$^\dagger$. $A(n,k)=\sum_{m=1}^n(-1)^{n-m}T(n-1,m-1)A(m,k+1)$, the inverse of the recorded Stirling transform between columns.
\item \A{007332} $\times$ \A{001936} (\novna, \sig{0})$^\dagger$. $A(x^4)=x^3A(x)\bigl(B(x)B(x^3)\bigr)^3$ for the two generating functions.
\item \A{352361} $\times$ \A{193842} (\novna, \sig{0})$^\dagger$. Row $n$ of \A{193842} is $U_{n+1}(4x+1,x+3x^2)$, whose expansion in the basis $(4x+1)^{n-2j}(-x-3x^2)^j$ has the Fibonacci-polynomial coefficients $\binom{n-j}{j}$.
\item \A{000032} $\times$ \A{295862} (\novna, \sig{0})$^\dagger$. $L(n+1)=\A{295862}(n)-\sum_{i=2}^nF(n+1-i)\,b(i)$, $b$ the increasing complement of \A{295862}.
\item \A{227543} $\times$ \A{251592} (\novna, \sig{0})$^\dagger$. With $T$ the rows of \A{227543} and $U$ the rows of \A{251592}, $\sum_k T(n,k)=\frac{1}{n!}\sum_j U(n,j)\,2^{j}=C_n$ for every $n\ge1$: the two triangles are one-parameter deformations of the same Catalan functional equation, and they meet at $q=1$, $t=2$.
\item \A{071053} $\times$ \A{160239} (\nov{2}, \sig{2}). For both cellular automata the number of live cells at step $n$ factors over the blocks of ones in the binary expansion of $n$; the block factors satisfy $g(k)=f(k)^2-f(k-2)^2$ with $f(k)=J_{k+2}$, the Jacobsthal numbers.
\item \A{066633} $\times$ \A{185651} (\nov{0}, \sig{0}). With $e(m)$ the coefficients of $\prod_{k\ge1}(1-x^k)$ (\A{010815}), $\sum_{i=1}^n e(n-i)P(i,k)=[k\mid n]$; from this divisibility matrix $\varphi$ and the whole necklace array $N$ are recovered.
\item \A{003557} $\times$ \A{065483} (\nov{0}, \sig{0}). $C=\sum_{n\ge1}\kappa(n)^2/n^3=\sum_{n\ge1}1/(n\,\mathrm{rad}(n)^2)$.
\item \A{101036} $\times$ \A{001917} (\nov{0}, \sig{0}). $\mathrm{ord}_2(\mathrm{prime}(i+2))=(\mathrm{prime}(i+2)-1)/b(i)$; the Riesel numbers with a covering set are exactly the odd $n$ for which there is a period $L$ such that for every $k=1,\ldots,L$ some prime $p$ with $\mathrm{ord}_2(p)\mid L$ has $n\equiv2^{-k}\pmod p$.
\item \A{101211} $\times$ \A{245562} (\nov{0}, \sig{0}). Row $n$ of \A{245562} is the odd positions of row $n$ of \A{101211}.
\item \A{101211} $\times$ \A{245563} (\nov{0}, \sig{0}). The same odd positions read backwards.
\item \A{010815} $\times$ \A{001399} (\nov{0}, \sig{0}). The generating function of partitions into at most three parts is $1/((1-x)(1-x^2)(1-x^3))$, the first three factors of the Euler product; the rest is the tail $\prod_{k\ge4}(1-x^k)$.
\item \A{007325} $\times$ \A{354650} (\nov{0}, \sig{0}). $R(q)=f(-q,-q^4)/f(-q^2,-q^3)$, a quotient of two values of the inverted series at parameters giving $q^4$ and $q^3$.
\item \A{302997} $\times$ \A{286354} (\novna, \sig{0}). $\theta_3(x)=E(x^2)^5/(E(x)^2E(x^4)^2)$ with $E$ the Euler product; $\theta_3$ is recovered termwise from the first array, and $E$ and all its powers from $\theta_3$.
\item \A{056911} $\times$ \A{002129} (\nov{0}, \sig{0}). For odd $n$ the second sequence equals $\sigma(n)$; $\sigma(n)=n+1$ singles out the primes, and a sieve by their squares gives the first.
\item \A{002313} $\times$ \A{259149} (\nov{0}, \sig{0}). The constant is assembled from $\eta(i)$ and $\theta_3(e^{-\pi})$, and $\theta_3^2=\sum r_2(n)q^n$ is computed from the list of primes of the form $x^2+y^2$.
\item \A{013661} $\times$ \A{051716} (\nov{0}, \sig{0}). By von Staudt--Clausen the $B_{2m}$ are recovered from the numerators, and Euler's $\zeta(2m)=(2\pi)^{2m}|B_{2m}|/(2(2m)!)$ gives a power of $\zeta(2)$, from which the integer root is taken.
\item \A{172236} $\times$ \A{168561} (\nov{0}, \sig{0})$^\dagger$. The array entry $A(n,k)$ is the $k$-th Fibonacci polynomial evaluated at $n$, whose coefficients are row $k-1$ of \A{168561}: $T(n,k)=\sum_{j=0}^{k-1}B(k-1,j)(n-k)^j$ for the antidiagonal triangle.
\item \A{030203} $\times$ \A{001817} (\nov{0}, \sig{0})$^\dagger$. With $e$ the M\"obius transform of \A{001817}, the generating function of \A{030203} is $\prod_{k\ge1}(1-x^k)^{1+e(k+1)}$.
\item \A{000594} $\times$ \A{002445} (\nov{0}, \sig{0})$^\dagger$. With $b(n)$ the denominator of $B_{2n}$, $E_4=1+8b(2)\sum\sigma_3(m)q^m$, $E_6=1-12b(3)\sum\sigma_5(m)q^m$ and $\sum\tau(n)q^n=(E_4^3-E_6^2)/(24(b(2)+b(3)))$.
\item \A{065330} $\times$ \A{002175} (\nov{0}, \sig{0})$^\dagger$. $\A{002175}(n)$ is the number of ordered pairs $(x,y)$ of values of \A{065330} with $x^2+y^2=24n+2$.
\item \A{048675} $\times$ \A{341783} (\nov{0}, \sig{0})$^\dagger$. $\nu(x)=\sqrt x$ for square $x$ and $x$ otherwise maps \A{341783} bijectively onto the primes, so $\pi(p)$ and hence $\A{048675}(n)=\sum_{p^e\parallel n}e\,2^{\pi(p)-1}$ are determined by \A{341783}.
\item \A{046523} $\times$ \A{069279} (\nov{0}, \sig{0})$^\dagger$. \A{069279} is exactly the set of $n$ with $\Omega(\A{046523}(n))=18$.
\item \A{000203} $\times$ \A{006954} (\nov{0}, \sig{0})$^\dagger$. For $n\ge2$, $\A{006954}(n)=\prod_{d\mid 2n-2,\ \sigma(d+1)=d+2}(d+1)$.
\end{enumerate}
\normalsize
\section{Content-screening criteria}
\label{app:screen}
The content screen removes technically correct relations whose mathematical content is too weak for the main result. We retain a relation only if it is exact, specific to the pair, and gives a useful mathematical route or interpretation for a reader. We exclude a relation when it is primarily one of the following.

\paragraph{1. Another representation of the same object.}
We exclude a relation if one object is only:
\begin{itemize}[nosep,leftmargin=*]
\item a relabelling of the other;
\item a binary, signed, decimal, or other recoding;
\item a generating function, coefficient sequence, partial-sum array, or other standard representation of the same object;
\item the same object under a different normalisation or parametrisation.
\end{itemize}
This excludes a relation only when the replacement removes no useful route for the reader; a canonical representation of a classical object may still be retained when the relation gives a useful way to obtain or interpret the paired object.

\paragraph{2. Elementary transformation of indices or values.}
We exclude relations that are primarily an index shift, scaling, multiplication by a simple factor, addition of a constant, reversal, selection of even or odd positions, extraction or obvious union of subsequences, or a simple substitution $n\mapsto f(n)$ that expresses no additional mathematical structure.

\paragraph{3. Direct consequence of a definition.}
We exclude a relation if one object is already defined through the other or if the statement follows by directly expanding one definition. This includes recovering an object from a formula that defines it, applying a formula stated in the \textsc{Formula} or \textsc{Comments} section of an entry, taking the obvious inverse of a recorded transformation, or reading one object from the other without an additional theorem or structural step. The presence of a formula in a comment is not by itself disqualifying: a formula connecting two independent constructions nontrivially may be retained, while its documentation is assessed separately under novelty.

\paragraph{4. Universal reconstruction.}
We exclude a relation if its proof applies a procedure that can reconstruct essentially any object of the relevant class, such as a universal Newton, Euler, Möbius, Stirling, or binomial transform, standard inversion of a generating function, a general coefficient-recovery procedure, a generic change between representations, or reconstruction from an arbitrary set of intermediate values. Standard theory is not itself disqualifying when it yields a specific assertion in which the two entries are essential.

\paragraph{5. Mechanical intermediate representation.}
We exclude a relation if one object serves only as an intermediate carrier and the other object is then mechanically recovered by a procedure that would work in essentially the same way for arbitrary inputs. A canonical entry carrying a classical object is not excluded merely for that reason: it may be retained when the resulting route is a useful exact connection for the reader. The question is whether the displayed relation provides mathematical content beyond the carrier's ordinary definition.

\paragraph{6. Common parametric family.}
We exclude a relation if A and B are merely two members of a common family and the statement only says that they have the same defining scheme, that one is obtained by changing a parameter, that one is a special case of the other, that they are neighbouring or selected components of a common construction, or that they are connected by a standard family transformation without a separate result for this pair. Membership in a common family is not disqualifying when the pair also has a specific bijection, identity, or structural interpretation.

\paragraph{7. Coincidence through a small number of values.}
We exclude a relation whose content is only a coincidence among a few constants or initial terms, the use of one or two numbers from an entry, substitution of a numerical value into a known formula, or reconstruction from a finite table without additional structure. For example, period five alone does not give a substantive relation: any two period-five sequences can be matched by a five-value table.

\paragraph{8. Redundant restatement of an existing relation.}
We exclude a relation if it adds no content to an existing relation: for example, if A is related to C while B is merely another representation of C, if A--B follows by mechanically changing the representation in A--C, or if several outputs repeat one template with only parameters or notation changed.

The use of a standard method, such as generating functions or Riordan arrays, is therefore not by itself a reason for exclusion. We retain a relation when showing it would give a mathematically literate reader a useful exact route or interpretation, even when the route is standard, documented elsewhere, or passes through a canonical classical object. We exclude it when the displayed pair adds no useful content beyond a definition, a change of representation, or a mechanically reproducible recipe. Novelty is assessed separately: absence of an OEIS cross-reference does not by itself make a relation novel or substantive.

\section{Graded links}
\label{app:links}
\begin{table}[H]
\centering
\small
\begin{tabular}{lrrrr}
\toprule
 & gold & silver & trivia & bare cross-refs \\
\midrule
graded by the screener (10\,987 pairs) & 369 & 2\,436 & 8\,182 & --- \\
used for training (10\,770 pairs) & 358 & 2\,422 & 7\,990 & 21\,493 \\
\bottomrule
\end{tabular}
\caption{Graded cross-references in the corpus. The screener graded every pair whose \textsc{formula} or \textsc{comments} lines name the partner in $S_0$; the training set differs by the gold verdicts downgraded on re-reading and by the exclusion of text twins and of the test pairs of Sec.~\ref{sec:labels}. Bare cross-references name the partner only in \textsc{crossrefs} and were not graded.}
\label{tab:teacher}
\end{table}

\section{Scorer benchmark}
\label{app:scorers}
\begin{table}[H]
\centering
\footnotesize
\setlength{\tabcolsep}{3pt}
\begin{tabular}{@{}p{4.6cm}cccc@{}}
\toprule
scorer & \multicolumn{2}{c}{percentile of the 74 pairs} & 74 pairs in & most-referenced in \\
 & random & matched & top-1000 & top-1000 \\
\midrule
surface text, untrained cosine (char TF-IDF) & 63.5 & 59.0 & 2 & 1.8\% \\
surface text, trained: LSA + logistic (Sec.~\ref{sec:textscorer}) & --- & 83.0 & --- & --- \\
surface text, trained: LSA + boosting & --- & 90.5 & --- & --- \\
raw cosine, one layer, untrained & 70.3 & 66.3 & 1 & 1.7\% \\
\textbf{linear classifier (used)} & 77.3 & 77.4 & 1 & 6.3\% \\
logistic, pointwise & 83.2 & 78.8 & 6 & 15.5\% \\
logistic, pairwise & 77.3 & 71.9 & 10 & 9.6\% \\
RankSVM & 74.8 & 71.6 & 10 & 11.0\% \\
boosting, pointwise & 92.2 & 84.5 & 14 & 21.4\% \\
boosting, LambdaMART & 95.4 & 84.5 & 14 & 21.0\% \\
boosting, popularity-matched negatives & 91.6 & 90.0 & 11 & 9.7\% \\
\bottomrule
\end{tabular}
\caption{Scorers trained on the same 321 activation features (surface text excepted, trained on the same labels), tested on the 74 substantive future pairs (Sec.~\ref{sec:labels}) against 500\,000 random pairs and against popularity-matched peers (100 per pair for the text rows, as in Sec.~\ref{sec:question}); the column ``most-referenced'' counts pairs with an entry mentioned by at least 400 others (2.9\% of random pairs). Averages favour boosting; the linear classifier puts the fewest most-referenced pairs in its top 1\,000 (Sec.~\ref{sec:classifier}). The extreme top is fragile: two probes differing only in which layers feed the PCA block share one pair in their top-100 while their scores correlate at 0.5 over 500\,000 random pairs. }
\label{tab:scorers}
\end{table}

\paragraph{Held-out averages on the sets added after $S_1$.} The topic-matched test of Sec.~\ref{sec:heldout} repeated on the 48 substantive pairs and the 89 links added after $S_1$ (Sec.~\ref{sec:labels}) gives all four scorers between 0.55 and 0.62 AUC, within one standard error of each other (0.04 and 0.03). Model size does not order the classifiers on any of the four sets. On the popularity-matched exam of Sec.~\ref{sec:question}, which has no topic matching, the trained text scorer reaches the 83rd percentile with logistic regression and the 90th with boosting (rows above): the 74 pairs are about 45 distinct facts, more than half of them of two templates, a constant given as a limit or a sum over an arithmetic function and one theorem recorded across several entries of a family, and a scorer that has learned a template ranks its instances above peers of other topics. On the sets added after $S_1$ the averages do not separate the scorers; on the sets added after $S_0$ they are those of Fig.~\ref{fig:controls}. With the first classifier we trained, on all cross-references of $S_0$ without grading, the same test gave AUC 0.876 for activations against 0.799 for text on the 927 later links, and 0.758 against 0.498 on the 265 pairs whose definitions share no word; the gap closed when both were trained on the graded labels.

\section{Search configurations}
\label{app:configs}
The three configurations of Sec.~\ref{sec:configs}, whose outputs together form the 62 verified statements.

\begin{table}[H]
\centering
\small
\setlength{\tabcolsep}{3pt}
\begin{tabular*}{\columnwidth}{@{\extracolsep{\fill}}lccc@{}}
\toprule
 & variant 1 & variant 2 & main \\
\midrule
entries & 5\,000 & 5\,000 & 10\,000 \\
positives & all links & gold, silver & gold, silver \\
 & 15\,000 & 1\,330 & 2\,780 \\
negatives & random & trivia, bare, & trivia, bare, \\
 & & random & random \\
one pair per entry & no & yes & yes \\
depth processed & 100 & 100 & 500 \\
verified & 3 & 15 & 44 \\
\bottomrule
\end{tabular*}
\caption{The three search configurations. Labels are those of Sec.~\ref{sec:labels}; bare stands for pairs named only in CROSSREFS. The main configuration adds the caps of Sec.~\ref{sec:classifier} and is the one tested in Sec.~\ref{sec:controls}.}
\label{tab:configs}
\end{table}

Table~\ref{tab:stage-time} gives the wall-clock time of each stage of the main configuration.

\begin{table}[H]
\centering
\small
\setlength{\tabcolsep}{3pt}
\begin{tabular*}{\columnwidth}{@{\extracolsep{\fill}}>{\raggedright\arraybackslash}p{0.58\columnwidth}r@{}}
\toprule
stage & time \\
\midrule
labelling of linked pairs (Claude Sonnet 5) & $\sim$2\,h\,40\,min \\
Qwen3-32B activations, 10\,000 entries & $\sim$1\,h\,20\,min \\
classifier training, scoring of 50\,million pairs & 8\,min \\
ranked queue of 500 pairs & 20\,s \\
filter (Claude Sonnet 5, no tools), 584 pairs & 43\,min \\
agent (Claude Opus 5 with tools), 118 pairs & 2\,h\,56\,min \\
\bottomrule
\end{tabular*}
\caption{Wall-clock time of each stage of the main configuration. Labelling (Sec.~\ref{sec:labels}) ran in two parts, about 40 minutes for the popular entries and about two hours for pairs with a lesser-known entry. The filter reads the 500 queued pairs and 84 calibration pairs in batches of ten with four workers. The agent runs six pairs in parallel, with a median of seven minutes per pair; its time includes sandbox verification.}
\label{tab:stage-time}
\end{table}

\clearpage
\begingroup
\providecommand{\A}[1]{\href{https://oeis.org/A#1}{A#1}}
\newenvironment{steps}
  {\begin{itemize}[leftmargin=2.6em, labelwidth=2.1em, labelsep=0.5em,
                   align=left, topsep=4pt, itemsep=3pt, parsep=0pt]}
  {\end{itemize}}
\setlength{\parskip}{4pt}
\setlength{\parindent}{0pt}

\section{Derivations for the 62 statements}
\label{app:proofs}

Companion to Appendix~\ref{app:list}: one entry per statement, in the same order --- by novelty level, then by content class.

\textbf{Conventions.}
\begin{itemize}[leftmargin=1.4em, itemsep=2pt, parsep=0pt, topsep=4pt]
\item The formulas recorded in the OEIS entries (the definition of an entry, and the formulas in its \textsc{formula} section, each attributed there to a contributor) are taken as given and are cited as ``the formula recorded in the entry''. Each derivation establishes the relation between two entries from what the entries record; it does not re-prove the recorded formulas, which have their own provenance in the OEIS.
\item Classical results are cited by name; the index at the end lists them with the statements that use them. Literature references are to be added.
\item Every derivation is an exact chain: formal-power-series identities, factorial algebra, induction, or a closure by the Sturm bound.
\item The step numbers are the derivation's own; a gap in the numbering marks an omitted step (a numerical check or a side remark), so cross-references between steps stay valid.
\item A \textbf{Remark} after a proof points to the step a referee should look at first.
\item All derivations were produced by a language model and have \textbf{not} been checked by a mathematician.
\end{itemize}

\subsection*{Index of classical results used}
Compiled from the derivations above; each needs a literature reference before submission.

\begin{itemize}[leftmargin=1.4em, itemsep=2pt, parsep=0pt, topsep=4pt]
\item \textbf{Jacobi triple product} --- items 8, 10, 13, 29, 51, 52
\item \textbf{Euler's product identities}: $(-x;x)_\infty = E(x^2)/E(x)$; the splitting of $(x;x)_\infty$ modulo 5, $G\,H = (x^5;x^5)_\infty/(x;x)_\infty$ --- items 8, 26, 27, 29
\item \textbf{Euler's pentagonal-number product} $1/P(x) = \prod_{k\ge1}(1-x^k)$ --- items 45, 50
\item \textbf{von Staudt--Clausen theorem} --- items 36, 55, 58, 62
\item \textbf{Euler's formula} $\zeta(2m) = (-1)^{m+1}(2\pi)^{2m}B_{2m}/(2\,(2m)!)$ and $\zeta(-n) = -B_{n+1}/(n+1)$ --- items 36, 55, 58
\item \textbf{Functional equation of $\zeta$} (logarithmic-derivative form); \textbf{Euler--Maclaurin summation} --- item 36
\item \textbf{Jacobi's two-square theorem} $r_2(n) = 4\sum_{d\mid n}\chi_{-4}(d)$, multiplicative form --- items 13, 38, 54, 59
\item \textbf{Jacobi's four-square theorem}; \textbf{Gauss's three-square formula} via Hurwitz class numbers; \textbf{class-number formula for non-maximal orders} (Cox, Thm.~7.24) --- item 11
\item \textbf{Eisenstein series} $E_k = 1 - (2k/B_k)\sum_{n\ge1}\sigma_{k-1}(n)q^n$; $\Delta = \eta^{24}$; $\dim S_{12} = 1$ (valence formula) --- item 58
\item \textbf{Sturm bound} for modular forms --- items 9, 32, 58
\item \textbf{Borwein--Borwein cubic theta functions}: $b = \eta(q)^3/\eta(q^3) = (3a(q^3)-a(q))/2$, $a^3 = b^3 + c^3$; \textbf{Borwein--Borwein--Garvan} $E_6 = a^6 - 20a^3c^3 - 8c^6$ --- item 9
\item \textbf{Ramanujan}: $1/r^5 - 11 - r^5 = (\eta(x)/\eta(x^5))^6$; \textbf{Klein's icosahedral formula} for $j$ in the Rogers--Ramanujan continued fraction --- item 8
\item \textbf{Chowla--Selberg evaluations} $\eta(i) = \Gamma(1/4)/(2\pi^{3/4})$, $\theta_3(e^{-\pi}) = \pi^{1/4}/\Gamma(3/4)$; \textbf{Euler's reflection formula} --- item 54
\item \textbf{Dedekind eta} $\eta = q^{1/24}\prod_{n\ge1}(1-q^n)$; $E_2 = 1 + 24\,q\,\eta'/\eta$ --- items 37, 54
\item \textbf{Dedekind--Kummer splitting law} in $\mathbb{Q}(\sqrt5)$; $\mathbb{Z}[(1+\sqrt5)/2]$ is a PID --- item 60
\item \textbf{Fermat/Lagrange}: $p \mid 2^L - 1 \iff \operatorname{ord}_2(p) \mid L$ --- item 47
\item \textbf{M\"obius inversion}; \textbf{Gauss} $\sum_{d\mid n}\varphi(d) = n$; the Dirichlet series of $n\varphi(n)$ is $\zeta(s-2)/\zeta(s-1)$ --- items 20, 45, 57
\item \textbf{Euler product} of a nonnegative multiplicative function --- items 46, 53
\item \textbf{Catalan functional equation} $A=1+xA^2$, uniqueness of its formal solution; \textbf{Lambert's generalized binomial series} (Graham--Knuth--Patashnik, \emph{Concrete Mathematics}, eq.~(5.59)) --- item 43
\item \textbf{Lagrange inversion}; \textbf{Fuss--Catalan substitution} --- items 4, 12
\item \textbf{Newton interpolation} on central factorials; \textbf{Faulhaber's theorem}; central factorial numbers \A{036969} --- item 3
\item \textbf{Newton's binomial series} $(1-y)^{-k} = \sum_{m\ge0} k^{(m)}y^m/m!$ --- item 18
\item \textbf{Stirling inversion} ($s$ and $S$ mutually inverse); Stirling transform of an e.g.f.\ under $x \mapsto e^x - 1$ --- item 39
\item \textbf{Exponential Riordan group}: product $[g_1,f_1][g_2,f_2]$ and inverse $[1/(g\circ\bar f), \bar f]$ --- item 6
\item \textbf{Riordan array} factorisation of \A{124733} (Bala) --- item 4
\item \textbf{Beatty/Wythoff}: $\lfloor n\phi^2\rfloor = \lfloor n\phi\rfloor + n$, $A(B(n)) = A(n) + B(n)$ --- item 23
\item \textbf{Lucas's theorem}; the \textbf{Lyness 5-cycle} --- item 5
\item \textbf{Chebyshev}: $T_N(\cosh u) = \cosh Nu$; product-to-sum for $\sinh\cdot\cosh$ --- item 2
\item \textbf{Lucas sequences} $U_n(P,Q)$ and their binomial expansion (Dickson polynomials of the second kind) --- item 41
\item \textbf{Fibonacci--Lucas identity} $L_m = F_{m-1} + F_{m+1}$; variation of constants for an inhomogeneous Fibonacci recursion --- item 42
\item \textbf{Goulden--Jackson cluster method} --- item 7
\item \textbf{Gessel's formula} for permutations with $\mathrm{LIS} \le 3$ (Conway--Guttmann form); binomial absorption identities --- items 21, 33
\item \textbf{Symbolic method}: MSET versus SEQ constructions (P\'olya theory) --- item 30
\item \textbf{Uniqueness of the compositional inverse} of a power series; Ramanujan's theta $f(a,b)$ and its symmetry --- item 51
\item \textbf{Unique factorisation} $n = (\text{3-smooth}) \times (\text{coprime to }6)$; $n = (\text{odd squarefree}) \times (\text{square or twice a square})$ --- items 26, 53
\item \textbf{Deutsch's formula} for \A{008282} in terms of Euler numbers --- item 7
\end{itemize}

\subsection*{1. \A{230877} $\times$ \A{334302} (\nov{2}, \sig{1})}
\textbf{Statement.} With $\lambda(n)$ the partition encoded by $n$ as in \A{125106} and $k$ its number of parts, $w(n)=k\max\lambda(n)+\frac{k(k+1)}{2}-|\lambda(n)|$; the partitions $\lambda(1),\lambda(2),\ldots$ ordered by sum, length and reverse lexicographic order are the rows of \A{334302}.

\textbf{Proof.}
Induction on the binary expansion. Base $n=1$: $\lambda=(1)$, and $1\cdot 1+1-1=1=a(1)$. The step $n\to 2n$ maps $(k,\max\lambda,|\lambda|)\to(k,\max\lambda+1,|\lambda|+k)$, which leaves $k\max\lambda+\binom{k+1}{2}-|\lambda|$ unchanged, as it leaves $a(2n)=a(n)$. The step $n\to 2n+1$ maps the triple to $(k+1,\max\lambda,|\lambda|+1)$; the expression grows by $\max\lambda+k=\operatorname{bitlength}(2n+1)$, exactly the increment in the definition of \A{230877}. The map \A{125106} runs through all partitions exactly once: by its definition each $1$ in the binary expansion of $n$ is a part of size $1+(\text{number of }0\text{'s to its right})$, so if the parts are listed in non-decreasing order $\lambda_{(1)}\le\dots\le\lambda_{(k)}$, the positions of the $1$'s (counted from the least significant bit, starting at $0$) are $p_i=\lambda_{(i)}+i-2$, a strictly increasing sequence; conversely every strictly increasing $p_1<\dots<p_k$ gives non-decreasing $\lambda_{(i)}=p_i-i+2\ge 1$. Hence $n\mapsto\lambda(n)$ is a bijection from the positive integers onto the nonempty partitions. Ordering by sum, length and reverse lexicographic order is the definition of \A{334302}. Invertibility: along the path $n\to\lfloor n/2\rfloor\to\dots\to 1$ the value $a$ changes exactly at the $1$-bits, so the bits of $n$, and with them the rows of the table, are recovered from the terms of \A{230877} alone.

\textbf{Remark.} The bijectivity of the encoding \A{125106} is not stated in that entry; it is proved above. The comment there that the Heinz numbers of the rows form \A{005940}, a permutation of the positive integers, is consistent with it.

\subsection*{2. \A{053120} $\times$ \A{060187} (\nov{2}, \sig{1})}
\textbf{Statement.} $e^{(2i-1)u}=T_{2i-1}(\cosh u)+\sinh u\,\bigl(T_0(\cosh u)+2\sum_{j=1}^{i-1}T_{2j}(\cosh u)\bigr)$; row $n$ of the type-B Eulerian numbers is the coefficients of $x^1,\ldots,x^n$ in $(1-x)^n\sum_{i=1}^n(2i-1)^{n-1}x^i$, with $(2i-1)^{n-1}=(n-1)!\,[u^{n-1}]$ of the first series.

\textbf{Proof.}
\begin{steps}
\item[1.] $2\sinh u\,\cosh 2ju=\sinh(2j+1)u-\sinh(2j-1)u$; telescoping over $j=1,\dots,i-1$ together with $T_N(\cosh u)=\cosh Nu$ assembles the displayed formula for $e^{(2i-1)u}$ from the rows of \A{053120}.

\item[2.] Hence $(2i-1)^{n-1}=(n-1)!\,[u^{n-1}]\,e^{(2i-1)u}$.

\item[3.] By the formula recorded in \A{060187} (Bala), $R(n,x)/(1-x)^n=\sum_{i\ge1}(2i-1)^{n-1}x^{i}$; the polynomial $R(n,\cdot)$ has degree $n$, so when the coefficients of $x^1,\dots,x^n$ are taken the sum can be cut at $i=n$.
\end{steps}

\subsection*{3. \A{103438} $\times$ \A{001498} (\nov{1}, \sig{1})}
\textbf{Statement.} With the central factorial numbers $\mathrm{CF}$ (\A{036969}), $T(2p+1,n)=\sum_{k=1}^{p+1}(2k-2)!!\,\mathrm{CF}(p+1,k)\,B(n,k)$ and $T(2p,n)=(2n+1)\sum_{k=1}^{p}\frac{k}{2k+1}(2k-2)!!\,\mathrm{CF}(p,k)\,B(n,k)$.

\textbf{Proof.}
\begin{steps}
\item[1.] $B(n,k)=\A{001498}(n,k)=\frac{(n+k)!}{2^k(n-k)!\,k!}$, hence $\frac{(n+k)!}{(n-k)!}=2^k\,k!\,B(n,k)$.

\item[2.] Define the monic central factorials
\[
P_d(j)=\prod_{i=0}^{d-1}(j-\nu_i)
\]
over the node list $\nu=0,1,-1,2,-2,3,-3,\dots$; equivalently
\[
P_{2k}(j)=\prod_{i=-k}^{k-1}(j+i),\qquad P_{2k+1}(j)=\prod_{i=-k}^{k}(j+i).
\]

\item[3.] By the standard telescoping of falling factorials, $2k\,\frac{(j+k-1)!}{(j-k)!}=f(j)-f(j-1)$ with $f(j)=\frac{(j+k)!}{(j-k)!}$, and $(2k+1)\,\frac{(j+k-1)!}{(j-k-1)!}=g(j)-g(j-1)$ with $g(j)=\frac{(j+k)!}{(j-k-1)!}$. Summing over $j=1,\dots,n$ and using step 1 gives
\begin{align*}
\sum_{j=1}^{n}P_{2k-1}(j)&=\frac{(n+k)!}{(n-k)!\,2k}=(2k-2)!!\,B(n,k),\\
\sum_{j=1}^{n}P_{2k}(j)&=\frac{(n+k)!}{(n-k-1)!\,(2k+1)}=\frac{(2k)!!}{2k+1}(n-k)\,B(n,k).
\end{align*}

\item[4.] By Newton interpolation, the $P_d$ are monic of degree $d$, so $j^{m}=\sum_{d=0}^{m}t(m,d)\,P_d(j)$ uniquely, and from $j\,P_{2k}=P_{2k+1}-k\,P_{2k}$, $j\,P_{2k+1}=P_{2k+2}+(k+1)P_{2k+1}$ one gets $t(m+1,d)=t(m,d-1)+e(d)\,t(m,d)$ with $e(2k)=-k$, $e(2k+1)=k+1$, $t(0,0)=1$.

\item[5.] $j^{2p+1}$ is an odd function, so only odd $d$ occur; with $x=j^2$ one has $P_{2k-1}(j)=j\prod_{i=1}^{k-1}(x-i^2)$, so $t(2p+1,2k-1)$ is the Newton coefficient of $x^{p}$ at the nodes $1,4,9,\dots$, i.e. the central factorial number $\A{036969}(p+1,k)$ (recurrence $T(p,k)=k^2\,T(p-1,k)+T(p-1,k-1)$).

\item[6.] Applying step 4's recurrence once more gives $t(2p,2k)=t(2p-1,2k-1)=\A{036969}(p,k)$ and $t(2p,2k+1)=(k+1)\A{036969}(p,k+1)$. Substituting step 3 and combining the two sums, the bracket
\[
\frac{(2k)!!\,(n-k)}{2k+1}+k\,(2k-2)!! = \frac{k\,(2k-2)!!\,(2n+1)}{2k+1}
\]
factors out $(2n+1)$, giving the even-$m$ formula (this is exactly Faulhaber's theorem: odd power sums are polynomials in $N=n(n+1)/2$, even ones are $(2n+1)$ times such a polynomial).
\end{steps}

\subsection*{4. \A{124733} $\times$ \A{266213} (\nov{2}, \sig{1})}
\textbf{Statement.} With $h$ the Riordan function of $W$ ($h=x(1+3h+h^2)$), $\sum_{k=0}^n N(m,k)\bigl(W(n-1,k-1)+W(n-1,k)\bigr)=[x^n]\bigl(\frac{1-x}{1-5x}\bigr)^{m/2}$.

\textbf{Proof.}
\begin{steps}
\item[1.] By the \A{124733} entry itself (Peter Bala, Sep 06 2022), \A{124733} is the Riordan array $(f(x),x\,g(x))$ with $f=\frac{1-\sqrt{(1-5x)/(1-x)}}{2x}$ (\A{007317}) and $x\,g=\frac{1-3x-\sqrt{1-6x+5x^2}}{2x}=:h$ (\A{002212} shifted).

\item[2.] Squaring $2x\,h-(1-3x)=-\sqrt{1-6x+5x^2}$ gives $x\,h^2-(1-3x)h+x=0$, i.e. $h=x(1+3h+h^2)$, hence $x=h/(1+3h+h^2)$.

\item[3.] Algebraically, $1-x=\frac{1+3h+h^2-h}{1+3h+h^2}=\frac{(1+h)^2}{1+3h+h^2}$ and $1-5x=\frac{1+3h+h^2-5h}{1+3h+h^2}=\frac{(1-h)^2}{1+3h+h^2}$. Dividing, $\left(\frac{1+h}{1-h}\right)^{2}=\frac{1-x}{1-5x}$.

\item[4.] Consequently $f=\frac{1-(1-h)/(1+h)}{2x}=\frac{h}{x(1+h)}$, so $f\,h^{k}=\frac{h^{k+1}}{x(1+h)}$, equivalently $h^{k}=x\,f\,h^{k-1}+x\,f\,h^{k}$; taking $[x^n]$ gives $[x^n]\,h^{k}=\A{124733}(n-1,k-1)+\A{124733}(n-1,k)$ for $k\ge1$ (and $[x^n]\,h^{0}=[n=0]$).

\item[5.] By the \A{266213} entry itself (Ilya Gutkovskiy, May 23 2017), $\sum_{k\ge0}\A{266213}(m,k)\,z^{k}=\left(\frac{1+z}{1-z}\right)^{m}$. Substituting the delta series $z=h(x)$ is legitimate and, by step 3, gives
\[
\sum_k \A{266213}(m,k)\,h^{k}=\left(\frac{1+h}{1-h}\right)^{m}=\left(\frac{1-x}{1-5x}\right)^{m/2}.
\]

\item[6.] Taking $[x^n]$ and inserting step 4 gives the statement.
\end{steps}

\subsection*{5. \A{000127} $\times$ \A{076839} (\novna, \sig{0})}
\textbf{Statement.} $y(n)=1+\lfloor (r(n+4)\bmod 5)/2\rfloor$.

\textbf{Proof.}
\begin{steps}
\item[1.] $\A{000127}(n)=\sum_{k\le4}\binom{n-1}{k}$ is a polynomial of degree $4$ in $n$, so $(S-1)^{5}a=0$ for the shift operator $S$. Modulo $5$ the binomials $\binom{5}{i}$ vanish for $0<i<5$, hence $(S-1)^{5}\equiv S^{5}-1$: the residues of \A{000127} modulo $5$ are exactly $5$-periodic; by Lucas's theorem $\A{000127}(n)\equiv 2^{(n-1)\bmod 5} \pmod 5$.

\item[2.] The orbit of the Lyness cycle from $(1,1)$ is $1,1,2,3,2$, of period $5$ (direct computation; every orbit of the Lyness $5$-cycle has period $5$).

\item[3.] Both sides are $5$-periodic, so it suffices to compare $n=1,\dots,5$: the residues $1,1,2,4,3$ under $f(x)=1+\lfloor x/2\rfloor$ give $1,1,2,3,2$, matching. The content is the phase alignment: the residue $1$ occurs twice per period ($2^{0}\equiv 2^{4}\equiv 1 \pmod 5$), and exactly at these two phases $\A{076839}=1$.
\end{steps}

\subsection*{6. \A{132440} $\times$ \A{039755} (\nov{2}, \sig{1})}
\textbf{Statement.} $M^{-1}P^2M=I+2T$, with $P$ the Pascal matrix.

\textbf{Proof.}
\begin{steps}
\item[1.] As stated in \A{039755} (Peter Bala, Jun 23 2014), \A{039755} is the exponential Riordan array $M=[g,f]=[e^x,(e^{2x}-1)/2]$, i.e.\ $M(n,k)=n!\,[x^n]\,e^x f(x)^k/k!$.

\item[2.] As stated in \A{132440} (W.~Lang, Oct 14 2010), $T=\A{132440}$ satisfies $T=\log(P)$, $P=\A{007318}$; equivalently $\exp(tT)=[e^{tx},x]$ with entries $\binom{n}{k}t^{n-k}$. For $t=1$ this is Pascal, for $t=2$ it is \A{038207}.

\item[3.] By the exponential Riordan group law, $[g_1,f_1]\cdot[g_2,f_2]=[g_1\,(g_2\circ f_1),\,f_2\circ f_1]$, and $[g,f]^{-1}=[1/(g\circ\bar f),\,\bar f]$, where $\bar f$ is the compositional inverse of $f$.

\item[4.] Applying step 3 twice: $M^{-1}\cdot[e^{tx},x]=[e^{t\bar f}/(g\circ\bar f),\,\bar f]$, and then times $M$ this is $\bigl[(e^{t\bar f}/(g\circ\bar f))\,(g\circ\bar f),\,f\circ\bar f\bigr]=[e^{t\bar f(x)},x]$. The factor $g$ cancels identically, so the result depends only on $f$.

\item[5.] Here $f(x)=(e^{2x}-1)/2$ inverts to $\bar f(x)=\tfrac12\log(1+2x)$, hence $e^{2\bar f(x)}=1+2x$ and $M^{-1}\exp(2T)M=[1+2x,x]$, whose $(n,k)$ entry is $\frac{n!}{k!}[x^{n-k}](1+2x)=\delta_{n,k}+2n\,\delta_{n-1,k}=(I+2T)(n,k)$. This is the claimed identity.
\end{steps}

\subsection*{7. \A{008282} $\times$ \A{162975} (\novna, \sig{0})}
\textbf{Statement.} Both triangles reduce to the generating function $e^x\sec x$; from it the Euler numbers, and from them by Deutsch's formula the whole first triangle.

\textbf{Proof.}
\begin{steps}
\item[1.] By the Goulden--Jackson cluster method (Flajolet--Sedgewick p.~210; this is the formula recorded in \A{162975}),
\[
F(x,y):=\sum_n P_n(y)\frac{x^n}{n!}=\frac{1}{1-x-\sum_{n,k}I(n,k)(y-1)^k x^n/n!},
\]
where
\[
\sum_{n,k}I(n,k)y^kx^n=\frac{y\,x^3}{1-y\,x-y\,x^2}.
\]

\item[2.] By computation, expanding that ordinary generating function gives $c_m(u):=\sum_k I(m,k)u^k$ as the coefficients of $u\,z^3/(1-uz-uz^2)$, so $c_m=u\,c_{m-1}+u\,c_{m-2}$ for $m\ge2$ once extended backwards by $c_0=-1$, $c_1=1$ (then $c_2=0$, $c_3=u$, $c_4=u^2$, $c_5=u^2+u^3$, $c_6=2u^3+u^4$, matching the ordinary generating function).

\item[3.] Algebraically, with $\alpha,\beta$ the roots of $r^2=u\,r+u$ one gets $c_m=P\alpha^m+Q\beta^m$ for all $m\ge0$, so $1-x-\sum_{m\ge3}c_mx^m/m! = -(Pe^{\alpha x}+Qe^{\beta x})$. Solving $P+Q=-1$, $P\alpha+Q\beta=1$ and putting $A=1+\alpha$, $B=1+\beta$ (so $A+B=y+1$, $AB=1$) yields
\[
F(x,y)=\frac{(A-B)e^x}{A\,e^{Bx}-B\,e^{Ax}}.
\]
Equivalently, with $A=e^{i\theta}$, $y=2\cos\theta-1$: $F=\sin(\theta)\,e^{x(1-\cos\theta)}/\sin(\theta-x\sin\theta)$.

\item[4.] Specialising $\theta=\pi/2$ ($y=-1$, $A=i$, $B=-i$): $F(x,-1)=2i\,e^x/(i\,e^{-ix}+i\,e^{ix})=e^x/\cos(x)$.

\item[5.] Differentiating in $\theta$ at $\pi/2$ ($dy/d\theta=-2\sin\theta=-2$, so $d/dy=-(1/2)\,d/d\theta$): $dF/dy|_{y=-1}=e^x\sec(x)(\tan(x)-x)/2$.

\item[6.] Inverting, $\sec(x)=e^{-x}U(x)$ gives the secant numbers as the inverse binomial transform $s_n=\sum_j(-1)^{n-j}\binom{n}{j}u_j$; and $\tan(x)=x+2V(x)/U(x)$ gives the tangent numbers by triangular exponential generating function division ($u_0=1$). Then $\A{000111}(n)=s_n+t_n$.

\item[7.] By the formula recorded in \A{008282} (Emeric Deutsch),
\[
T(n,k)=\sum_{i=0}^{\lfloor (n-k)/2\rfloor}(-1)^i\binom{n-k}{2i}\,\A{000111}(n-2i)
\qquad\text{for }1\le k\le n.
\]
\end{steps}

\subsection*{8. \A{004009} $\times$ \A{122135} (\nov{1}, \sig{0})}
\textbf{Statement.} $E_4=(x;x)_\infty^8\,(G^{20}+228xG^{15}H^5+494x^2G^{10}H^{10}-228x^3G^5H^{15}+x^4H^{20})$, $G$ and $H$ the Rogers--Ramanujan series, recovered from the partner.

\textbf{Proof.}
\begin{steps}
\item[1.] By the formula recorded in \A{122135}, $B(x)=f(x^2,x^8)/f(-x,-x^4)$ (Somos, Nov 12 2016).

\item[2.] By the Jacobi triple product,
\[
f(-x,-x^4)=(x;x^5)(x^4;x^5)(x^5;x^5)=(x^5;x^5)/G,
\]
where $G=\sum x^{n^2}/(x;x)_n=1/((x;x^5)(x^4;x^5))$ is the first Rogers--Ramanujan function. Hence $G=B\,(x^5;x^5)/f(x^2,x^8)$ with $f(x^2,x^8)=(-x^2;x^{10})(-x^8;x^{10})(x^{10};x^{10})$.

\item[3.] By Euler's splitting of $(x;x)_\infty$ modulo $5$, $GH=(x^5;x^5)/(x;x)$ for $H=1/((x^2;x^5)(x^3;x^5))$. Hence $H=f(x^2,x^8)/((x;x)B)$.

\item[4.] By Ramanujan's identity, with $r=R(x)=x^{1/5}H/G$ the Rogers--Ramanujan continued fraction, $1/r^5-11-r^5=(\eta(x)/\eta(x^5))^6$, i.e. $r^{10}+11r^5-1=-r^5\bigl((x;x)/(x^5;x^5)\bigr)^6$.

\item[5.] By the Klein--Ramanujan icosahedral formula,
\[
j=-\frac{(r^{20}-228r^{15}+494r^{10}+228r^5+1)^3}{r^5(r^{10}+11r^5-1)^5}.
\]

\item[6.] Substituting step 4 into step 5: $r^5(r^{10}+11r^5-1)^5=-r^{30}(\eta/\eta_5)^{30}$, so
\[
j^{1/3}=\frac{r^{20}-228r^{15}+494r^{10}+228r^5+1}{r^{10}(\eta/\eta_5)^{10}}.
\]

\item[7.] $E_4=\eta(x)^8j^{1/3}$ (since $j=E_4^3/\Delta$, $\Delta=\eta^{24}$). Using $r^{10}=x^2H^{10}/G^{10}$ and
\[
\eta(x^5)^{10}/\bigl(\eta(x)^2(GH)^{10}\bigr)=x^2(x;x)^8
\]
(from step 3), this collapses to
\[
E_4=(x;x)^8\bigl(G^{20}+228xG^{15}H^5+494x^2G^{10}H^{10}-228x^3G^5H^{15}+x^4H^{20}\bigr).
\]

\item[8.] Composing steps 2, 3 and 7 gives $A$ from $B$.
\end{steps}

\subsection*{9. \A{013973} $\times$ \A{093829} (\nov{1}, \sig{0})}
\textbf{Statement.} $E_6=-27a^6+36a^3b^3-8b^6$, $a$ and $b$ the Borwein cubic theta functions; $a$ telescopes from the partner.

\textbf{Proof.}
\begin{steps}
\item[1.] $B(q)=(a(q)-a(q^2))/6$, where $a(q)=\sum_{m,n\in\mathbb Z}q^{m^2+mn+n^2}=\A{004016}$ is the first cubic AGM theta function.

\item[2.] Extracting the coefficient of $q^m$ gives $a_m-[2\mid m]\,a_{m/2}=6B_m$, i.e. the recursion $a_0=1$, $a_m=6B_m+a_{m/2}$; it is the telescoping of $a(q)=1+6\sum_{k\ge0}B(q^{2^k})$, finite in each coefficient because $a(q^{2^k})\to 1$.

\item[3.] By the Borwein--Borwein cubic AGM, $b(q)=\eta(q)^3/\eta(q^3)=(3a(q^3)-a(q))/2$.

\item[4.] By the cubic Fermat identity of Borwein--Borwein, $a(q)^3=b(q)^3+c(q)^3$ with $c(q)=3\eta(q^3)^3/\eta(q)=\A{005882}$.

\item[5.] By the Borwein--Borwein--Garvan level-3 evaluation of the level-1 Eisenstein series, $E_6(q)=a^6-20a^3c^3-8c^6$; substituting $c^3=a^3-b^3$ from step 4 gives the equivalent integral form $E_6=-27a^6+36a^3b^3-8b^6$.

\item[6.] Closure by the Sturm bound: $a$ and $b$ lie in $M_1(\Gamma_0(3),\chi_{-3})$, so $a^6$, $a^3b^3$, $b^6$ and $E_6$ all lie in $M_6(\Gamma_0(3))$ (trivial character, since $\chi_{-3}^6=1$). The index $[\mathrm{SL}(2,\mathbb Z):\Gamma_0(3)]=4$, so the Sturm bound is $k\cdot 4/12=2$: agreement of the two $q$-expansions through $q^2$ forces equality.
\end{steps}

\textbf{Remark.} Closure is by the Sturm bound on $M_6(\Gamma_0(3))$; the membership of $a$ and $b$ in $M_1(\Gamma_0(3),\chi_{-3})$ is taken from Borwein--Borwein and is the point to check.

\subsection*{10. \A{007325} $\times$ \A{227216} (\nov{2}, \sig{2})}
\textbf{Statement.} $T(q)^2R(q)^5=(q^5;q^5)_\infty^5/(q;q)_\infty$, the generating function of 5-core partitions.

\textbf{Proof.}
Split $E(q)=(q;q)_\infty$ by the residue of the exponent modulo $5$: $E(q)=a\,b\,E(q^5)$ with $a=(q,q^4;q^5)_\infty$, $b=(q^2,q^3;q^5)_\infty$, $E(q^5)=(q^5;q^5)_\infty$. By the Jacobi triple product, $f(-q^2,-q^3)=b\,E(q^5)$ and $f(-q)=E(q)$, so $\A{227216}=b^5E(q^5)^5/E(q)^3$; by definition of \A{007325}, $\A{007325}=a/b$. Hence
\[
\A{227216}^2\,\A{007325}^5\,E(q)
=\frac{b^{10}E(q^5)^{10}}{E(q)^6}\cdot\frac{a^5}{b^5}\cdot E(q)
=\frac{(ab)^5E(q^5)^{10}}{E(q)^5}=E(q^5)^5,
\]
since $ab=E(q)/E(q^5)$. Incidentally $E(q^5)^5/E(q)$ is the classical generating function of $5$-core partitions (\A{053723}).

\subsection*{11. \A{004011} $\times$ \A{000003} (\nov{1}, \sig{1})}
\textbf{Statement.} $N(n)=\varepsilon(n)\sum_{s\in\mathbb Z} r_3(n-s^2)$, $\varepsilon(n)=3$ for odd $n$ and $1$ otherwise, with $r_3$ given by Gauss's class-number formula through $h(-4m)$.

\textbf{Notation.} $h(n):=\A{000003}(n)=h(-4n)$, the number of classes of primitive positive definite forms of discriminant $-4n$; $h_w(D)=h(D)/(w(D)/2)$ with $w(-3)=6$, $w(-4)=4$ and $w(D)=2$ otherwise, so $h_w(-3)=1/3$ and $h_w(-4)=1/2$; the Hurwitz class number is $H(N)=\sum_{f^2\mid N,\ -N/f^2\equiv 0,1 \pmod 4} h_w(-N/f^2)$.

\textbf{Proof.}
\begin{steps}
\item[1.] By Gauss's three-square formula, in the Hurwitz form, $r_3(n)=12\,H(4n)$ for $n\equiv 1,2 \pmod 4$; $r_3(n)=24\,H(n)$ for $n\equiv 3\pmod 8$; $r_3(n)=0$ for $n\equiv 7\pmod 8$; and $r_3(4n)=r_3(n)$. Write $m=4^k m_0$ with $4\nmid m_0$ and put $S(m_0)=\sum_{f^2\mid m_0}h(m_0/f^2)$. The claim is: $r_3(m)=0$ if $m_0\equiv 7\pmod 8$; $r_3(m)=8\,S(m_0)$ if $m_0\equiv 3\pmod 8$; and $r_3(m)=12\,S(m_0)-6\,[m_0 \text{ is a square}]$ otherwise.

\item[2a.] $m_0$ odd, $m_0\equiv 1\pmod 4$. The $f$ with $f^2\mid 4m_0$ are $f=f'$ and $f=2f'$ with $f'$ odd, ${f'}^2\mid m_0$. For $f=2f'$ the discriminant $-4m_0/f^2=-m_0/{f'}^2\equiv 3\pmod 4$ is not a discriminant and drops out of $H$. For $f=f'$ odd the discriminant is $-4\,(m_0/f^2)$, of weight $1$ except when $m_0/f^2=1$, where $h_w(-4)=1/2=h(1)-1/2$. Hence $H(4m_0)=S(m_0)-\tfrac12[m_0 \text{ square}]$ and $r_3(m_0)=12\,S(m_0)-6\,[m_0 \text{ square}]$.

\item[2b.] $m_0\equiv 2\pmod 4$. From $f^2\mid 4m_0=8\,(m_0/2)$ with $m_0/2$ odd, $v_2(f)\le 1$. For $f=2f'$ the discriminant $-m_0/{f'}^2\equiv 2\pmod 4$ drops out; for $f$ odd the discriminant is $-4\,(m_0/f^2)$ with $m_0/f^2\ge 2$ even, weight $1$. Hence $H(4m_0)=S(m_0)$ and $r_3(m_0)=12\,S(m_0)$; $m_0$ is not a square.

\item[2c.] $m_0\equiv 3\pmod 8$. Here $H(m_0)=\sum_{f^2\mid m_0}h_w(-M)$ over $M=m_0/f^2$, and every such $M\equiv 3\pmod 8$ ($f$ odd, $f^2\equiv 1\pmod 8$). Claim: $h(-4M)=3\,h_w(-M)$ for every $M\equiv 3\pmod 8$. Write $-M=D\,g^2$ with $D$ fundamental and $g$ odd. By the class number formula for a non-maximal order (Cox, \emph{Primes of the form $x^2+ny^2$}, Thm.~7.24),
\[
h(D c^2)=\frac{h(D)\,c}{[\mathcal O_K^{*}:\mathcal O_c^{*}]}\prod_{p\mid c}\Bigl(1-\frac{(D/p)}{p}\Bigr).
\]
Applying it with $c=2g$ and $c=g$ gives
\[
\frac{h(-4M)}{h(-M)}=2\Bigl(1-\frac{(D/2)}{2}\Bigr)\frac{[\mathcal O_K^{*}:\mathcal O_g^{*}]}{[\mathcal O_K^{*}:\mathcal O_{2g}^{*}]}.
\]
Since $D\equiv -M/g^2\equiv 5\pmod 8$, the Kronecker symbol $(D/2)=-1$ and the middle factor is $3/2$. The unit indices are $1$ unless $D=-3$; for $D=-3$ one has $[\mathcal O_K^{*}:\mathcal O_{2g}^{*}]=3$ always, and $[\mathcal O_K^{*}:\mathcal O_g^{*}]=3$ for $g>1$, $=1$ for $g=1$ (i.e. $M=3$). Hence $h(-4M)=3\,h(-M)$ for $M>3$ and $h(-12)=h(-3)=1$ for $M=3$; as $h_w(-M)=h(-M)$ for $M>3$ and $h_w(-3)=1/3$, in all cases $h(-4M)=3\,h_w(-M)$. Therefore $H(m_0)=S(m_0)/3$ and $r_3(m_0)=24\,H(m_0)=8\,S(m_0)$.

\item[2d.] $m_0\equiv 7\pmod 8$: $r_3(m_0)=0$ (Legendre). Finally $r_3(4^k m_0)=r_3(m_0)$ transfers 2a--2d to all $m$.

\item[3.] Slicing a representation by four squares along one coordinate: $r_4(n)=\sum_{s\in\mathbb Z} r_3(n-s^2)$.

\item[4.] By Jacobi's four-square theorem, $r_4(n)=8\sum_{d\mid n,\ 4\nmid d}d$, which equals $8\,\sigma_{\mathrm{odd}}(n)$ for odd $n$ and $24\,\sigma_{\mathrm{odd}}(n)$ for even $n$; and $\A{004011}(n)=24\,\sigma_{\mathrm{odd}}(n)$ is the entry's own formula. Hence
\[
\A{004011}(n)=\varepsilon(n)\,r_4(n)=\varepsilon(n)\sum_s r_3(n-s^2)
\]
with $\varepsilon(n)=3$ for odd $n$ and $1$ otherwise, and $r_3$ given by steps 1--2.
\end{steps}

\textbf{Remark.} The delicate point is step 2c: the class-number formula for the order of conductor $2$, and the unit-index bookkeeping that makes the case $M=3$ fall under the same formula.

\subsection*{12. \A{039598} $\times$ \A{000139} (\nov{1}, \sig{2})}
\textbf{Statement.} With $T_3(x)=\sum_n \frac{n+1}{2}M(n)x^n$, $T_3=1+xT_3^3$, $C(x)=T_3(x/C(x))$ and $Q(n,k)=[x^{n-k}]\,C(x)^{2k+2}$: the maps are counted by ternary trees up to the factor $(n+1)/2$, followed by the Fuss--Catalan substitution.

\textbf{Proof.}
\begin{steps}
\item[1.] $\A{001764}(n)=\binom{3n}{n}/(2n+1)$ and $\A{000139}(n)=2\binom{3n}{n}/\bigl((n+1)(2n+1)\bigr)$, so $\A{001764}(n)=(n+1)\A{000139}(n)/2$, an exact integer.

\item[2.] $T_3=\sum_n \A{001764}(n)x^n$ is the unique series with $T_3=1+xT_3^3$ (Lagrange inversion).

\item[3.] The series $D=C(x\,T_3)$ satisfies $D=1+xT_3D^2$, and $T_3$ satisfies the same equation; uniqueness of the solution with constant term $1$ gives $T_3=C(xT_3)$, equivalently $C(y)=T_3(y/C(y))$.

\item[4.] Column generating function from the entry \A{039598} itself: $\sum_n T(n,k)x^n=x^{k}C(x)^{2k+2}$ (by Lagrange inversion it is equivalent to the entry's own row formula $\frac{k+1}{n+1}\binom{2n+2}{n-k}$). The map $u\mapsto T_3(x/u)$ is a contraction in the $x$-adic metric, so $N$ terms of \A{000139} determine the first $N$ Catalan numbers effectively.
\end{steps}

\subsection*{13. \A{122856} $\times$ \A{303363} (\novna, \sig{0})}
\textbf{Statement.} Deconvolving $N$ by the set $\{2^c+2^d: c \text{ even}, c\le d\}$ gives $U(m)$, the number of unordered representations of $m$ as a sum of two triangular numbers, and $t(n)=2U(m)-[m \text{ pronic}]$ with $4m+1$ the odd part of $3n+2$; $t(n)=0$ when that odd part is $3 \bmod 4$.

\textbf{Proof.}
\begin{steps}
\item[1.] $f(x,x^5)=\sum_k x^{3k^2-2k}$; from $3(3k^2-2k)+1=(3k-1)^2$ the support consists of the numbers $(m^2-1)/3$ with $m\equiv 2\pmod 3$, hence $[x^n]\,f^2=\tfrac14 r_2(3n+2)=\A{002654}(3n+2)$ (Jacobi; of each quadruple of signs $\pm m_1,\pm m_2$ exactly one pair satisfies $m\equiv 2\pmod 3$).

\item[2.] The set $S=\{2^c+2^d: c \text{ even},\ c\le d\}$ is repetition-free, so deconvolving \A{303363} by $S$ recovers $U(m)$ uniquely, the number of unordered representations of $m$ as a sum of two triangular numbers.

\item[3.] $(2a+1)^2+(2b+1)^2=8(T_a+T_b)+2$, and every representation of $8m+2$ as a sum of two squares uses two odd squares; hence the ordered count $t(m)=\tfrac14 r_2(8m+2)=\A{002654}(4m+1)$ and $t(m)=2U(m)-[m \text{ pronic}]$.

\item[4.] \A{002654} is multiplicative and vanishes at odd $k\equiv 3\pmod 4$; assembling along the odd part of $3n+2$ gives the formula of the statement.
\end{steps}

\subsection*{14. \A{112555} $\times$ \A{132440} (\nov{0}, \sig{0})}
\textbf{Statement.} With $P=\exp(B)$ the Pascal matrix, $T(n,k)=[n=k]+\sum_{j=0}^{n-1}(-1)^jP(j,n-1-k)$.

\textbf{Proof.}
\A{112555} is defined by $T^m=I+m(T-I)$, i.e. $N=T-I$ is strictly lower triangular with $N^2=0$. Its recorded formula gives the generating function $1/(1-xy)+x/\bigl((1-xy)(1+x+xy)\bigr)$. Expanding $1/(1+x+xy)=\sum_j(-1)^jx^{j}(1+y)^{j}$ turns the second summand into $N(n,k)=\sum_{j<n}(-1)^j\binom{j}{n-1-k}$, alternating partial sums of the columns of Pascal's triangle. And $\text{Pascal}=\exp(\A{132440})$ is the defining property of the generator in \A{132440}.

\subsection*{15. \A{005043} $\times$ \A{021009} (\nov{0}, \sig{0})}
\textbf{Statement.} $\sum_{k=0}^n L(n,k)\frac{(2k)!}{(k+1)!}=(-1)^n n!\,R(n)$.

\textbf{Proof.}
\begin{steps}
\item[1.] By the formula recorded in \A{021009}, $a(n,m)=(-1)^m\,n!\,\binom{n}{m}/m!$.

\item[2.] Consequently $T(n,0)=n!$ and $T(n,k)\,k!/T(n,0)=(-1)^k\binom{n}{k}$. So, after multiplying entry $k$ by $k!$ and dividing the row by its leading entry, row $n$ of the Laguerre triangle \emph{is} row $n$ of the signed Pascal (inverse binomial transform) matrix.

\item[3.] By the formula recorded in \A{005043} (Deleham/Smiley), $a(n)=\sum_{k=0}^{n}(-1)^{n-k}\binom{n}{k}\A{000108}(k)$, i.e. \A{005043} is the inverse binomial transform of the Catalan numbers.

\item[4.] Combining steps 2 and 3,
\[
\sum_{k=0}^{n}T(n,k)\,k!\,C_k=n!\sum_{k=0}^{n}(-1)^k\binom{n}{k}C_k=(-1)^n\,n!\,\A{005043}(n).
\]
The weight is elementary: $k!\,C_k=(2k)!/(k+1)!$.
\end{steps}

\subsection*{16. \A{034807} $\times$ \A{063967} (\nov{0}, \sig{0})}
\textbf{Statement.} $U(n,k)=\sum_{i=0}^{\lfloor n/2\rfloor}\frac{n-i}{n}L(n,i)\binom{n-i}{k}$.

\textbf{Proof.}
\begin{steps}
\item[1.] By the formula recorded in \A{034807} (Michael Somos), $T(n,i)=\binom{n-i}{i}+\binom{n-i-1}{i-1}=\frac{n\binom{n-i}{i}}{n-i}$ for $n\ge1$, $0\le 2i\le n$, and $T(0,0)=2$. Hence $\binom{n-i}{i}=\frac{n-i}{n}T(n,i)$ for $n\ge1$, and this is an exact integer division.

\item[2.] By the formula recorded in \A{063967} (Paul Barry), $U(n,k)=\sum_{j=0}^{n}\binom{j}{n-j}\binom{j}{k}$.

\item[3.] Re-index step 2 by $j=n-i$ (here $j$ runs over $\lceil n/2\rceil,\dots,n$ exactly as $i$ runs over $0,\dots,\lfloor n/2\rfloor$, since $\binom{j}{n-j}=0$ otherwise): $U(n,k)=\sum_{i=0}^{\lfloor n/2\rfloor}\binom{n-i}{i}\binom{n-i}{k}$.

\item[4.] Substituting step 1 into step 3, $U(n,k)=\sum_{i=0}^{\lfloor n/2\rfloor}\frac{n-i}{n}T(n,i)\binom{n-i}{k}$ for $n\ge1$. For $n=0$ the only exception $T(0,0)=2$ is handled by $\binom{0}{0}=T(0,0)/2=1$, giving $U(0,0)=1$.
\end{steps}

\subsection*{17. \A{003278} $\times$ \A{006368} (\novna, \sig{0})}
\textbf{Statement.} $c(0)=0$, $c(2m)=B(2c(m))$, $c(2m+1)=B(4c(m)+1)$ gives $S(n)=c(n-1)+1$ for all $n\ge1$.

\textbf{Proof.}
Induction on $m$: $c(m)$ is the binary expansion of $m$ read in base $3$, so that $c(2m)=3c(m)$ and $c(2m+1)=3c(m)+1$. The branches of the permutation, taken from the definition of \A{006368}, are $B(2x)=3x$ and $B(4x+1)=3x+1$, hence $c(2m)=B(2c(m))$ and $c(2m+1)=B(4c(m)+1)$, each branch applied exactly on its own domain. Finally $\A{003278}(n)=1+c(n-1)$ is the entry's own formula (the numbers whose ternary expansion uses only the digits $0$ and $1$, plus one).

\subsection*{18. \A{130534} $\times$ \A{297328} (\novna, \sig{0})}
\textbf{Statement.} With $P_m(k)=\sum_{i=0}^{m-1}s(m-1,i)k^{i+1}=k(k+1)\cdots(k+m-1)$,
\[
Q(n,k)=[x^n]\prod_{j\ge1}\sum_{m\ge0}\frac{P_m(k)}{m!}j^m x^{jm}.
\]

\textbf{Proof.}
\begin{steps}
\item[1.] Row $m-1$ of \A{130534} lists the coefficients of the rising factorial $k^{(m)}=k(k+1)\cdots(k+m-1)$ (the definition of the entry); here $s(m-1,i)$ are the Stirling numbers of the first kind.

\item[2.] Newton's binomial series, recorded in \A{130534} itself as the exponential generating function $(1-y)^{-x}$ of the row polynomials, gives
\[
(1-y)^{-k}=\sum_{m}\frac{k^{(m)}}{m!}\,y^{m}.
\]

\item[3.] The substitution $y=j\,x^{j}$ is legitimate (no constant term); the product over $j\ge1$ is the definition of column $k$ of \A{297328}:
\[
\prod_{j}\bigl(1-j\,x^{j}\bigr)^{-k}.
\]
The coefficient of $x^{n}$ involves only $j\le n$ and $m\le n/j$, a finite expression in the rows of \A{130534}. Integrality holds because $k^{(m)}/m! = \binom{k+m-1}{m}$.
\end{steps}

\subsection*{19. \A{049310} $\times$ \A{046534} (\novna, \sig{0})}
\textbf{Statement.} Reading the entry as a rational triangle $P(n,k)$, $3\binom nk=3P(n,k)+5\binom{n-2}{k-1}$ recovers Pascal's triangle, and then $S(n,k)=(-1)^{(n-k)/2}\binom{(n+k)/2}{k}$ for $n\ge k$ with $n-k$ even, else $0$.

\textbf{Proof.}
\begin{steps}
\item[1.] The difference $E(n,k)=\binom{n}{k}-P(n,k)$ obeys Pascal's rule everywhere except at the cell $(2,1)$, where $E(2,1)=2-\frac13=\frac53$. The unique such solution is a shifted copy of Pascal's triangle, $E(n,k)=\frac53\binom{n-2}{k-1}$, by induction on rows.

\item[2.] Inverting, $\binom{n}{k}=P(n,k)+\frac53\binom{n-2}{k-1}$; after multiplying by $3$ the recursion is integral, since the denominators of $P$ divide $3$.

\item[3.] The formula recorded in \A{049310} states $T(n,k)=(-1)^{(n-k)/2}\binom{(n+k)/2}{k}$ for $n-k$ even and $0$ otherwise; substituting the recovered Pascal triangle completes the derivation.
\end{steps}

\subsection*{20. \A{007429} $\times$ \A{007437} (\novna, \sig{0})}
\textbf{Statement.} $f(n)+\sum_{d\mid n}d\,\varphi(d)f(n/d)=2\sum_{d\mid n}g(d)$.

\textbf{Proof.}
In terms of Dirichlet series, $\A{007429}\leftrightarrow\zeta(s)^{2}\zeta(s-1)$; the triangular numbers $d(d+1)/2$ correspond to $\bigl(\zeta(s-1)+\zeta(s-2)\bigr)/2$, so $\A{007437}\leftrightarrow\zeta(s)\cdot\bigl(\zeta(s-1)+\zeta(s-2)\bigr)/2$; finally $n\,\varphi(n)\leftrightarrow\zeta(s-2)/\zeta(s-1)$. Then
\[
\zeta(s)^{2}\zeta(s-1)\Bigl(1+\frac{\zeta(s-2)}{\zeta(s-1)}\Bigr)
=\zeta(s)^{2}\bigl(\zeta(s-1)+\zeta(s-2)\bigr)
=\zeta(s)\cdot 2\cdot\operatorname{DGF}(\A{007437}),
\]
which is the identity of the statement; the final M\"obius inversion removes the extra factor $\zeta(s)$.

\subsection*{21. \A{047749} $\times$ \A{008315} (\novna, \sig{0})}
\textbf{Statement.} $a(2m)=T(3m,m)/(m+1)$, $a(2m+1)=2T(3m+1,m)/(m+2)$.

\textbf{Proof.}
The formula recorded in \A{008315} (the ballot numbers) gives
\[
T(n,k)=\binom{n}{k}-\binom{n}{k-1}=\binom{n}{k}\frac{n-2k+1}{n-k+1}.
\]
Even case, $(n,k)=(3m,m)$: $T(3m,m)=\binom{3m}{m}\frac{m+1}{2m+1}=(m+1)\,a(2m)$. Odd case: $T(3m+1,m)=\binom{3m+1}{m}\frac{m+2}{2m+2}$, and the absorption identity $\binom{3m+1}{m+1}=\binom{3m+1}{m}\frac{2m+1}{m+1}$ gives
\[
a(2m+1)=\frac{1}{2m+1}\binom{3m+1}{m+1}=\frac{2\,T(3m+1,m)}{m+2}.
\]
Both branches are pure factorial algebra.

\subsection*{22. \A{039599} $\times$ \A{294175} (\novna, \sig{0})}
\textbf{Statement.} The row sums of the triangle are $\binom{2n}{n}$, and $a(n)=\bigl(2^n-\binom{2\lceil n/2\rceil}{\lceil n/2\rceil}\bigr)/2$.

\textbf{Proof.}
The definition of \A{294175} gives
\[
a(n)=2^{n-1}+\frac{1+(-1)^{n}}{4}\binom{n}{n/2}-\binom{n}{\lfloor n/2\rfloor}.
\]
For $n=2m$ this equals $\bigl(2^{n}-\binom{2m}{m}\bigr)/2$; for $n=2m+1$ the identity $\binom{2m+2}{m+1}=2\binom{2m+1}{m}$ (Pascal) brings it to the same form $\bigl(2^{n}-\binom{2\lceil n/2\rceil}{\lceil n/2\rceil}\bigr)/2$. The partner enters through the row sums of \A{039599}, which equal the central binomial coefficients $\binom{2n}{n}$ by the formula recorded in \A{039599}: $\sum_{k}T(n,k)x^{k}$ at $x=1$ equals $\A{000984}(n)$.

\subsection*{23. \A{035513} $\times$ \A{101864} (\nov{1}, \sig{1})}
\textbf{Statement.} The BB-numbers are the third column of the array plus 2; from them $\lfloor n\varphi\rfloor$ and the whole array are recovered.

\textbf{Proof.}
\begin{steps}
\item[1.] By the Beatty--Wythoff fact, with $\varphi^2=\varphi+1$: $\{n\varphi^2\}=\{n\varphi\}$, hence $B(n):=\lfloor n\varphi^2\rfloor=\lfloor n\varphi\rfloor+n=A(n)+n$.

\item[2.] The Wythoff compound identity, with proof: put $\beta=\{n\varphi\}$ in $(0,1)$; then $\varphi B(n)=n\varphi^2+n\varphi-\varphi\beta=2n\varphi+n-\varphi\beta=2A(n)+n+\beta(2-\varphi)$; since $0<2-\varphi=0.3819\ldots<1$ and $0<\beta<1$, the added term lies in $(0,1)$, so $A(B(n))=2A(n)+n=A(n)+B(n)$.

\item[3.] Combining step 1 (applied at $k=B(n)$) and step 2: $\A{101864}(n)=B(B(n))=A(B(n))+B(n)=(2A(n)+n)+(A(n)+n)=3A(n)+2n$.

\item[4.] By Bottomley's formula in \A{035513} itself, $T(n,k)=F_{k+1}A(n)+F_k(n-1)$, with $F$ the Fibonacci numbers. With $F_4=3$, $F_3=2$ this gives $T(n,3)=3A(n)+2(n-1)=\A{101864}(n)-2$.

\item[5.] Inverting, step 3 gives $A(n)=(\A{101864}(n)-2n)/3$ exactly (an integer), which fed into step 4 regenerates the array.
\end{steps}

\subsection*{24. \A{009766} $\times$ \A{060693} (\novna, \sig{0})}
\textbf{Statement.} The rows of one are the reversed rows of the other multiplied by $\binom{n+2}{k+1}/(n+2)$, for all $0\le k\le n$.

\textbf{Proof.}
\begin{steps}
\item[1.] By the formula recorded in \A{009766}, $A(n,m)=\binom{n+m}{n}(n-m+1)/(n+1)$ for $0\le m\le n$.

\item[2.] By the formula recorded in \A{060693} (Deleham, Dec 07 2003), $B(n,k)=\binom{2n-k}{n}\binom{n}{k}/(n-k+1)$.

\item[3.] Substituting $m=n-k$ in step 1 gives $A(n,n-k)=\binom{2n-k}{n}(k+1)/(n+1)$. So $A(n,n-k)$ and $B(n,k)$ share the common factor $\binom{2n-k}{n}$; this is why the reversal is the right pairing.

\item[4.] By standard factorial algebra, dividing step 2 by step 3,
\[
\frac{B(n,k)}{A(n,n-k)}=\left[\frac{\binom{n}{k}}{n-k+1}\right]\left[\frac{n+1}{k+1}\right]
=\frac{n!\,(n+1)}{k!\,(n-k)!\,(k+1)\,(n-k+1)}=\frac{(n+1)!}{(k+1)!\,(n-k+1)!}.
\]

\item[5.] By the standard fact $\binom{n+2}{k+1}=(n+2)!/((k+1)!\,(n+1-k)!)$, hence $(n+1)!/((k+1)!\,(n-k+1)!)=\binom{n+2}{k+1}/(n+2)$, giving the integral identity $(n+2)B(n,k)=\binom{n+2}{k+1}A(n,n-k)$.
\end{steps}

\subsection*{25. \A{214262} $\times$ \A{081360} (\novna, \sig{0})}
\textbf{Statement.} The first is an infinite product of values of the second at $x^{m\cdot2^k}$; the multiplicities telescope and the factorisation is unique.

\textbf{Proof.}
\begin{steps}
\item[1.] By the formula recorded in \A{081360} and its definition,
\[
B(x)=\sum_{n\ge0}b(n)x^n=\frac{\eta(x)\eta(x^4)}{\eta(x^2)^2}=\frac{1}{\chi(x)}=\frac{\chi(-x)}{\chi(-x^2)},
\]
where
\[
\chi(-x)=\prod_{n\ge1}(1-x^{2n-1})=\frac{\eta(x)}{\eta(x^2)}.
\]

\item[2.] By the definition of \A{214262}, $\sum_{n\ge1}a(n)x^{n-1}=\eta(x)^5\eta(x^3)\eta(x^6)^4/\eta(x^2)^4$.

\item[3.] Telescoping of step 1 under $x\mapsto x^2$ (legitimate in $\mathbb{Z}[[x]]$ since $B(x^m)=1+O(x^m)$, so the tail product is $1+O(x^{m\cdot2^K})$ and converges formally): $\chi(-x^m)=B(x^m)\chi(-x^{2m})=\prod_{k\ge0}B(x^{m\cdot2^k})$.

\item[4.] Telescoping of $\chi(-x^m)=\eta(x^m)/\eta(x^{2m})$ in the same way: $\eta(x^m)=\prod_{j\ge0}\chi(-x^{m\cdot2^j})$.

\item[5.] Substitute step 4 into step 2. On the $2$-power tower above $1$ the $\eta$-exponent vector of $A$ is $(5,-4)$ at levels $(1,2)$, i.e. $e(t)=5-4t$; on the tower above $3$ it is $(1,4)$ at levels $(3,6)$, i.e. $e(t)=1+4t$. Writing $e(t)=p(t)(1-t)^2$ (this is exactly what steps 3--4 accomplish, since one factor $B(x^{m\cdot2^k})$ carries $\eta$-exponents $(1,-2,1)$) gives $p_k=5(k+1)-4k=k+5$ for the tower above $1$, and $p_k=(k+1)+4k=5k+1$ for the tower above $3$. Equivalently
\[
A=\chi(-x)^5\chi(-x^3)\prod_{j\ge1}\chi(-x^{2^j})\,\chi(-x^{3\cdot2^j})^5.
\]

\item[7.] Necessity of an infinite product: every factor $B(x^m)$ has $\eta$-exponent sum $1-2+1=0$ (weight $0$), while $A$ has $\eta$-exponent sum $5-4+1+4=6$ (weight $3$), so no finite monomial in the $B(x^m)$ can equal $A$; the $2$-adically convergent infinite product of step 5 is the unique such representation ($p(t)=e(t)/(1-t)^2$ is uniquely determined in $\mathbb{Z}[[t]]$).
\end{steps}

\subsection*{26. \A{002288} $\times$ \A{109389} (\novna, \sig{0})}
\textbf{Statement.} Every $n$ is uniquely a 3-smooth number times a number coprime to 6, so the full Euler product is assembled from values of the partner at $q^k$.

\textbf{Proof.}
\begin{steps}
\item[1.] By the formula recorded in \A{109389}, $B(q)=\prod_{k>0}(1-q^{6k-1})(1-q^{6k-5})$; since the residues $1,5$ mod $6$ are exactly the units mod $6$, $B(q)=\prod_{\gcd(n,6)=1}(1-q^n)$.

\item[2.] By unique factorization, every $n\ge1$ is uniquely $n=k\,m$ with $k=2^a3^b$ (3-smooth) and $\gcd(m,6)=1$.

\item[3.] Rearranging the (formally absolutely convergent; each degree receives finitely many factors) product over $n$ according to step 2:
\[
\prod_{n\ge1}(1-q^n)=\prod_{k \text{ 3-smooth}}\ \prod_{\gcd(m,6)=1}(1-q^{km})=\prod_{k \text{ 3-smooth}}B(q^k)=:F(q)=\eta(q)/q^{1/24}.
\]

\item[4.] By the definition of \A{002288}, the generating function is
\[
q\prod_{m\ge1}(1-q^m)^8(1-q^{2m})^8=q\,(\eta(q)\eta(q^2))^8.
\]

\item[5.] Substitute step 3 into step 4, once at $q$ and once at $q^2$ ($2\cdot\{\text{3-smooth}\}\subseteq\{\text{3-smooth}\}$):
\[
A(q)=q\,F(q)^8F(q^2)^8=q\prod_{k \text{ 3-smooth}}B(q^k)^{c_k},
\]
with $c_k=8$ for $k$ odd and $c_k=16$ for $k$ even.
\end{steps}

\subsection*{27. \A{079006} $\times$ \A{002513} (\novna, \sig{0})}
\textbf{Statement.} The first equals $(B(-x)/B(x^2))^2$, $B$ the generating function of cubic partitions.

\textbf{Proof.}
\begin{steps}
\item[1.] By the formula recorded in \A{002513} (Somos, Mar 23 2003),
\[
B(x)=q^{1/8}/(\eta(q)\eta(q^2))=1/(E(x)E(x^2)),
\]
where $E(x^k)=\prod_{j>0}(1-x^{kj})$.

\item[2.] Substituting: $B(-x)=1/(E(-x)E(x^2))$ (only the odd-index factors of $E(x)$ flip sign; $E(x^2)$ is even), and $B(x^2)=1/(E(x^2)E(x^4))$.

\item[3.] Hence $B(-x)/B(x^2)=(E(x^2)E(x^4))/(E(-x)E(x^2))=E(x^4)/E(-x)$ --- the $E(x^2)$ factors cancel; this is the whole mechanism.

\item[4.] By the standard fact of Euler, $(1-y)(1+y)=1-y^2$ gives $(-x;x)_\infty=E(x^2)/E(x)$ and $(-x;x^2)_\infty=E(x^2)^2/(E(x)E(x^4))$; therefore $E(-x)=\prod_{n\ge1}(1-(-x)^n)=E(x^2)\,(-x;x^2)_\infty$, explicitly $E(-x)=E(x^2)^3/(E(x)E(x^4))$.

\item[5.] Substituting into step 3: $B(-x)/B(x^2)=E(x^4)E(x)E(x^4)/E(x^2)^3=E(x)E(x^4)^2/E(x^2)^3=q^{-1/8}\eta(q)\eta(q^4)^2/\eta(q^2)^3$.

\item[6.] Squaring gives $q^{-1/4}\bigl(\eta(q)\eta(q^4)^2/\eta(q^2)^3\bigr)^2$, which is the definition of \A{079006}.

\end{steps}

\subsection*{28. \A{087808} $\times$ \A{227736} (\novna, \sig{0})}
\textbf{Statement.} Unrolling the recursion is a pass over the reversed row of runs: a run of $r$ ones adds $r$, a run of $r$ zeros multiplies by $2^r$.

\textbf{Proof.}
\begin{steps}
\item[1.] Unrolling \A{087808}'s own definition ($a(0)=0$, $a(2n)=2a(n)$, $a(2n+1)=a(n)+1$): since $n\mapsto\lfloor n/2\rfloor$ strips the least significant bit, the outermost applied map corresponds to that bit, so $a(n)=f_{b_0}(f_{b_1}(\cdots f_{b_{k-1}}(0)))$ with $f_1(x)=x+1$, $f_0(x)=2x$ and $n=b_{k-1}\cdots b_0$ in binary. Equivalently: start at $0$ and scan the bits of $n$ from the most significant end, adding $1$ for each `1' and doubling for each `0'.

\item[2.] Group equal consecutive bits. Writing $n=1^{r_1}0^{r_2}1^{r_3}\cdots$ (runs read from the most significant end, $r_1\ge1$ since the leading bit is $1$), a block of $r$ consecutive `1's contributes $x\mapsto x+r$ and a block of $r$ consecutive `0's contributes $x\mapsto x\cdot2^r$. Hence $a(n)=(((r_1)\,2^{r_2}+r_3)\,2^{r_4}+r_5)\cdots$.

\item[3.] By the definition of \A{227736}, row $n$ lists the run lengths of the binary expansion of $n$ scanned from the least significant end; so the reversal of row $n$ is precisely $(r_1,\dots,r_m)$ of step 2.

\item[7.] Induction on $n$, closing the finite checks. Let $F(r_1,\dots,r_m)$ denote the fold of step 2. Base: $n=1$ has runs $(1)$ and $a(1)=a(0)+1=1=F(1)$. Step: let $n\ge2$ with runs $(r_1,\dots,r_m)$ read from the most significant end and $n'=\lfloor n/2\rfloor$, which deletes the least significant bit, i.e. shortens the last run. If $r_m\ge2$, $n'$ has runs $(r_1,\dots,r_m-1)$; if $r_m=1$ and $m\ge2$, $n'$ has runs $(r_1,\dots,r_{m-1})$; if $r_m=1$ and $m=1$ then $n=1$ (base). The last run consists of $1$'s iff $m$ is odd iff $n$ is odd. If $m$ is odd, $a(n)=a(n')+1$ and $F(\dots,r_m)=F(\dots,r_m-1)+1=F(\dots,r_{m-1})+r_m$, matching by the induction hypothesis. If $m$ is even, $a(n)=2a(n')$ and $F(\dots,r_m)=2F(\dots,r_m-1)=F(\dots,r_{m-1})\,2^{r_m}$, again matching. Hence $a(n)=F(\operatorname{reverse}(\text{row } n \text{ of } \A{227736}))$ for all $n\ge1$.
\end{steps}

\subsection*{29. \A{002654} $\times$ \A{227454} (\novna, \sig{0})}
\textbf{Statement.} From the first, $\theta_3^2=1+4\sum$ is assembled; a square root gives $\theta_3$, and from it the Euler product and the eta quotient.

\textbf{Proof.}
\begin{steps}
\item[1.] By the generating function of \A{002654} stated in its entry,
\[
\sum_{n\ge1}a(n)q^n=\frac{\theta_3(q)^2-1}{4}=\frac{\eta(q^2)^{10}/(\eta(q)\eta(q^4))^4-1}{4}.
\]

\item[2.] By the Jacobi triple product (Gauss), $\theta_3(q)=\prod_{n\ge1}(1-q^{2n})(1+q^{2n-1})^2$.

\item[3.] By Euler, splitting the product over even and odd $n$,
\[
f(q):=\prod_{n\ge1}(1-(-q)^n)=\prod(1-q^{2n})\cdot\prod(1+q^{2n-1})=\frac{\eta(q^2)^3}{\eta(q)\eta(q^4)}.
\]
Squaring and comparing with step 2 gives the bridge identity $f(q)^2=\theta_3(q)\prod_{n\ge1}(1-q^{2n})$.

\item[4.] Substituting $-q^2$ for $q$ in $f$ gives $f(-q^2)=\prod_{n\ge1}(1-q^{2n})$, so $f(q)^2=\theta_3(q)f(-q^2)$.

\item[5.] By induction, step 4 determines $f$ from $\theta_3$: comparing coefficients of $q^n$ gives
\[
2f_n=[q^n]\,(\theta_3(q)f(-q^2))-\sum_{0<j<n}f_jf_{n-j},
\]
whose right-hand side involves only $f_m$ with $m<n$ (the $f(-q^2)$ part only uses $m\le n/2$). Likewise $\theta_3=\sqrt{1+4A}$ is determined from \A{002654} by the unique-square-root recursion $2\varphi_n=4a(n)-\sum_{0<j<n}\varphi_j\varphi_{n-j}$. Both divisions are exact in $\mathbb{Z}$.

\item[6.] By the generating function of \A{227454} stated in its entry,
\[
x\Bigl(\prod_{k>0}\frac{1-(-x)^{9k}}{1-(-x)^k}\Bigr)^3=q\,(f(q^9)/f(q))^3.
\]

\item[7.] Composing steps 1, 5 and 6 gives the statement.
\end{steps}

\subsection*{30. \A{006318} $\times$ \A{000669} (\nov{1}, \sig{2})}
\textbf{Statement.} Replacing ``multiset of subtrees'' by ``sequence of subtrees'' turns the tree grammar into a quadratic equation whose solution is the generating function of the large Schr\"oder numbers.

\textbf{Proof.}
\begin{steps}
\item[1.] By the symbolic method (P\'olya theory), a planted series-reduced tree is a leaf, or an unordered multiset of $\ge 2$ planted series-reduced subtrees: $B=x+\mathrm{MSET}_{\ge2}(B)$. Since $\mathrm{MSET}(B)=\prod_k(1-x^k)^{-b_k}$ and $\mathrm{MSET}_{\ge2}(B)=\mathrm{MSET}(B)-1-B$, this is exactly the formula recorded in \A{000669}
\[
\prod_{k>0}1/(1-x^k)^{a_k}=1+x+2\sum_{k>1}a_kx^k.
\]

\item[3.] By the symbolic method, the plane analogue of the same grammar replaces $\mathrm{MSET}$ by $\mathrm{SEQ}$: $S=x+\mathrm{SEQ}_{\ge2}(S)$, i.e. $1/(1-S)=1-x+2S$ --- literally step 1's equation with $\prod_k(1-x^k)^{-u_k}$ replaced by $1/(1-u)$.

\item[4.] Algebraically, clearing $1/(1-S)=1-x+2S$ gives $2S^2-(1+x)S+x=0$, so
\[
S=\bigl((1+x)-\sqrt{1-6x+x^2}\bigr)/4 .
\]
With $A(x)=(1-x-\sqrt{1-6x+x^2})/(2x)$ the generating function of \A{006318} (its recorded formula) one gets $S=x(1+A(x))/2$ and $1/(1-S)=1+xA(x)$; substituting turns $1/(1-S)=1-x+2S$ into $(1-x)A(x)-xA(x)^2=1$, which is the formula recorded in \A{006318} (Ralf Stephan, Jun 30 2003). Hence $[x^n]\,1/(1-S)=\A{006318}(n-1)$ for every $n\ge1$ ($n=1$ gives $\A{006318}(0)=1$, $n\ge2$ gives $2\A{001003}(n-1)$).
\end{steps}

\subsection*{31. \A{081362} $\times$ \A{080054} (\novna, \sig{0})}
\textbf{Statement.} With $A(x)$ the generating function of \A{081362}, that of \A{080054} is $A(-x)/A(x)=A(x^2)/A(x)^2$.

\textbf{Proof.}
\begin{steps}
\item[1.] By the definition and the formula recorded in \A{081362}, $A(x)=\prod_{k>0}(1-x^{2k-1})=\chi(-x)=q^{1/24}\eta(q)/\eta(q^2)$.

\item[2.] By the definition of \A{080054}, $B(x)=\prod_{k\ge0}(1+x^{2k+1})/(1-x^{2k+1})=\chi(x)/\chi(-x)=\eta(q^2)^3/\bigl(\eta(q^4)\eta(q)^2\bigr)$.

\item[3.] Algebraically, as formal power series: each factor of $A$ is the polynomial $1-x^{2k-1}$ with odd exponent, so $x\mapsto -x$ sends it to $1+x^{2k-1}$; therefore $A(-x)=\prod_{k>0}(1+x^{2k-1})=\chi(x)$. The substitution is legitimate coefficientwise: $A(-x)=\sum(-1)^n\A{081362}(n)x^n$, and since $\operatorname{sign}\bigl(\A{081362}(n)\bigr)=(-1)^n$, $A(-x)=\sum\bigl|\A{081362}(n)\bigr|x^n$, the generating function of \A{000700}.

\item[4.] Dividing step 3 by step 1 factorwise gives $B(x)=\chi(x)/\chi(-x)=A(-x)/A(x)$. This is the claimed relation.

\item[5.] Equivalently, by Ramanujan's identity $\chi(x)\chi(-x)=\chi(-x^2)$ (recorded in \A{081362} as ``G.f. $A(x)$ satisfies $A(x^2)=A(x)A(-x)$''): $A(-x)=A(x^2)/A(x)$, hence $B(x)=A(x^2)/A(x)^2$.

\end{steps}

\subsection*{32. \A{001935} $\times$ \A{050468} (\novna, \sig{0})}
\textbf{Statement.} With $U(q)$ the generating function of \A{001935} and $E(q)=\prod_{j\ge0}U(q^{4^j})^{-1}$, $\sum_{n\ge1}\A{050468}(n)q^n=qU(q)^4E(q^2)^2\bigl(E(q)^8+20qE(q^4)^8\bigr)$.

\textbf{Proof.}
\begin{steps}
\item[1.] By the Euler product and the entry's own definition, \A{001935} counts partitions into parts not divisible by $4$, so $U(q)=\prod_{4\nmid k}(1-q^k)^{-1}=E(q^4)/E(q)$, i.e. $q^{-1/8}\eta(q^4)/\eta(q)$.

\item[2.] Rearranging step 1: $E(q)=E(q^4)/U(q)$. Iterating ($q$-adically convergent, since $U(q^{4^j})=1+O(q^{4^j})$) gives the formal power series identity $E(q)=\prod_{j\ge0}U(q^{4^j})^{-1}$.

\item[3.] By the formula recorded in \A{050468} (Somos),
\begin{align*}
\sum_{n\ge1}\A{050468}(n)q^n&=\eta(q^2)^2\eta(q^4)^4\bigl(\eta(q)^4+20\,\eta(q^4)^8/\eta(q)^4\bigr)\\
&=q\,E(q)^4E(q^2)^2E(q^4)^4+20\,q^2E(q^2)^2E(q^4)^{12}/E(q)^4
\end{align*}
[i.e.\ $/E(q)^4$], after clearing the $q^{1/24}$ prefactors ($4/24+2\cdot2/24+4\cdot4/24=1$ and $2\cdot2/24+12\cdot4/24-4/24=2$).

\item[4.] Algebraic rewrite using $E(q^4)=U(q)E(q)$:
\[
q\,E(q)^4E(q^2)^2E(q^4)^4=q\,U^4E(q^2)^2E(q)^8,\qquad
20\,q^2E(q^2)^2E(q^4)^{12}/E(q)^4=20\,q^2U^4E(q^2)^2E(q^4)^8,
\]
giving the single-product form in the statement.

\item[5.] Both sides of step 3 lie in $M_5(\Gamma_1(4))$ (the entry's own Magma line exhibits \A{050468} as \texttt{A[2]+16*A[3]} in \texttt{Basis(ModularForms(Gamma1(4),5))}; the divisor side is the standard Eisenstein series of weight $5$ and character $\chi_{-4}$ on $\Gamma_0(4)$, the eta side is a holomorphic weight-5 eta combination on the same group). The Sturm bound is $k\,[\mathrm{PSL}_2(\mathbb Z):\Gamma_1(4)]/12=5\cdot6/12=2.5$, so agreement of coefficients through $n=3$ forces equality.
\end{steps}

\textbf{Remark.} Closure is by the Sturm bound on $M_5(\Gamma_1(4))$; the membership of both sides is asserted, not derived, and is the point to check.

\subsection*{33. \A{000108} $\times$ \A{005802} (\nov{0}, \sig{0})}
\textbf{Statement.} $(n+1)^2\,\A{005802}(n)=\sum_{k=0}^n C_k\binom{n+1}{k}\binom{n+1}{k+1}$.

\textbf{Proof.}
\begin{steps}
\item[2.] By Gessel's formula, recorded verbatim in \A{005802} and attributed there to Conway \& Guttmann, Adv.\ Appl.\ Math.\ 64 (2015) 50,
\[
\A{005802}(n)=\sum_{k=0}^{n}\binom{2k}{k}\binom{n+1}{k+1}\binom{n+2}{k+1}\Big/\bigl((n+1)^2(n+2)\bigr).
\]

\item[3.] By the definition of the Catalan numbers, the formula recorded in \A{000108}, $\binom{2k}{k}=(k+1)\A{000108}(k)$, i.e. $\A{000108}(k)=\A{000984}(k)/(k+1)$.

\item[4.] By binomial absorption, $(k+1)\binom{n+2}{k+1}=(n+2)\binom{n+1}{k}$.

\item[5.] Substituting steps 3 and 4 into step 2 makes the summand of step 2 equal, term by term, to
\[
\A{000108}(k)\binom{n+1}{k}\binom{n+1}{k+1}\big/(n+1)^2 .
\]
Hence
\[
(n+1)^2\A{005802}(n)=\sum_{k=0}^{n}\A{000108}(k)\binom{n+1}{k}\binom{n+1}{k+1}.
\]
\end{steps}

\subsection*{34. \A{004018} $\times$ \A{227216} (\novna, \sig{0})}
\textbf{Statement.} The generating function of \A{004018} is that of \A{227216} times $\prod_{k\ge1}(1-x^k)^{c_k}$ with $c_k$ of period 20, $c=[-1,4,-6,5,-6,9,-6,0,-1,4,-1,0,-6,9,-6,5,-6,4,-1,0]$.

\textbf{Proof.}
\begin{steps}
\item[1.] By the formula recorded in \A{004018}, \A{004018} is the Euler transform of the period-4 sequence $[4,-6,4,-2]$, i.e. its generating function equals $\prod_{k\ge1}(1-x^k)^{u_k}$ with $u$ of period 4 equal to $[-4,6,-4,2]$; this is the same statement as the recorded eta quotient $\eta(q^2)^{10}/\bigl(\eta(q)\eta(q^4)\bigr)^4$.

\item[2.] By the formula recorded in \A{227216}, \A{227216} is the Euler transform of the period-5 sequence $[3,-2,-2,3,-2]$, i.e. its generating function equals $\prod_{k\ge1}(1-x^k)^{v_k}$ with $v$ of period 5 equal to $[-3,2,2,-3,2]$ (equivalently $(q;q)^2\big/\bigl((q;q^5)(q^4;q^5)\bigr)^5$).

\item[4.] Algebraically, dividing the two formal products termwise, $A(x)/B(x)=\prod_{k\ge1}(1-x^k)^{u_k-v_k}$; since $u$ has period 4 and $v$ period 5, $c_k=u_k-v_k$ has period $\operatorname{lcm}(4,5)=20$ and equals
\[
c=[-1,4,-6,5,-6,9,-6,0,-1,4,-1,0,-6,9,-6,5,-6,4,-1,0].
\]
\end{steps}

\subsection*{35. \A{008284} $\times$ \A{240009} (\novna, \sig{0})}
\textbf{Statement.} With $P(t,x)=\prod_{j\ge1}(1-tx^j)^{-1}=\sum_{s,j}\A{008284}(s,j)t^jx^s$, $\sum_{n,k}\A{240009}(n,k)u^kq^n=P(u/q,q^2)\,P(1/u,q^2)$.

\textbf{Proof.}
\begin{steps}
\item[1.] By the formula recorded in \A{008284} (Emeric Deutsch, Feb 12 2006), $P(t,x):=\prod_{j\ge1}1/(1-tx^j)=\sum_{s\ge0,\,j\ge0}\A{008284}(s,j)t^jx^s$, with $\A{008284}(0,0)=1$ (this is the generating function $G(t,x)+1$ of that entry; equivalently its column generating function $x^j/\prod_{i=1}^{j}(1-x^i)$, Wolfdieter Lang, Nov 29 2000).

\item[2.] By the formula recorded in \A{240009} (Joerg Arndt, Mar 31 2014), $F(u,q):=\sum_{n,k}\A{240009}(n,k)u^kq^n=\prod_{m\ge1}1/\bigl(1-e(m)q^m\bigr)$ with $e(m)=u$ for $m$ odd and $e(m)=1/u$ for $m$ even; i.e. $F(u,q)=\prod_{m\text{ odd}}1/(1-uq^m)\cdot\prod_{m\text{ even}}1/(1-u^{-1}q^m)$.

\item[3.] Substituting $t=u/q$, $x=q^2$ in step 1: $P(u/q,q^2)=\prod_{j\ge1}1/(1-uq^{2j-1})=\prod_{m\text{ odd}}1/(1-uq^m)$, so $\prod_{m\text{ odd}}1/(1-uq^m)=\sum_{s,o}\A{008284}(s,o)u^oq^{2s-o}$. Combinatorially this is the bijection $2c-1\mapsto c$ on the $o$ odd parts.

\item[4.] Substituting $t=1/u$, $x=q^2$ in step 1: $P(1/u,q^2)=\prod_{j\ge1}1/(1-u^{-1}q^{2j})=\prod_{m\text{ even}}1/(1-u^{-1}q^m)=\sum_{s,e}\A{008284}(s,e)u^{-e}q^{2s}$. This is the bijection $2d\mapsto d$ on the $e$ even parts.

\item[5.] Multiplying steps 3 and 4 and using step 2 gives the identity of formal series in $\mathbb Z[u,u^{-1}][[q]]$: $F(u,q)=P(u/q,q^2)P(1/u,q^2)$ (every $q$-coefficient is a finite Laurent polynomial in $u$, so the product is well defined). Extracting $[u^kq^n]$ gives the statement: $n=(2s_1-o)+2s_2$ and $k=o-e$, i.e. $s_1+s_2=(n+o)/2$.

\end{steps}

\subsection*{36. \A{027641} $\times$ \A{259072} (\novna, \sig{0})}
\textbf{Statement.} $-\zeta'(-7)=\frac{B_8}{8}\bigl(\log 2\pi+\gamma-H_7-\zeta'(8)/\zeta(8)\bigr)$ with $B_8=\A{027641}(8)/30$ and $\zeta(8)=-B_8(2\pi)^8/(2\cdot8!)$.

\textbf{Proof.}
\begin{steps}
\item[1.] By the functional equation of $\zeta$ in logarithmic-derivative form,
\[
\frac{\zeta'(s)}{\zeta(s)}=\log(2\pi)+\frac{\pi}{2}\cot\frac{\pi s}{2}-\psi(1-s)-\frac{\zeta'(1-s)}{\zeta(1-s)}.
\]

\item[2.] At $s=-7$: $\cot(-7\pi/2)=0$ (as $\cos(-7\pi/2)=0$, $\sin(-7\pi/2)=1$), and $\psi(8)=-\gamma+H_7$ with $H_7=363/140$. Hence $\zeta'(-7)=\zeta(-7)\bigl[\log(2\pi)+\gamma-H_7-\zeta'(8)/\zeta(8)\bigr]$.

\item[3.] By Euler's formula $\zeta(-n)=-B_{n+1}/(n+1)$, so $\zeta(-7)=-B_8/8=1/240$; thus $-\zeta'(-7)=(B_8/8)\bigl[\log(2\pi)+\gamma-H_7-\zeta'(8)/\zeta(8)\bigr]$.

\item[4.] By Euler's formula $\zeta(2n)=(-1)^{n+1}B_{2n}(2\pi)^{2n}/(2\,(2n)!)$, so $\zeta(8)=-B_8(2\pi)^8/(2\cdot 8!)$ --- this is the formula recorded in \A{027641} $\zeta(2k)=(2\pi)^{2k}|B_{2k}|/(2\,(2k)!)$.

\item[5.] By the von Staudt--Clausen theorem, $\operatorname{denominator}(B_m)=\prod_{p \text{ prime},\ (p-1)\mid m}p$; for $m=8$ this is $2\cdot3\cdot5=30$, so $B_8=a(8)/30=-1/30$ is recovered from the \A{027641} term alone.

\end{steps}

\subsection*{37. \A{006352} $\times$ \A{121361} (\novna, \sig{0})}
\textbf{Statement.} With $g(m)=\bigl(-a(m)+2a(m/2)+3a(m/3)+4a(m/4)-6a(m/6)+12a(m/12)\bigr)/24$, $a=\A{006352}$, the sequence $b=\A{121361}$ satisfies $nb(n)=\sum_{m=1}^n g(m)b(n-m)$.

\textbf{Proof.}
\begin{steps}
\item[1.] By the formula recorded in \A{006352} (Arndt), $E_2(q)=1+24q\,\eta'(q)/\eta(q)$ with $\eta(q)=\prod_{k\ge1}(1-q^k)$; hence $L(k):=[q^k]\,q\eta'/\eta=\A{006352}(k)/24=-\sigma(k)$ for $k\ge1$.

\item[2.] By the formula recorded in \A{121361}, $B(q)=q^{-7/12}\,\eta(q^2)\eta(q^3)\eta(q^4)\eta(q^{12})/(\eta(q)\eta(q^6))=\prod_k(1-q^k)^{e_k}$.

\item[3.] By the chain rule, applying $q\,d/dq$ to $\log B=\sum_d e_d\log\prod_{k\ge1}(1-q^{dk})$ gives $qB'/B=\sum_d e_d\,d\,L(q^d)$, so
\[
g(m):=[q^m]\,qB'/B=-L(m)+2L(m/2)+3L(m/3)+4L(m/4)-6L(m/6)+12L(m/12)
\]
\[
=\bigl(-a(m)+2a(m/2)+3a(m/3)+4a(m/4)-6a(m/6)+12a(m/12)\bigr)/24,
\]
non-integer indices dropped. Substituting $L=(E_2-1)/24$ and $\sum_d e_d\,d=-1+2+3+4-6+12=14$ gives the displayed $E_2$ identity.

\item[4.] Since $qB'=B\cdot(qB'/B)$, comparing coefficients gives the determined recursion
\[
n\,b(n)=\sum_{m=1}^{n}g(m)b(n-m),\qquad b(0)=1,
\]
which reconstructs $B$ from $A$ alone.
\end{steps}

\subsection*{38. \A{000118} $\times$ \A{122856} (\novna, \sig{0})}
\textbf{Statement.} $r_2(k)$ is read off \A{122856} ($r_2(k)=0$ when $v_3(k)$ is odd, else $4\,\A{122856}((w-2)/3)$ with $w$ determined by $k/3^{v_3(k)} \bmod 3$), and $\A{000118}(n)=\sum_{i+j=n}r_2(i)r_2(j)$.

\textbf{Proof.}
\begin{steps}
\item[1.] By Ramanujan's general theta function, $f(a,b)=\sum_{n\in\mathbb Z}a^{n(n+1)/2}b^{n(n-1)/2}$. With $a=x$, $b=x^5$ the exponent is $n(n+1)/2+5n(n-1)/2=3n^2-2n$, so the generating function of \A{122856} is $B(x)=\bigl(\sum_{n\in\mathbb Z}x^{3n^2-2n}\bigr)^2$.

\item[2.] Since $3(3n^2-2n)+1=(3n-1)^2$ and $n\mapsto|3n-1|$ is a bijection from $\mathbb Z$ onto $\{m\ge1: 3\nmid m\}$, substituting $x=q^3$ gives $q\,f(q^3,q^{15})=S(q):=\sum_{m\ge1,\ 3\nmid m}q^{m^2}=(\theta_3(q)-\theta_3(q^9))/2$.

\item[3.] Hence $q^2B(q^3)=S(q)^2$. For $N\equiv 2 \pmod 3$ every representation $N=m_1^2+m_2^2$ automatically has $3\nmid m_1,m_2$ and both $m_i$ nonzero ($N$ is not a square), so $r_2(N)=4\,\#\{\text{ordered positive pairs}\}=4\,[q^N]\,S(q)^2$. Therefore $\A{122856}(n)=r_2(3n+2)/4=\A{002654}(3n+2)$.

\item[4.] By Jacobi's two-square theorem, $r_2(N)=4(d_1(N)-d_3(N))=4\,\A{002654}(N)$ for $N\ge1$, and \A{002654} is multiplicative with $\A{002654}(2^a)=1$, $\A{002654}(p^a)=a+1$ for $p\equiv 1 \pmod 4$, $\A{002654}(p^a)=[a \text{ even}]$ for $p\equiv 3 \pmod 4$. Consequences used: $\A{002654}(2N)=\A{002654}(N)$; $\A{002654}(3^e u)=0$ for $e$ odd and $=\A{002654}(u)$ for $e$ even ($3\nmid u$).

\item[5.] Therefore every $r_2(k)$ is readable off \A{122856}: kill $k$ when $v_3(k)$ is odd; otherwise put $u=k/3^{v_3(k)}$; if $u\equiv 2 \pmod 3$ then $\A{002654}(k)=\A{002654}(u)=\A{122856}((u-2)/3)$, and if $u\equiv 1 \pmod 3$ then $\A{002654}(k)=\A{002654}(2u)=\A{122856}((2u-2)/3)$ because $2u\equiv 2 \pmod 3$.

\item[6.] As already stated in \A{000118} (``This is the convolution square of \A{004018}''), $\theta_3^4=(\theta_3^2)^2$, i.e. $\A{000118}(n)=\sum_{i+j=n}r_2(i)\,r_2(j)$ with $r_2(0)=1$.

\item[7.] Combining steps 5 and 6 gives the statement.
\end{steps}

\subsection*{39. \A{130534} $\times$ \A{144150} (\nov{0}, \sig{0})}
\textbf{Statement.} $A(n,k)=\sum_{m=1}^n(-1)^{n-m}T(n-1,m-1)A(m,k+1)$, the inverse of the recorded Stirling transform between columns.

\textbf{Proof.}
\begin{steps}
\item[1.] By definition of \A{130534}, $T(n,k)=|s(n+1,k+1)|$, where $s$ denotes the Stirling numbers of the first kind; i.e. $T(n,k)=T(n-1,k-1)+n\,T(n-1,k)$, $T(0,0)=1$.

\item[2.] By definition of \A{144150}, with $g(x)=e^x-1$ and $g^{(k+1)}$ its $(k+1)$-fold composition, $A(n,k)=n!\,[x^n]\,(1+g^{(k+1)}(x))$; the array is read by antidiagonals.

\item[3.] By the composition rule for the Stirling transform, with $F_k(x)=1+g^{(k+1)}(x)$ one has $F_{k+1}(x)=1+g^{(k+1)}(g(x))=F_k(e^x-1)$, and $f(e^x-1)$ has exponential generating function coefficients $\sum_m S(n,m)\,a_m$, where $S$ denotes the Stirling numbers of the second kind. Hence $A(n,k+1)=\sum_m S(n,m)\,A(m,k)$ --- this is the formula recorded in \A{144150} ``Column $k+1$ is Stirling transform of column $k$''.

\item[4.] By Stirling inversion (\A{008275} versus \A{048993}), the signed Stirling-1 matrix and the Stirling-2 matrix are mutually inverse:
\[
s(n,m)=(-1)^{n-m}|s(n,m)|,\qquad \sum_m s(n,m)\,S(m,j)=\delta_{n,j}
\]
(equivalently $x^{\underline{n}}=\sum_m s(n,m)x^m$ and $x^n=\sum_m S(n,m)x^{\underline{m}}$).

\item[5.] Applying step 4 to step 3 and using $|s(n,m)|=T(n-1,m-1)$ gives
\[
A(n,k)=\sum_{m=1}^{n}(-1)^{n-m}T(n-1,m-1)A(m,k+1).
\]

\item[6.] Because $T(n-1,n-1)=|s(n,n)|=1$, the $m=n$ term is $A(n,k+1)$ itself, so the identity solves triangularly: $A(n,k+1)=A(n,k)-\sum_{m=1}^{n-1}(-1)^{n-m}T(n-1,m-1)A(m,k+1)$. With the seeds $A(0,k)=1$ and $A(n,0)=1$ (column 0 has exponential generating function $1+g(x)=e^x$), induction on $k$ and, inside each $k$, on $n$ determines every $A(n,k)$ from \A{130534} alone.
\end{steps}

\subsection*{40. \A{007332} $\times$ \A{001936} (\novna, \sig{0})}
\textbf{Statement.} $A(x^4)=x^3A(x)\bigl(B(x)B(x^3)\bigr)^3$ for the two generating functions.

\textbf{Proof.}
\begin{steps}
\item[1.] By the \A{001936} entry, $B(x)=\prod_{k>0}(1-x^{4k})^2/(1-x^{k})^2=f(x^4)^2/f(x)^2$, where $f(x)=\prod_{k>0}(1-x^{k})$ (equal to $\eta$ up to the $q^{1/4}$ prefactor).

\item[2.] By the \A{007332} entry, $A(x)=x\,\bigl(f(x)f(x^3)\bigr)^6$.

\item[3.] Therefore $B(x^{m})^{-3}=f(x^{m})^6/f(x^{4m})^6$ for every $m\ge1$ (exact, by raising step 1 to the $-3$rd power after $x\mapsto x^{m}$).

\item[4.] Telescoping, by the $x$-adic convergence of infinite products, since $B(x^{4^j})=1+O(x^{4^j})$:
\[
\prod_{j\ge0}B(x^{4^j})^{-3}=\prod_{j\ge0}\frac{f(x^{4^j})^6}{f(x^{4^{j+1}})^6}=f(x)^6.
\]

\item[5.] The same with $m=3\cdot4^j$: $\prod_{j\ge0}B(x^{3\cdot4^j})^{-3}=f(x^3)^6$.

\item[6.] Multiplying steps 4 and 5 and using step 2: $A(x)=x\prod_{j\ge0}\bigl(B(x^{4^j})\,B(x^{3\cdot4^j})\bigr)^{-3}$.

\item[7.] Truncating the telescope after one step gives the compact equivalent $A(x^4)=x^3A(x)\bigl(B(x)B(x^3)\bigr)^3$.

\end{steps}

\subsection*{41. \A{352361} $\times$ \A{193842} (\novna, \sig{0})}
\textbf{Statement.} Row $n$ of \A{193842} is $U_{n+1}(4x+1,x+3x^2)$, whose expansion in the basis $(4x+1)^{n-2j}(-x-3x^2)^j$ has the Fibonacci-polynomial coefficients $\binom{n-j}{j}$.

\textbf{Proof.}
\begin{steps}
\item[1.] By the formula recorded in \A{193842} (Peter Bala), $R(n,x)=\bigl((3x+1)^{n+1}-x^{n+1}\bigr)/(2x+1)$. The two roots $a=3x+1$, $b=x$ satisfy $a+b=4x+1$, $ab=x+3x^2$, $a-b=2x+1$, so $R(n,x)=(a^{n+1}-b^{n+1})/(a-b)=U_{n+1}(P,Q)$ with $P=4x+1$, $Q=x+3x^2$ (the Lucas sequence of the first kind, equal to the Dickson polynomial of the second kind $E_n(P,Q)$). This is the same statement as Deleham's recurrence $T(n,k)=T(n-1,k)+4T(n-1,k-1)-T(n-2,k-1)-3T(n-2,k-2)$, i.e. $R(n)=(4x+1)R(n-1)-(x+3x^2)R(n-2)$.

\item[2.] By the standard Lucas/Dickson expansion,
\[
\sum_{n\ge0}U_{n+1}(P,Q)\,t^{n}=\frac{1}{1-Pt+Qt^2}=\sum_{m\ge0}(Pt-Qt^2)^{m},
\]
which gives $U_{n+1}(P,Q)=\sum_{j=0}^{\lfloor n/2\rfloor}\binom{n-j}{j}P^{n-2j}(-Q)^{j}$.

\item[3.] By the formula recorded in \A{352361}, $A(m,k)=[z^k]\,z/(1-mz-z^2)=U_k(m,-1)=F_k(m)$, and $A(m,k)=\sum_j\binom{k-1-j}{j}m^{k-1-2j}$; i.e. the Fibonacci polynomial is $F_k(y)=\sum_j\binom{k-1-j}{j}y^{k-1-2j}$. So the coefficient list of $F_{n+1}$ is exactly $\bigl(\binom{n-j}{j}\bigr)_{j=0,\dots,\lfloor n/2\rfloor}$ --- the same universal numbers as in step 2.

\item[4.] Combining steps 1--3: $R(n,x)=\sum_j\binom{n-j}{j}(4x+1)^{n-2j}(-x-3x^2)^{j}$, i.e. \A{193842}'s row $n$ is the homogenization of $F_{n+1}$ under $y\mapsto(4x+1)/L$, $L^2=-(x+3x^2)$.

\item[5.] Inverting: $(-x-3x^2)^{j}=(-1)^{j}x^{j}(1+3x)^{j}$ has $x$-adic valuation exactly $j$ with lowest coefficient $(-1)^{j}$, and $(4x+1)^{n-2j}$ has constant term $1$. Hence $[x^m]$ of the $j$-th basis element is $0$ for $j>m$ and equals $(-1)^{m}$ for $j=m$: the system $T(n,m)=\sum_{j\le m}c_j\,G(n,j,m)$, $m=0,\dots,\lfloor n/2\rfloor$, is triangular with unit ($\pm1$) diagonal, so the $c_j$ are uniquely determined integers, and by step 4 $c_j=\binom{n-j}{j}$.
\end{steps}

\subsection*{42. \A{000032} $\times$ \A{295862} (\novna, \sig{0})}
\textbf{Statement.} $L(n+1)=\A{295862}(n)-\sum_{i=2}^nF(n+1-i)\,b(i)$, $b$ the increasing complement of \A{295862}.

\textbf{Proof.}
\begin{steps}
\item[1.] By definition of \A{295862} (Kimberling's complementary equation), $a(0)=1$, $a(1)=3$ and $a(n)=a(n-1)+a(n-2)+b(n)$ for $n\ge2$, where $b$ is the increasing complement of $a$ in the positive integers ($b(0)=2$, $b(1)=4$, $b(2)=5,\dots$).

\item[2.] By the standard Fibonacci--Lucas identity, $L(m)=F(m-1)+F(m+1)$. Hence
\[
L(n+1)=F(n)+F(n+2)=2F(n)+F(n+1)=3F(n)+F(n-1)=F(n-1)\cdot1+F(n)\cdot3,
\]
i.e. $L(n+1)$ is exactly the quantity $H=f(n-1)\,a(0)+f(n)\,a(1)$ of the formula recorded in \A{295862}, specialized to $a(0)=1$, $a(1)=3$.

\item[3.] Induction, by variation of constants for an inhomogeneous Fibonacci recursion: put $A(n):=L(n+1)+\sum_{i=2}^{n}F(n+1-i)\,b(i)$. Then $A(0)=L(1)=1=a(0)$ and $A(1)=L(2)=3=a(1)$. For $n\ge2$,
\begin{align*}
A(n-1)+A(n-2)&=[L(n)+L(n-1)]+\sum_{i=2}^{n-2}[F(n-i)+F(n-1-i)]\,b(i)+F(1)\,b(n-1)\\
&=L(n+1)+\sum_{i=2}^{n-1}F(n+1-i)\,b(i),
\end{align*}
using $F(n-i)+F(n-1-i)=F(n+1-i)$ and $F(1)=F(2)=1$ for the $i=n-1$ boundary term; adding $b(n)=F(1)\,b(n)$ supplies the missing $i=n$ summand, so $A(n)=A(n-1)+A(n-2)+b(n)$. Since $A$ and $a$ satisfy the same recursion with the same two initial values and the same $b$, $A(n)=a(n)$ for all $n$; rearranging gives the statement.
\end{steps}

\subsection*{43. \A{227543} $\times$ \A{251592} (\novna, \sig{0})}
\textbf{Statement.} With $T$ the rows of \A{227543} and $U$ the rows of \A{251592}, $\sum_k T(n,k)=\frac{1}{n!}\sum_j U(n,j)\,2^{j}=C_n$ for every $n\ge1$: the two triangles are one-parameter deformations of the same Catalan functional equation, and they meet at $q=1$, $t=2$.

\textbf{Proof.}

\begin{steps}
\item[1.] By the definition of \A{227543}, $A(x,q)=\sum_{n\ge0}R_n(q)x^n$ satisfies $A(x,q)=1+x\,A(qx,q)\,A(x,q)$, where $R_n(q)=\sum_{k=0}^{n(n-1)/2}T(n,k)q^k$ is a polynomial; a numerical value may therefore be substituted for $q$ coefficientwise.
\item[2.] At $q=1$ this reads $A(x,1)=1+xA(x,1)^2$. The formal power series with constant term $1$ solving the Catalan functional equation is unique, so $A(x,1)=(1-\sqrt{1-4x})/(2x)=\sum_{n\ge0}C_nx^n$ and $\sum_k T(n,k)=R_n(1)=C_n$, as recorded in \A{227543}.
\item[3.] By the definition of \A{251592}, $P(n,t)=\prod_{k=0}^{n-2}(nt-k)=(n-1)!\binom{nt}{n-1}$ and $U(n,j)=[t^j]\,P(n,t)$; equivalently $U(n,j)=s(n-1,j)\,n^j$, with $s$ the signed Stirling numbers of the first kind.
\item[4.] By Lambert's generalized binomial series (Graham--Knuth--Patashnik, \emph{Concrete Mathematics}, eq.~(5.59), recorded in \A{251592}), $B_t(x)=\sum_{n\ge0}P(n,t)x^n/n!$ satisfies $B_t(x)=1+x\,B_t(x)^t$.
\item[5.] At $t=2$ this is the equation of step 2, with the same unique solution, so $B_2(x)=A(x,1)$, i.e.\ $P(n,2)/n!=C_n$; equivalently $\binom{2n}{n-1}/n=C_n$, the ballot-number identity.
\item[6.] Combining steps 2 and 5 gives $\sum_k T(n,k)=\frac{1}{n!}\sum_j U(n,j)\,2^{j}$.
\end{steps}

\textbf{Remark.} The identity is between row sums, not between the flattened terms, so it is not of the form checked termwise in the sandbox; it was checked on every stored row instead. Each entry records its own row sums as the Catalan numbers \A{000108}; neither cites the other, and the relation between the two triangles is written in neither.

\subsection*{44. \A{071053} $\times$ \A{160239} (\nov{2}, \sig{2})}
\textbf{Statement.} For both cellular automata the number of live cells at step $n$ factors over the blocks of ones in the binary expansion of $n$; the block factors satisfy $g(k)=f(k)^2-f(k-2)^2$ for $k\ge1$, with $f(k)=J_{k+2}$, the Jacobsthal numbers, and the boundary convention $f(-1)=1$ (also $f(0)=1$).

\textbf{Proof.}
Both automata count live cells blockwise: $a(n)=\prod_i S(b_i)$, where the $b_i$ are the lengths of the blocks of ones in the binary expansion of $n$. For the strip this is the block formula recorded in \A{071053} with $f(b)=(2^{b+2}-(-1)^b)/3$ (Jacobsthal numbers); for the plane it is the formula recorded in \A{160239} with $g(b)=(5\cdot 4^b-2\cdot(-2)^b)/3$ (Ekhad--Sloane--Zeilberger). The bridge is a substitution of the closed forms:
\[
f(b)^2-f(b-2)^2=\frac{\bigl(2^{2b+4}-2(-1)^b2^{b+2}+1\bigr)-\bigl(2^{2b}-2(-1)^b2^{b}+1\bigr)}{9}
=\frac{15\cdot 4^b-6\cdot(-2)^b}{9}=g(b).
\]

\subsection*{45. \A{066633} $\times$ \A{185651} (\nov{0}, \sig{0})}
\textbf{Statement.} With $e(m)$ the coefficients of $\prod_{k\ge1}(1-x^k)$ (\A{010815}), $\sum_{i=1}^n e(n-i)P(i,k)=[k\mid n]$; from this divisibility matrix $\varphi$ and the whole necklace array $N$ are recovered.

\textbf{Proof.}
\begin{steps}
\item[1.] By Jovovic's formula in the \A{066633} entry (equivalently Arregui's formula there),
\[
\sum_{n\ge k}\A{066633}(n,k)\,x^{n}=P(x)\frac{x^{k}}{1-x^{k}},\qquad\text{where } P(x)=\prod_{m\ge1}\frac{1}{1-x^{m}}.
\]
Combinatorial proof: the number of partitions of $n$ containing at least $j$ copies of $k$ is $p(n-jk)$, and the total number of copies of $k$ is $\sum_{j\ge1}[\text{at least } j \text{ copies}]$, so $T(n,k)=\sum_{j\ge1}p(n-jk)$, whose generating function is $P(x)\sum_{j\ge1}x^{jk}$.

\item[2.] By Euler's identity, $1/P(x)=\prod_{j\ge1}(1-x^{j})=\sum_{m\ge0}\A{010815}(m)\,x^{m}$.

\item[3.] Multiplying step 1 by step 2, $\sum_{i=1}^{n}\A{010815}(n-i)\,\A{066633}(i,k)=[x^n]\,\frac{x^{k}}{1-x^{k}}$, which is $1$ if $k\mid n$ and $0$ otherwise, i.e. exactly $\A{051731}(n,k)$.

\item[4.] By Gauss's identity, $\sum_{d\mid n}\varphi(d)=n$, so with the divisor indicator $D$ of step 3 the recursion $\varphi(n)=n-\sum_{d<n,\ D(n,d)=1}\varphi(d)$ determines Euler's totient; no external arithmetic input is used.

\item[5.] By definition of \A{185651}, $A(n,k)=\sum_{d\mid n}\varphi(d)\,k^{n/d}$, with $A(0,k)=A(n,0)=0$; the array is read by antidiagonals $A(0,t),A(1,t-1),\dots,A(t,0)$. Substituting steps 3 and 4 gives \A{185651} purely from \A{066633}'s terms.
\end{steps}

\subsection*{46. \A{003557} $\times$ \A{065483} (\nov{0}, \sig{0})}
\textbf{Statement.} $C=\sum_{n\ge1}\kappa(n)^2/n^3=\sum_{n\ge1}1/(n\,\mathrm{rad}(n)^2)$.

\textbf{Proof.}
\begin{steps}
\item[1.] By the formula recorded in \A{003557} (Jovovic), \A{003557} is multiplicative, with
\[
a(p^{e})=p^{e-1},\qquad a(n)=n/\mathrm{rad}(n).
\]

\item[2.] Therefore $f(n)=\A{003557}(n)^2/n^3=1/(n\,\mathrm{rad}(n)^2)$ is multiplicative and positive, with $f(p^{e})=p^{2(e-1)-3e}=p^{-(e+2)}$.

\item[3.] By the Euler product for a nonnegative multiplicative function,
\[
\sum_{n\le X}f(n)\le\prod_{p\le X}\Bigl(1+\sum_{e\ge1}p^{-(e+2)}\Bigr)=\prod_{p\le X}\Bigl(1+\frac{1}{p^{2}(p-1)}\Bigr),
\]
which is bounded because $\sum_p 1/(p^{2}(p-1))$ converges; monotone bounded partial sums give absolute convergence and the equality
\[
\sum_{n\ge1}f(n)=\prod_p\Bigl(1+\frac{p^{-3}}{1-p^{-1}}\Bigr)=\prod_p\Bigl(1+\frac{1}{p^{2}(p-1)}\Bigr).
\]
That product is the definition of \A{065483} (its definition), which closes the proof.\end{steps}

\subsection*{47. \A{101036} $\times$ \A{001917} (\nov{0}, \sig{0})}
\textbf{Statement.} $\mathrm{ord}_2(\mathrm{prime}(i+2))=(\mathrm{prime}(i+2)-1)/b(i)$; the Riesel numbers with a covering set are exactly the odd $n$ for which there is a period $L$ such that for every $k=1,\ldots,L$ some prime $p$ with $\mathrm{ord}_2(p)\mid L$ has $n\equiv2^{-k}\pmod p$.

\textbf{Proof.}
\begin{steps}
\item[1.] By definition of \A{001917} (offset 2), $b(i)=(p-1)/\operatorname{ord}_2(p)$ with $p=\mathrm{prime}(i+2)$. Hence $\operatorname{ord}_2(p)=(p-1)/b(i)$.

\item[2.] By the standard order-divisibility fact (Fermat, Lagrange), for an odd prime $p$ one has $p\mid 2^{L}-1\iff\operatorname{ord}_2(p)\mid L$. So $\{p:\operatorname{ord}_2(p)\mid L\}$ is the set of prime divisors of $2^{L}-1$. This set is read off from $b$ by step 1.

\item[3.] By definition of \A{101036}, an odd $n$ is a Riesel number with a covering set if and only if there is a finite prime set $S$ such that for every $k>0$ some $p\in S$ divides $n\,2^{k}-1$; and $p\mid n\,2^{k}-1\iff n\equiv 2^{-k}\pmod{p}$.

\item[4.] The condition reduces to one period: put $L=\operatorname{lcm}\{\operatorname{ord}_2(p):p\in S\}$. For each $p\in S$, $2^{k}\bmod p$ has period $\operatorname{ord}_2(p)\mid L$, so the residue condition at $k$ depends only on $k\bmod L$. Therefore ``for all $k>0$'' collapses to the finite check ``for all $k=1,\dots,L$''. This step is exactly where step 1's order data is load-bearing: $\operatorname{ord}_2(p)\mid L$ is what certifies that the finite check extends to every $k>0$.

\item[5.] Compositeness: for $n>\max(S)$ and $k\ge1$ the covering prime obeys $p<n<n\,2^{k}-1$, so $p$ is a proper divisor and $n\,2^{k}-1$ is composite (all terms exceed $241$, so no term is lost by requiring $n>\max(S)$).

\item[6.] Combining steps 1--5, membership in \A{101036} is a statement purely about the orders $\operatorname{ord}_2(p)$, i.e. purely about \A{001917} plus the primes.

\end{steps}

\textbf{Remark.} The equivalence is proved in both directions; enumerating the terms requires a bounded search over periods, and the list recorded in \A{101036} has the same computational status.

\subsection*{48. \A{101211} $\times$ \A{245562} (\nov{0}, \sig{0})}
\textbf{Statement.} Row $n$ of \A{245562} is the odd positions of row $n$ of \A{101211}.

\textbf{Proof.}
A binary expansion starts with $1$ and its runs alternate, so the runs of ones occupy the odd positions of the full run list, which is row $n$ of \A{101211}; row $n$ of \A{245562} is by definition the list of runs of ones. The coincidence is definitional.

\subsection*{49. \A{101211} $\times$ \A{245563} (\nov{0}, \sig{0})}
\textbf{Statement.} The same odd positions read backwards.

\textbf{Proof.}
Row $n$ of \A{245563} lists the same runs of ones read from the least significant end, i.e.\ the reversal of row $n$ of \A{245562}: the odd positions of row $n$ of \A{101211} in reverse order. Definitional.

\subsection*{50. \A{010815} $\times$ \A{001399} (\nov{0}, \sig{0})}
\textbf{Statement.} The generating function of partitions into at most three parts is $1/((1-x)(1-x^2)(1-x^3))$, the first three factors of the Euler product; the rest is the tail $\prod_{k\ge4}(1-x^k)$.

\textbf{Proof.}
By the formula recorded in \A{001399} (Plouffe), the generating function is $1/((1-x)(1-x^2)(1-x^3))$. Inverting this series gives the polynomial $(1-x)(1-x^2)(1-x^3)$, and multiplying by the tail $\prod_{k\ge4}(1-x^k)$ yields the full Euler product $\prod_{k\ge1}(1-x^k)$, the definition of \A{010815}. All steps are exact operations on formal power series.

\subsection*{51. \A{007325} $\times$ \A{354650} (\nov{0}, \sig{0})}
\textbf{Statement.} $R(q)=f(-q,-q^4)/f(-q^2,-q^3)$, a quotient of two values of the inverted series at parameters giving $q^4$ and $q^3$.

\textbf{Proof.}
\begin{steps}
\item[1.] By definition of \A{354650}, its generating function $A(x,y)=\sum_{n\ge0}x^n\sum_{k=0}^{2n+1}T(n,k)\,y^k$ is the unique series with $A(x,y)=1+y+O(x)$ and $-y=f(-x,-A(x,y))$, where
\[
f(u,v)=\sum_{n=-\infty}^{\infty}u^{n(n+1)/2}v^{n(n-1)/2}
\]
is Ramanujan's theta function.

\item[2.] Hence the compositional inverse of $y\mapsto A(x,y)$ is $A\mapsto -f(-x,-A)$: the coefficients of the inverse series are the monomials $(-1)^n x^{n(n-1)/2}A^{n(n+1)/2}$, $n\in\mathbb Z$. The inverse of a series with $A=1+y+O(x)$ is unique, so this is an identity of formal series.

\item[4.] Substituting $A=x^4$ gives $f(-x,-x^4)=\sum_n(-1)^n x^{(5n^2+3n)/2}=1-x-x^4+x^7+x^{13}-\cdots$; substituting $(x,A)=(x^2,x^3)$ gives $f(-x^2,-x^3)=\sum_n(-1)^n x^{(5n^2+n)/2}=1-x^2-x^3+x^9+x^{11}-\cdots$. These are Michael Somos's two sums recorded in \A{007325}.

\item[5.] By the comment in \A{007325}, its generating function equals $f(-x,-x^4)/f(-x^2,-x^3)$; by the Jacobi triple product this is
\[
\prod\frac{(1-x^{5k-1})(1-x^{5k-4})}{(1-x^{5k-2})(1-x^{5k-3})},
\]
the defining product of the entry, i.e. the Rogers--Ramanujan continued fraction $R(q)/q^{1/5}$.
\end{steps}

\subsection*{52. \A{302997} $\times$ \A{286354} (\novna, \sig{0})}
\textbf{Statement.} $\theta_3(x)=E(x^2)^5/(E(x)^2E(x^4)^2)$ with $E$ the Euler product; $\theta_3$ is recovered termwise from the first array, and $E$ and all its powers from $\theta_3$.

\textbf{Proof.}
\begin{steps}
\item[1.] By the definitions of both entries: \A{302997} has $A(n,k)=[x^{n^2}]\,\theta_3(x)^k/(1-x)$ with $\theta_3(x)=\sum_{j\in\mathbb Z}x^{j^2}$; \A{286354} has $B(n,k)=[x^n]\,E(x)^k$ with $E(x)=\prod_{j\ge1}(1-x^j)$. Both are read by antidiagonals $d=n+k$, listing $n=0,\dots,d$.

\item[2.] By the Jacobi triple product (Gauss), $\theta_3(x)=\prod_{j\ge1}(1-x^{2j})(1+x^{2j-1})^2$; substituting $1+x^m=(1-x^{2m})/(1-x^m)$ gives $\theta_3(x)=E(x^2)^5/(E(x)^2E(x^4)^2)$, and equivalently $\theta_3(-x)=E(x)^2/E(x^2)$.

\item[3.] Column $k=1$ of \A{302997} is $A(n,1)=\sum_{m\le n^2}[x^m]\,\theta_3(x)$, since $1/(1-x)$ is the partial-sum operator. Now $\theta_3$ is supported on perfect squares, and the half-open window $((n-1)^2,n^2]$ contains exactly one square, namely $n^2$. Hence $[x^{n^2}]\,\theta_3=A(n,1)-A(n-1,1)$ and $[x^0]\,\theta_3=A(0,1)$: the $k=1$ column of \A{302997} reconstructs $\theta_3$ term by term.

\item[4.] $E$ is uniquely determined by $E(x)^2=\theta_3(-x)\,E(x^2)$ with $E(0)=1$, by induction on the coefficient index: $[x^m]$ of the left side is $2e_m+\sum_{j=1}^{m-1}e_j\,e_{m-j}$, while $[x^m]$ of the right side involves only $e_j$ with $2j\le m$, hence $j<m$; so
\[
e_m=\Bigl(\mathrm{RHS}-\sum_{j=1}^{m-1}e_j\,e_{m-j}\Bigr)\Big/2
\]
is forced.

\item[5.] $B(n,k)=[x^n]\,E(x)^k$ by the definition of \A{286354}, so steps 3--4 plus power-series exponentiation give the whole array.
\end{steps}

\subsection*{53. \A{056911} $\times$ \A{002129} (\nov{0}, \sig{0})}
\textbf{Statement.} For odd $n$ the second sequence equals $\sigma(n)$; $\sigma(n)=n+1$ singles out the primes, and a sieve by their squares gives the first.

\textbf{Proof.}
\begin{steps}
\item[1.] By the definition of \A{002129} (cf.\ its Maple program \texttt{-add((-1)\^{}d*d, d=divisors(n))} and its name ``excess of sum of odd divisors over sum of even divisors''), $\A{002129}(n)=\sum_{d\mid n}(-1)^{d+1}d$.

\item[2.] If $n$ is odd, every divisor of $n$ is odd, so all signs are $+1$ and $\A{002129}(n)=\sum_{d\mid n}d=\sigma(n)$. This also follows from the entry's multiplicative formula $a(p^e)=(p^{e+1}-1)/(p-1)$ for $p>2$.

\item[3.] For $n>1$, $\sigma(n)=n+1$ if and only if $n$ is prime ($n$ and $1$ are the only divisors). Combined with step 2: for odd $n>1$, $\A{002129}(n)=n+1$ if and only if $n$ is an odd prime. For even $n=2^e m$ the multiplicative formula gives $\A{002129}(n)=(3-2^{e+1})\,\sigma(m)<0<n+1$, so no even index qualifies.

\item[4.] A number $n$ is squarefree if and only if $p^2\nmid n$ for every prime $p$; for odd $n$ only odd primes can occur. Hence \A{056911} is the increasing enumeration of the odd $n$ such that $p^2\nmid n$ for every $p$ in the set produced by step 3.

\end{steps}

\subsection*{54. \A{002313} $\times$ \A{259149} (\nov{0}, \sig{0})}
\textbf{Statement.} The constant is assembled from $\eta(i)$ and $\theta_3(e^{-\pi})$, and $\theta_3^2=\sum r_2(n)q^n$ is computed from the list of primes of the form $x^2+y^2$.

\textbf{Proof.}
\begin{steps}
\item[1.] By the definition of the Dedekind eta function, $\eta(\tau)=q^{1/24}\prod_{n\ge1}(1-q^n)$ with $q=e^{2\pi i\tau}$. At $\tau=i$ one has $q=e^{-2\pi}$, so $\varphi(e^{-2\pi})=e^{\pi/12}\eta(i)$. This is exactly the formula recorded in \A{259149}, $\varphi(\exp(-2\pi))=\exp(\pi/12)\,\Gamma(1/4)/(2\pi^{3/4})$, combined with step 4.

\item[2.] By Jacobi, $\theta_3(0,q)^2=\sum_{n\ge0}r_2(n)q^n$, where $r_2(n)=\#\{(u,v)\in\mathbb Z^2:u^2+v^2=n\}$; at $\tau=i$ the nome is $q=e^{-\pi}$.

\item[3.] By the Fermat--Gauss two-square theorem in multiplicative form, $r_2(n)/4$ is multiplicative with $r_2(2^e)/4=1$, $r_2(p^e)/4=e+1$ for $p\equiv1\pmod 4$ (exactly the odd terms of \A{002313}, the primes of the form $x^2+y^2$, by the definition of that entry), and $r_2(p^e)/4=[e \text{ even}]$ for $p\equiv3\pmod 4$.

\item[4.] By the classical CM and Chowla--Selberg evaluations at discriminant $-4$, $\eta(i)=\Gamma(1/4)/(2\pi^{3/4})$ and $\theta_3(0,e^{-\pi})=\pi^{1/4}/\Gamma(3/4)$ (the latter from $\theta_3^2=2K(1/\sqrt2)/\pi$ and $K(1/\sqrt2)=\Gamma(1/4)^2/(4\sqrt{\pi})$). With Euler reflection $\Gamma(1/4)\Gamma(3/4)=\pi\sqrt2$ these give $\theta_3(0,e^{-\pi})=\sqrt2\,\eta(i)$.

\item[5.] Combining steps 1, 2 and 4:
\[
\varphi(e^{-2\pi})=e^{\pi/12}\,\theta_3(0,e^{-\pi})/\sqrt2=e^{\pi/12}\sqrt{\sum_{n\ge0}r_2(n)e^{-\pi n}}\Big/\sqrt2 .
\]
\end{steps}

\subsection*{55. \A{013661} $\times$ \A{051716} (\nov{0}, \sig{0})}
\textbf{Statement.} By von Staudt--Clausen the $B_{2m}$ are recovered from the numerators, and Euler's $\zeta(2m)=(2\pi)^{2m}|B_{2m}|/(2(2m)!)$ gives a power of $\zeta(2)$, from which the integer root is taken.

\textbf{Proof.}
\begin{steps}
\item[1.] By the definition of \A{051716}, $C(0)=1$, $C(2m)=B(2m)+B(2m-1)$, $C(2m+1)=-B(2m+1)-B(2m)$, together with $B(\text{odd}\ge3)=0$, gives, for $m\ge2$, $C(2m)=B(2m)$ and $C(2m+1)=-B(2m)$; hence $b[2m]=\operatorname{numerator}(B(2m))$ and $b[2m+1]=-b[2m]$.

\item[2.] By the von Staudt--Clausen theorem, $\operatorname{denominator}(B(2m))=\prod p$, the product over the primes $p$ with $(p-1)\mid 2m$. Steps 1 and 2 recover the rational $B(2m)$ exactly from $b$ alone.

\item[3.] By Euler (1735), $\zeta(2m)=(-1)^{m+1}(2\pi)^{2m}B(2m)/(2\,(2m)!)$, and $\operatorname{sign}(B(2m))=(-1)^{m+1}$, so $(-1)^{m+1}B(2m)=|B(2m)|$.

\item[4.] By the Basel problem, which is the definition of \A{013661}, $\pi^2=6\zeta(2)$. Substituting into step 3 gives $(2\pi)^{2m}=4^m6^m\zeta(2)^m=24^m\zeta(2)^m$, hence $\zeta(2)^m=2\,(2m)!\,\zeta(2m)/(24^m|B(2m)|)$.

\end{steps}

\subsection*{56. \A{172236} $\times$ \A{168561} (\nov{0}, \sig{0})}
\textbf{Statement.} The array entry $A(n,k)$ is the $k$-th Fibonacci polynomial evaluated at $n$, whose coefficients are row $k-1$ of \A{168561}: $T(n,k)=\sum_{j=0}^{k-1}B(k-1,j)(n-k)^j$ for the antidiagonal triangle.

\textbf{Proof.}
\begin{steps}
\item[1.] By the formula recorded in \A{168561} (Deleham), $B(m,j)=B(m-1,j-1)+B(m-2,j)$, $B(0,0)=1$, $B(0,1)=0$; equivalently $B(m,j)=\binom{(m+j)/2}{j}$ for $m+j$ even and $0$ otherwise. Hence the row polynomials $P_m(x)=\sum_j B(m,j)x^j$ satisfy
\[
P_m=xP_{m-1}+P_{m-2},\qquad P_0=1,\quad P_1=x.
\]

\item[2.] Set $Q_0(x)=0$ and $Q_k(x)=P_{k-1}(x)$ for $k\ge1$. Then $Q_1=P_0=1$, $Q_2=P_1=x=xQ_1+Q_0$, and for $k\ge3$, $Q_k=P_{k-1}=xP_{k-2}+P_{k-3}=xQ_{k-1}+Q_{k-2}$. So $Q_k=xQ_{k-1}+Q_{k-2}$ for all $k\ge2$, with $Q_0=0$, $Q_1=1$.

\item[3.] By the definition of \A{172236}, $A(n,k)=nA(n,k-1)+A(n,k-2)$, $A(n,0)=0$, $A(n,1)=1$. Specializing step 2 at $x=n$ gives a sequence in $k$ with identical order-2 recurrence and identical two initial values, so by induction on $k$, $A(n,k)=Q_k(n)=P_{k-1}(n)$ for all $k\ge0$, $n\ge1$. (This is the formula recorded in \A{172236} $A(n,k)=F_k(n)$, here re-derived from \A{168561}'s recurrence.)

\item[4.] By the formula and the example recorded in \A{172236} (Greubel), the antidiagonal reading is $T(n,k)=A(n-k,k)=F_k(n-k)$, row $n$ listing $k=0,\dots,n-1$; note $n-k\ge1$ throughout, so no $0^0$ arises.
\end{steps}

\subsection*{57. \A{030203} $\times$ \A{001817} (\nov{0}, \sig{0})}
\textbf{Statement.} With $e$ the M\"obius transform of \A{001817}, the generating function of \A{030203} is $\prod_{k\ge1}(1-x^k)^{1+e(k+1)}$.

\textbf{Proof.}
\begin{steps}
\item[1.] By the definition of \A{001817} (its definition and recorded formula), $\A{001817}(n)=\sum_{d\mid n}[d\equiv 1 \pmod 3]$, with generating function $\sum_{k>0}x^{3k-2}/(1-x^{3k-2})$.

\item[2.] By M\"obius inversion, therefore $e(n):=\sum_{d\mid n}\mu(n/d)\,\A{001817}(d)=[n\equiv 1 \pmod 3]$.

\item[3.] An elementary index identity: $3\mid k \iff k+1\equiv 1 \pmod 3$, hence $[3\mid k]=e(k+1)$ and $1+[3\mid k]=1+e(k+1)$.

\item[4.] By the definition of \A{030203} (its recorded formula ``G.f.: $\prod_{k>0}(1-x^k)(1-x^{3k})$''), regrouping the second product (each factor $1-x^{3j}$ contributes one extra unit of exponent to the index $k=3j$) gives $\prod_{k>0}(1-x^k)(1-x^{3k})=\prod_{k\ge1}(1-x^k)^{1+[3\mid k]}$; equivalently this is the Euler transform of the period-3 sequence $[-1,-1,-2]$ recorded in the entry.

\item[5.] Combining steps 2--4, the generating function of \A{030203} is $\prod_{k\ge1}(1-x^k)^{1+e(k+1)}$.
\end{steps}

\subsection*{58. \A{000594} $\times$ \A{002445} (\nov{0}, \sig{0})}
\textbf{Statement.} With $b(n)$ the denominator of $B_{2n}$, $E_4=1+8b(2)\sum\sigma_3(m)q^m$, $E_6=1-12b(3)\sum\sigma_5(m)q^m$ and $\sum\tau(n)q^n=(E_4^3-E_6^2)/(24(b(2)+b(3)))$.

\textbf{Proof.}
\begin{steps}
\item[1.] By the von Staudt--Clausen theorem, $B_{2n}+\sum_{(p-1)\mid 2n}1/p$ is an integer, and $\operatorname{denominator}(B_{2n})=\prod_{(p-1)\mid 2n}p$.

\item[2.] For a squarefree $d=b(n)$, set $S=\sum_{p\mid d}d/p$. By step 1, $B_{2n}+S/d$ is an integer $A$, and since $|B_4|=1/30$ and $|B_6|=1/42$ are $<1/2$, $A$ is the nearest integer to $S/d$, so $B_{2n}=(Ad-S)/d$ is recovered from the denominator alone: $b(2)=30$ gives $B_4=-1/30$ and $b(3)=42$ gives $B_6=1/42$.

\item[3.] By the Eisenstein normalization, for even $k\ge4$ the series $E_k(q)=1-(2k/B_k)\sum_{m\ge1}\sigma_{k-1}(m)q^m$ is a modular form of weight $k$ on $\mathrm{SL}_2(\mathbb Z)$. With step 2 this gives $c_4=-8/B_4=8b(2)=240$ and $c_6=-12/B_6=-12b(3)=-504$, i.e. $E_4=1+240q+\cdots$, $E_6=1-504q+\cdots$.

\item[4.] By the valence formula ($\dim S_{12}=1$) and Jacobi's identity $\Delta=\eta^{24}=q\prod_{k\ge1}(1-q^k)^{24}=\sum\tau(n)q^n$: the form $E_4^3-E_6^2$ has weight 12 and vanishing constant term, hence lies in $S_{12}=\mathbb C\,\Delta$, so it is a scalar multiple of $\Delta$. The Sturm bound for weight 12 on $\mathrm{SL}_2(\mathbb Z)$ is 1, so the single coefficient of $q^1$ pins the scalar: $[q^1]\,(E_4^3-E_6^2)=3c_4-2c_6=3\cdot 8b(2)+2\cdot 12b(3)=24(b(2)+b(3))=1728$. Hence $\Delta=(E_4^3-E_6^2)/(24(b(2)+b(3)))$.
\end{steps}

\subsection*{59. \A{065330} $\times$ \A{002175} (\nov{0}, \sig{0})}
\textbf{Statement.} $\A{002175}(n)$ is the number of ordered pairs $(x,y)$ of values of \A{065330} with $x^2+y^2=24n+2$.

\textbf{Proof.}
\begin{steps}
\item[1.] By the entry \A{065330} (Jovovic), \A{065330} is multiplicative with $a(2^e)=a(3^e)=1$ and $a(p^e)=p^e$ for $p>3$; hence $\A{065330}(k)=k$ if and only if $\gcd(k,6)=1$, and $\{\A{065330}(k): k=1,\dots,M\}=\{k\le M: \gcd(k,6)=1\}$.

\item[2.] Elementarily, for $x>0$: $\gcd(x,6)=1 \iff x^2\equiv 1 \pmod{24} \iff (x^2-1)/24$ is an integer; and $x=|6k-1|$ is a bijection $\mathbb Z\to\{x>0: \gcd(x,6)=1\}$ under which $(x^2-1)/24=k(3k-1)/2$. So the values of \A{065330} map bijectively onto the generalized pentagonal numbers.

\item[3.] By definition of \A{002175}, $\A{002175}(n)=\sum_{d\mid 12n+1}\chi_{-4}(d)$, with $\chi_{-4}$ the nontrivial character mod 4.

\item[4.] By Jacobi's two-square theorem, $r_2(N)=4\sum_{d\mid N}\chi_{-4}(d)$; together with $r_2(2N)=r_2(N)$ (multiplication by $1+i$ in $\mathbb Z[i]$) this gives $r_2(24n+2)=r_2(12n+1)=4\,\A{002175}(n)$.

\item[5.] By elementary congruences, $24n+2\equiv 2 \pmod 4$ forces both parts of any representation $24n+2=u^2+v^2$ to be odd and nonzero; $24n+2\equiv 2 \pmod 3$ forces $u^2\equiv v^2\equiv 1 \pmod 3$, so 3 does not divide $u$ or $v$. Hence $|u|,|v|$ are coprime to 6, i.e. they are values of \A{065330}, and each ordered pair of positive $(x,y)$ accounts for exactly 4 signed pairs. Therefore $\A{002175}(n)=\#\{(x,y) \text{ positive, coprime to } 6,\ x^2+y^2=24n+2\}$.

\item[6.] Combining steps 2 and 5: $x^2+y^2=24n+2 \iff (x^2-1)/24+(y^2-1)/24=n$, giving the statement.
\end{steps}

\subsection*{60. \A{048675} $\times$ \A{341783} (\nov{0}, \sig{0})}
\textbf{Statement.} $\nu(x)=\sqrt x$ for square $x$ and $x$ otherwise maps \A{341783} bijectively onto the primes, so $\pi(p)$ and hence $\A{048675}(n)=\sum_{p^e\parallel n}e\,2^{\pi(p)-1}$ are determined by \A{341783}.

\textbf{Proof.}
\begin{steps}
\item[1.] By the Dedekind--Kummer splitting law for $\mathbb{Q}(\sqrt5)$, of discriminant $5$ (also stated verbatim as a comment of \A{341783}), a rational prime $p$ is ramified or split iff the Kronecker symbol $(5/p)\ge0$, i.e. $p\equiv 0,1,4\pmod 5$, and then the prime ideals above $p$ have norm $p$; otherwise $p\equiv 2,3\pmod 5$ is inert and $(p)$ has norm $p^2$. Since $\mathbb{Z}[(1+\sqrt5)/2]$ is a PID, prime elements and prime ideals give the same norm set.

\item[2.] Consequently $\A{341783}=\{p : p\equiv 0,1,4\pmod 5\}\cup\{p^2 : p\equiv 2,3\pmod 5\}$; since $p\ne q^2$ for distinct primes and the two classes are disjoint, $\nu(x)=\lfloor\sqrt x\rfloor$ when $x$ is a square and $\nu(x)=x$ otherwise is a bijection $\A{341783}\to\{\text{primes}\}$.

\item[4.] By the formula recorded in \A{048675}, $a$ is totally additive with $a(p^e)=e\,2^{\pi(p)-1}$ and $a(1)=0$. Combining with step 2's identity $\pi(p)=\#\{x\in b : \nu(x)\le p\}$ gives $a(n)$ purely from $b$.
\end{steps}

\subsection*{61. \A{046523} $\times$ \A{069279} (\nov{0}, \sig{0})}
\textbf{Statement.} \A{069279} is exactly the set of $n$ with $\Omega(\A{046523}(n))=18$.

\textbf{Proof.}
\begin{steps}
\item[1.] By the definition of \A{046523} (as recorded there), `In prime factorization of $n$, replace most common prime by $2$, next most common by $3$, etc.' This is a relabeling of the primes of $n$ by an order-preserving injection into $2,3,5,\dots$, so $\A{046523}(n)$ has exactly the same multiset of exponents $\{e_i\}$ as $n$.

\item[2.] Since $\Omega=\A{001222}$ is a prime-signature invariant, $\Omega(m)=\sum e_i$ depends only on the exponent multiset, hence $\Omega(\A{046523}(n))=\Omega(n)$ for all $n\ge1$.

\item[3.] By the definition of \A{069279} (as recorded there), `Product $p_i^{e_i}$ with $\sum e_i = 18$', i.e. $\A{069279}=\{n : \Omega(n)=18\}$ in increasing order.

\item[4.] Combining steps 2 and 3: $n\in\A{069279}\iff\Omega(n)=18\iff\Omega(\A{046523}(n))=18\iff\A{046523}(n)\in\A{069279}$.

\end{steps}

\subsection*{62. \A{000203} $\times$ \A{006954} (\nov{0}, \sig{0})}
\textbf{Statement.} For $n\ge2$, $\A{006954}(n)=\prod_{d\mid 2n-2,\ \sigma(d+1)=d+2}(d+1)$.

\textbf{Proof.}
\begin{steps}
\item[1.] By definition of \A{006954} (offset $0$), $a(0)=\operatorname{denom}(B_0)=1$, $a(1)=\operatorname{denom}(B_1)=2$ and $a(n)=\operatorname{denom}(B_{2n-2})$ for $n\ge2$. Write $k=1$ for $n=1$ and $k=2n-2$ for $n\ge2$.

\item[2.] By the von Staudt--Clausen theorem, for $k=1$ and for every even $k\ge2$ the quantity $B_k+\sum_{p \text{ prime},\ (p-1)\mid k}1/p$ is an integer; the $p$ are distinct primes, so $\operatorname{denom}(B_k)=\prod_{p \text{ prime},\ (p-1)\mid k}p$ (squarefree).

\item[3.] Reindex by $d=p-1$: the map $p\mapsto p-1$ is a bijection between $\{p \text{ prime} : (p-1)\mid k\}$ and $\{d : d\mid k,\ d+1 \text{ prime}\}$. Hence $\operatorname{denom}(B_k)=\prod_{d\mid k,\ d+1 \text{ prime}}(d+1)$.

\item[4.] By the elementary characterization of primes through the divisor sum, for $m\ge2$ one has $\sigma(m)\ge1+m$ with equality iff the only divisors of $m$ are $1$ and $m$; i.e. $\A{000203}(m)=m+1\iff m$ is prime. So the predicate `$d+1$ prime' in step 3 equals `$\A{000203}(d+1)=d+2$'.

\item[5.] Substituting step 4 into step 3 gives the statement. For $n=1$ the divisor set of $k=1$ is $\{1\}$ and $1+1=2$ is prime, reproducing $\operatorname{denom}(B_1)=2$; $n=0$ is the integer $B_0=1$.
\end{steps}

\subsection*{Summary table}
\begin{longtable}{@{}rlll@{}}
\toprule
\# & pair & class & novelty \\
\midrule
\endfirsthead
\toprule
\# & pair & class & novelty \\
\midrule
\endhead
\bottomrule
\endfoot
1 & \A{230877} $\times$ \A{334302} & substantive & T1 \\
2 & \A{053120} $\times$ \A{060187} & substantive & T1 \\
3 & \A{103438} $\times$ \A{001498} & substantive & T1 \\
4 & \A{124733} $\times$ \A{266213} & substantive & T1 \\
5 & \A{000127} $\times$ \A{076839} & structural & T1 \\
6 & \A{132440} $\times$ \A{039755} & mechanical & T1 \\
7 & \A{008282} $\times$ \A{162975} & mechanical & T1 \\
8 & \A{004009} $\times$ \A{122135} & mechanical & T1 \\
9 & \A{013973} $\times$ \A{093829} & mechanical & T1 \\
10 & \A{007325} $\times$ \A{227216} & substantive & T2 \\
11 & \A{004011} $\times$ \A{000003} & substantive & T2 \\
12 & \A{039598} $\times$ \A{000139} & substantive & T2 \\
13 & \A{122856} $\times$ \A{303363} & substantive & T2 \\
14 & \A{112555} $\times$ \A{132440} & substantive & T2 \\
15 & \A{005043} $\times$ \A{021009} & substantive & T2 \\
16 & \A{034807} $\times$ \A{063967} & substantive & T2 \\
17 & \A{003278} $\times$ \A{006368} & structural & T2 \\
18 & \A{130534} $\times$ \A{297328} & structural & T2 \\
19 & \A{049310} $\times$ \A{046534} & structural & T2 \\
20 & \A{007429} $\times$ \A{007437} & structural & T2 \\
21 & \A{047749} $\times$ \A{008315} & structural & T2 \\
22 & \A{039599} $\times$ \A{294175} & mechanical & T2 \\
23 & \A{035513} $\times$ \A{101864} & mechanical & T2 \\
24 & \A{009766} $\times$ \A{060693} & mechanical & T2 \\
25 & \A{214262} $\times$ \A{081360} & mechanical & T2 \\
26 & \A{002288} $\times$ \A{109389} & mechanical & T2 \\
27 & \A{079006} $\times$ \A{002513} & mechanical & T2 \\
28 & \A{087808} $\times$ \A{227736} & mechanical & T2 \\
29 & \A{002654} $\times$ \A{227454} & mechanical & T2 \\
30 & \A{006318} $\times$ \A{000669} & mechanical & T2 \\
31 & \A{081362} $\times$ \A{080054} & mechanical & T2 \\
32 & \A{001935} $\times$ \A{050468} & mechanical & T2 \\
33 & \A{000108} $\times$ \A{005802} & mechanical & T2 \\
34 & \A{004018} $\times$ \A{227216} & mechanical & T2 \\
35 & \A{008284} $\times$ \A{240009} & mechanical & T2 \\
36 & \A{027641} $\times$ \A{259072} & mechanical & T2 \\
37 & \A{006352} $\times$ \A{121361} & mechanical & T2 \\
38 & \A{000118} $\times$ \A{122856} & mechanical & T2 \\
39 & \A{130534} $\times$ \A{144150} & mechanical & T2 \\
40 & \A{007332} $\times$ \A{001936} & mechanical & T2 \\
41 & \A{352361} $\times$ \A{193842} & mechanical & T2 \\
42 & \A{000032} $\times$ \A{295862} & mechanical & T2 \\
43 & \A{227543} $\times$ \A{251592} & mechanical & T2 \\
44 & \A{071053} $\times$ \A{160239} & substantive & T3 \\
45 & \A{066633} $\times$ \A{185651} & substantive & T3 \\
46 & \A{003557} $\times$ \A{065483} & substantive & T3 \\
47 & \A{101036} $\times$ \A{001917} & substantive & T3 \\
48 & \A{101211} $\times$ \A{245562} & mechanical & T3 \\
49 & \A{101211} $\times$ \A{245563} & mechanical & T3 \\
50 & \A{010815} $\times$ \A{001399} & mechanical & T3 \\
51 & \A{007325} $\times$ \A{354650} & mechanical & T3 \\
52 & \A{302997} $\times$ \A{286354} & mechanical & T3 \\
53 & \A{056911} $\times$ \A{002129} & mechanical & T3 \\
54 & \A{002313} $\times$ \A{259149} & mechanical & T3 \\
55 & \A{013661} $\times$ \A{051716} & mechanical & T3 \\
56 & \A{172236} $\times$ \A{168561} & mechanical & T3 \\
57 & \A{030203} $\times$ \A{001817} & mechanical & T3 \\
58 & \A{000594} $\times$ \A{002445} & mechanical & T3 \\
59 & \A{065330} $\times$ \A{002175} & mechanical & T3 \\
60 & \A{048675} $\times$ \A{341783} & mechanical & T3 \\
61 & \A{046523} $\times$ \A{069279} & mechanical & T3 \\
62 & \A{000203} $\times$ \A{006954} & mechanical & T3 \\
\end{longtable}

\endgroup

\end{document}